\documentclass[acmtog]{acmart}
\acmSubmissionID{357}

\usepackage{booktabs} %
\usepackage{acronym} %
\usepackage[normalem]{ulem} 
\usepackage{balance}

\usepackage[ruled]{algorithm2e} %

\SetAlFnt{\small}
\SetAlCapFnt{\small}
\SetAlCapNameFnt{\small}
\SetAlCapHSkip{0pt}

\acmJournal{TOG}

\setcopyright{rightsretained}
\acmJournal{TOG}
\acmYear{2021}
\acmVolume{40}
\acmNumber{4}
\acmArticle{88}
\acmMonth{8}

\acmDOI{10.1145/3450626.3459936}

\newcommand{\qheading}[1]{\noindent\textbf{#1}}
\newcommand{\modelname}{DECA\xspace}
\newcommand{\supmat}{Sup.~Mat.\xspace}
\newcommand{\imsize}{0.48}
\newcommand{\varhspace}{0.065}

\acrodef{PCA}{Principal Component Analysis}
\acrodef{GAN}{generative adversarial network}
\acrodef{MDL}{minimum description length}
\acrodef{LBS}{linear blend skinning}
\acrodef{SfS}{shape from shading}

\begin{document}
\title{Learning an Animatable Detailed 3D Face Model from In-The-Wild Images}

\author{Yao Feng}
\authornote{Both authors contributed equally to the paper}
\affiliation{%
\institution{Max Planck Institute for Intelligent Systems}
\city{T{\"u}bingen}}
\affiliation{%
\institution{Max Planck ETH Center for Learning System}
\city{T{\"u}bingen}
\country{Germany}}
\email{yfeng@tuebingen.mpg.de}

\author{Haiwen Feng}
\authornotemark[1]
\affiliation{%
\institution{Max Planck Institute for Intelligent Systems}
\city{T{\"u}bingen}
\country{Germany}}
\email{hfeng@tuebingen.mpg.de}

\author{Michael J. Black}
\affiliation{%
\institution{Max Planck Institute for Intelligent Systems}
\city{T{\"u}bingen}
\country{Germany}}
\email{black@tuebingen.mpg.de}

\author{Timo Bolkart}
\affiliation{%
\institution{Max Planck Institute for Intelligent Systems}
\city{T{\"u}bingen}
\country{Germany}}
\email{tbolkart@tuebingen.mpg.de}

\renewcommand\shortauthors{Yao Feng, Haiwen Feng, Michael J. Black, Timo Bolkart}

\begin{abstract}
While current monocular 3D face reconstruction methods can recover fine geometric details, they suffer several limitations.
Some methods produce faces that cannot be realistically animated because they do not model how wrinkles vary with expression.
Other methods are trained on high-quality face scans and do not generalize well to in-the-wild images.
We present the first approach that regresses 3D face shape and animatable details that are specific to an individual but change with expression.
Our model, \modelname (Detailed Expression Capture and Animation), is trained to robustly produce a UV displacement map from a low-dimensional latent representation that consists of person-specific detail parameters and generic expression parameters, while a regressor is trained to predict detail, shape, albedo, expression, pose and illumination parameters from a single image. 
To enable this, we introduce a novel detail-consistency loss that disentangles person-specific details from expression-dependent wrinkles.
This disentanglement allows us to synthesize realistic person-specific wrinkles by controlling expression parameters while keeping person-specific details unchanged. 
\modelname is learned from in-the-wild images with no paired 3D supervision and
achieves state-of-the-art shape reconstruction accuracy on two benchmarks. 
Qualitative results on in-the-wild data demonstrate \modelname's robustness and its ability to disentangle identity- and expression-dependent details enabling animation of reconstructed faces.
The model and code are publicly available at \url{https://deca.is.tue.mpg.de}.
\end{abstract}




\maketitle

\newcommand{\vect}[1]{\mathbf{#1}}

\newcommand{\norm}[1]{\left\lVert#1\right\rVert}

\newcommand{\shapecoeff}{\boldsymbol{\beta}}
\newcommand{\shapedim}{{\left| \shapecoeff \right|}}
\newcommand{\shapespace}{\mathcal{S}}
\newcommand{\posecoeff}{\boldsymbol{\theta}}
\newcommand{\posedim}{{\left| \posecoeff \right|}}
\newcommand{\posespace}{\mathcal{P}}
\newcommand{\expcoeff}{\boldsymbol{\psi}}
\newcommand{\expdim}{{\left| \expcoeff \right|}}
\newcommand{\expspace}{\mathcal{E}}
\newcommand{\numverts}{n}
\newcommand{\template}{\textbf{T}}

\newcommand{\numjoints}{k}
\newcommand{\joints}{\textbf{J}}
\newcommand{\jointregressor}{\mathcal{J}}
\newcommand{\blendweights}{\mathcal{W}}
\newcommand{\blendweightsdim}{\left| \mathcal{W} \right|}

\newcommand{\landmark}{\textbf{k}}

\newcommand{\lighting}{\textbf{l}}
\newcommand{\cam}{\textbf{c}}

\newcommand{\albedo}{A}
\newcommand{\albedocoeffs}{\boldsymbol{\alpha}}
\newcommand{\albedodim}{\left| \albedocoeffs \right|}
\newcommand{\normalcoeffs}{\boldsymbol{\nu}}
\newcommand{\normaldim}{\left| \normalcoeffs \right|}

\newcommand{\uvsize}{d}
\newcommand{\image}{I}

\newcommand{\flamev}{M}
\newcommand{\normals}{N}

\newcommand{\displacements}{D}
\newcommand{\detailgeom}{G}

\newcommand{\coarseencoder}{E_c}
\newcommand{\detailencoder}{E_d}
\newcommand{\edm}{F_d}

\newcommand{\zcode}{\boldsymbol{\delta}}
\newcommand{\cons}{constraint}

\section{Introduction}

\begin{figure}[t]
    \centering
    \includegraphics[width=\columnwidth]{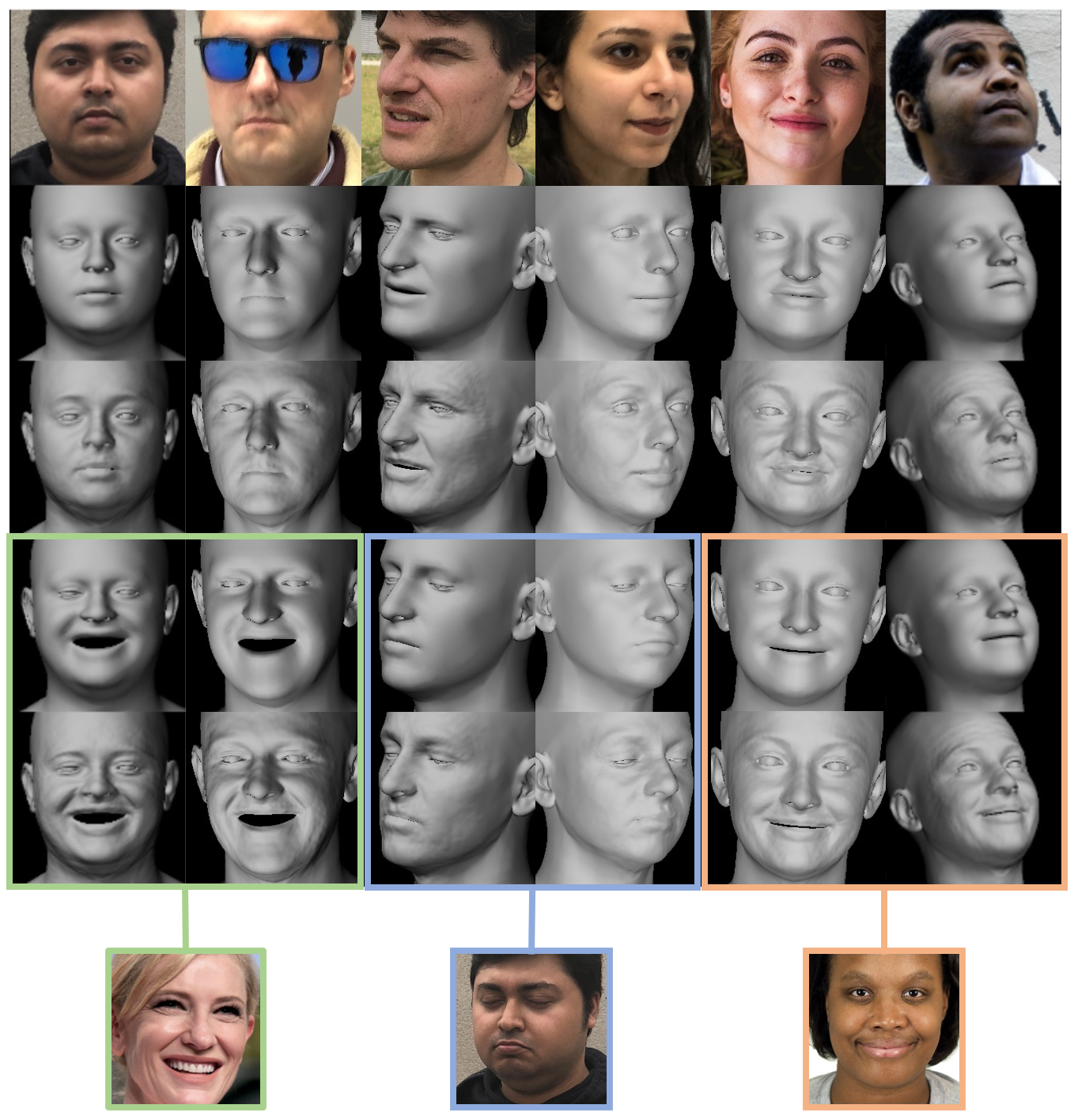}
	\caption{{\bf \modelname.}  Example images (row 1), the regressed coarse shape (row 2), detail shape (row 3) and reposed coarse shape (row 4), and reposed with person-specific details (row 5) where the source expression is extracted by \modelname from the faces in the corresponding colored boxes  (row 6).
	\modelname is robust to in-the-wild variations and captures person-specific details as well as expression-dependent wrinkles that appear in regions like the forehead and mouth.    
	Our novelty is that this detailed shape can be reposed ({\em animated}) such that the {\em  wrinkles are specific to the source shape and target expression}.
	Images are taken from Pexels~\shortcite{Pexels2021} (row 1; col. 5), Flickr~\shortcite{FlickrImg} (bottom left) @ Gage Skidmore, 
	Chicago~\cite{Chicago} (bottom right), and from NoW~\cite{Sanyal2019} (remaining images).
	}
    \label{fig:teaser}
\end{figure}

Two decades have passed since the seminal work of Vetter and Blanz~\shortcite{VetterBlanz1998} that first showed how to reconstruct 3D facial geometry from a single image.
Since then, 3D face reconstruction methods have rapidly advanced (for a comprehensive overview see~\cite{Morales2021,Zollhoefer2018}) enabling applications such as 3D avatar creation for VR/AR~\cite{Hu2017}, video editing~\cite{Thies2016,Kim2018_DeepVideo}, image synthesis~\cite{Ghosh2020,Tewari2020} face recognition~\cite{Blanz2002,Romdhani2002}, virtual make-up~\cite{Scherbaum2011}, or speech-driven facial animation~\cite{VOCA2019,Karras2017,Richard2021}.
To make the problem tractable, most existing methods incorporate prior knowledge about geometry or appearance by leveraging pre-computed 3D face models~\cite{Brunton2014,Egger2020}. 
These models reconstruct the coarse face shape but are unable to capture geometric details such as expression-dependent wrinkles, which are essential for realism and support analysis of human emotion.

Several methods recover detailed facial geometry~\cite{Abrevaya2020,Cao2015,Chen2019,Guo2018,Richardson2017,AnhTran2018,LuanTran2019}, however, they require high-quality training scans~\cite{Cao2015,Chen2019} or lack robustness to occlusions~\cite{Abrevaya2020,Guo2018,Richardson2017}. 
None of these explore how the recovered wrinkles change with varying expressions. 
Previous methods that learn expression-dependent detail models~\cite{chaudhuri2020personalized,yang2020facescape, Bickel2008} either use detailed 3D scans as training data and, hence, do not generalize to unconstrained images~\cite{yang2020facescape}, or model expression-dependent details as part of the appearance map rather than the geometry~\cite{chaudhuri2020personalized},  preventing realistic mesh relighting. 

We introduce \modelname (Detailed Expression Capture and Animation), which learns an {\em animatable} displacement model from in-the-wild images without 2D-to-3D supervision. 
In contrast to prior work, these {\em animatable expression-dependent wrinkles are specific to an individual} and are regressed from a single image.
Specifically, \modelname jointly learns
1) a geometric detail model that generates a UV displacement map from a low-dimensional representation that consists of subject-specific detail parameters and expression parameters, and
2) a regressor that predicts subject-specific detail, albedo, shape, expression, pose, and lighting parameters from an image. 
The detail model builds upon FLAME's~\cite{FLAME2017} coarse geometry, and we formulate the displacements as a function of subject-specific detail parameters and FLAME's jaw pose and expression parameters.

This enables important applications such as easy avatar creation from a single image.
While previous methods can capture detailed geometry in the image, most applications require a face that can be animated.
For this, it is not sufficient or recover accurate geometry in the input image.
Rather, we must be able to animate that detailed geometry and, more specifically, the details should be person specific.

To gain control over expression-dependent wrinkles of the reconstructed face, while preserving person-specific details (i.e.~moles, pores, eyebrows, and expression-independent wrinkles), the person-specific details and expression-dependent wrinkles must be disentangled. 
Our key contribution is a novel {\em detail consistency loss} that enforces this disentanglement.
During {\em training}, if we are given two images of the same person with different expressions, we observe that their 3D face shape and their person-specific details are the same in both images, but the expression and the intensity of the wrinkles differ with expression.
We exploit this observation during training by swapping the detail codes between different images of the same identity and enforcing the newly rendered results to look similar to the original input images. 
Once trained, \modelname reconstructs a detailed 3D face from a {\em single image} (Fig.~\ref{fig:teaser}, third row) in real time (about 120fps on a Nvidia Quadro RTX 5000), and is able to animate the reconstruction with realistic adaptive expression wrinkles (Fig.~\ref{fig:teaser}, fifth row). 

In summary, our main contributions are:
1) The first approach to learn an {\em animatable displacement model} from in-the-wild images that can synthesize plausible geometric details by varying expression parameters. 
2) A novel detail consistency loss that disentangles identity-dependent and expression-dependent facial details. 
3) Reconstruction of geometric details that is, unlike most competing methods, robust to common occlusions, wide pose variation, and illumination variation.
This is enabled by our low-dimensional detail representation, the detail disentanglement, and training from a large dataset of in-the-wild images.
4) State-of-the-art shape reconstruction accuracy on two different benchmarks. 
5) The code and model are available for research purposes  at \url{https://deca.is.tue.mpg.de}.
\section{Related work}

The reconstruction of 3D faces from visual input has received significant attention over the last decades after the pioneering work of Parke~\shortcite{Parke1974}, the first method to reconstruct 3D faces from multi-view images.
While a large body of related work aims to reconstruct 3D faces from various input modalities such as multi-view images~\cite{Beeler2010,Cao2018,Pighin1998}, video data~\cite{Garrido2016,Ichim2015,Jeni2015,Shi2014,Suwajanakorn2014}, RGB-D data~\cite{Li2013,Thies2015,Weise2011} or subject-specific image collections~\cite{KemelmacherSeitz2011,Roth2016}, our main focus is on methods that use only a single RGB image.
For a more comprehensive overview, see Zollh{\"o}fer et al.~\shortcite{Zollhoefer2018}.

\qheading{Coarse reconstruction:}
Many monocular 3D face reconstruction methods follow Vetter and Blanz~\shortcite{VetterBlanz1998} by estimating coefficients of pre-computed statistical models in an analysis-by-synthesis fashion. 
Such methods can be categorized into optimization-based~\cite{AldrianSmith2013,Bas2017fitting,Blanz2002,BlanzVetter1999,Gerig2018,RomdhaniVetter2005,Thies2016}, or learning-based methods~\cite{Chang2018,Deng2019,Genova2018,Kim2018,Ploumpis2020,Richardson2016,Sanyal2019,Tewari2017,AnhTran2017,Tu2019}. 
These methods estimate parameters of a statistical face model with a fixed linear shape space, which captures only low-frequency shape information.
This results in overly-smooth reconstructions. 

Several works are model-free and directly regress 3D faces (i.e. voxels~\cite{Jackson2017} or meshes~\cite{Dou2017,Feng2018,Guler2017,Wei2019}) and hence can capture more variation than the model-based methods. 
However, all these methods require explicit 3D supervision, which is provided either by an optimization-based model fitting~\cite{Feng2018,Guler2017,Jackson2017,Wei2019} or by synthetic data generated by sampling a statistical face model~\cite{Dou2017} and therefore also only capture coarse shape variations. 

Instead of capturing high-frequency geometric details, some methods reconstruct coarse facial geometry along with high-fidelity textures~\cite{Gecer2019,Saito2017,Slossberg2018,Yamaguchi2018}. 
As this ``bakes" shading details into the texture, lighting changes do not affect these details, limiting realism and the range of applications.
To enable animation and relighting, \modelname captures these details as part of the geometry.

\qheading{Detail reconstruction:}
Another body of work aims to reconstruct faces with ``mid-frequency" details.
Common optimization-based methods fit a statistical face model to images to obtain a coarse shape estimate, followed by a \ac{SfS} method to reconstruct facial details from monocular images~\cite{Jiang2018,Li2018,Riviere2020}, or videos~\cite{Garrido2016,Suwajanakorn2014}. 
Unlike \modelname, these approaches are slow, the results lack robustness to occlusions, and the coarse model fitting step requires facial landmarks, making them error-prone for large viewing angles and occlusions.

Most regression-based approaches~\cite{Cao2015,Chen2019,Guo2018,Lattas2020,Richardson2017,AnhTran2018} follow a similar approach by first reconstructing the parameters of a statistical face model to obtain a coarse shape, followed by a refinement step to capture localized details.
Chen et al.~\shortcite{Chen2019} and Cao et al.~\shortcite{Cao2015}  compute local wrinkle statistics from high-resolution scans and leverage these to constrain the fine-scale detail reconstruction from images~\cite{Chen2019} or videos~\cite{Cao2015}.
Guo et al.~\shortcite{Guo2018} and Richardson et al.~\shortcite{Richardson2017} directly regress per-pixel displacement maps.
All these methods only reconstruct fine-scale details in non-occluded regions, causing visible artifacts in the presence of occlusions. 
Tran et al.~\shortcite{AnhTran2018} gain robustness to occlusions by applying a face segmentation method~\cite{Nirkin2018} to determine occluded regions, and employ an example-based hole filling approach to deal with the occluded regions. 
Further, model-free methods exist that directly reconstruct detailed meshes~\cite{Sela2017,Zeng2019} or surface normals that add detail to coarse reconstructions~\cite{Abrevaya2020,Sengupta2018}.
Tran et al.~\shortcite{LuanTran2019} and Tewari et al.~\shortcite{Tewari2019,Tewari2018} jointly learn a statistical face model and reconstruct 3D faces from images. 
While offering more flexibility than fixed statistical models, these methods capture limited geometric details compared to other detail reconstruction methods. 
Lattas et al.~\shortcite{Lattas2020} use image translation networks to infer the diffuse normals and specular normals, resulting in realistic rendering.
Unlike \modelname, none of these detail reconstruction methods offer animatable details after reconstruction.

\qheading{Animatable detail reconstruction:}
Most relevant to \modelname are methods that reconstruct detailed faces while allowing animation of the result.
Existing methods \cite{Golovinskiy2006, Ma2008, Bickel2008, Shin2014, yang2020facescape} learn correlations between wrinkles or attributes like age and gender \cite{Golovinskiy2006}, pose \cite{Bickel2008} or expression~\cite{Shin2014,yang2020facescape} from high-quality 3D face meshes \cite{Bickel2008}. 
Fyffe et al.~\shortcite{Fyffe2014} use optical flow correspondence computed from dynamic video frames to animate static high-resolution scans. 
In contrast, \modelname learns an animatable detail model solely from in-the-wild images without paired 3D training data.
While FaceScape~\cite{yang2020facescape} predicts an animatable 3D face from a single image, the method is not robust to occlusions.
This is due to a two step reconstruction process: first optimize the coarse shape, then predict a displacement map from the texture map extracted with the coarse reconstruction.

Chaudhuri et al.~\shortcite{chaudhuri2020personalized} learn identity and expression corrective blendshapes with dynamic (expression-dependent) albedo maps \cite{Nagano2018}. 
They model geometric details as part of the albedo map, and therefore, the shading of these details does not adapt with varying lighting. 
This results in unrealistic renderings.
In contrast, \modelname models details as geometric displacements, which look natural when re-lit. 

In summary, \modelname occupies a unique space. 
It takes a single image as input and produces person-specific details that can be realistically animated. 
While some methods produce higher-frequency pixel-aligned details, these are not animatable.
Still other methods require high-resolution scans for training.
We show that these are not necessary and that animatable details can be learned from 2D images without paired 3D ground truth.
This is not just convenient, but means that \modelname learns to be robust to a wide variety of real-world variation.
We want to emphasize that, while elements of \modelname are built on well-understood principles (dating back to Vetter and Blanz), our core contribution is new and essential.
The key to making \modelname work is the detail consistency loss, which has not appeared previously in the literature.

\section{Preliminaries}
\label{FLAMEDECODER}

\qheading{Geometry prior:}
FLAME~\cite{FLAME2017} is a statistical 3D head model that combines separate linear identity shape and expression spaces with \ac{LBS} and pose-dependent corrective blendshapes to articulate the neck, jaw, and eyeballs.
Given parameters of facial identity $\shapecoeff \in \mathbb{R}^\shapedim$, pose $\posecoeff \in \mathbb{R}^{3\numjoints+3}$ (with $\numjoints=4$ joints for neck, jaw, and eyeballs), and expression $\expcoeff \in \mathbb{R}^\expdim$, FLAME outputs a mesh with $\numverts=5023$ vertices. 
The model is defined as
\begin{equation}
\flamev(\shapecoeff, \posecoeff, \expcoeff) = W(T_P(\shapecoeff, \posecoeff, \expcoeff),  \joints(\shapecoeff), \posecoeff, \blendweights),
\end{equation}
with the blend skinning function $W(\template, \joints, \posecoeff, \blendweights)$ that rotates the vertices in $\template \in \mathbb{R}^{3\numverts}$ around joints $\joints \in \mathbb{R}^{3\numjoints}$, linearly smoothed by blendweights $\blendweights \in \mathbb{R}^{\numjoints \times \numverts}$. The joint locations $\joints$ are defined as a function of the identity $\shapecoeff$. Further,
\begin{equation}
T_P(\shapecoeff, \posecoeff, \expcoeff) = \template + B_S(\shapecoeff; \shapespace) + B_P(\posecoeff; \posespace) + B_E(\expcoeff; \expspace)
\end{equation}
denotes the mean template $\template$ in ``zero pose'' with added shape blendshapes $B_S(\shapecoeff; \shapespace): \mathbb{R}^{\shapedim} \rightarrow \mathbb{R}^{3\numverts}$, pose correctives $B_P(\posecoeff; \posespace): \mathbb{R}^{3\numjoints+3} \rightarrow \mathbb{R}^{3\numverts}$, and expression blendshapes $B_E(\expcoeff; \expspace): \mathbb{R}^{\expdim} \rightarrow \mathbb{R}^{3\numverts}$, with the learned identity, pose, and expression bases (i.e.~linear subspaces) $\shapespace, \posespace$ and $\expspace$. See~\cite{FLAME2017} for details.

\qheading{Appearance model:}
FLAME does not have an appearance model, hence we convert the Basel Face Model's linear albedo subspace \cite{Paysan2009} into the FLAME UV layout to make it compatible with FLAME.
The appearance model outputs a UV albedo map $\albedo(\albedocoeffs) \in \mathbb{R}^{\uvsize \times \uvsize \times 3}$ for albedo parameters $\albedocoeffs \in \mathbb{R}^{\albedodim}$.

\qheading{Camera model:}
Photographs in existing in-the-wild face datasets are often taken from a distance.
We, therefore, use an orthographic camera model $\cam$ to project the 3D mesh into image space. 
Face vertices are projected into the image as $\vect{v} = s\Pi(\flamev_i)+\vect{t}$, where $\flamev_i \in \mathbb{R}^3$ is a vertex in $\flamev$, $\Pi \in \mathbb{R}^{2 \times 3}$ is the orthographic 3D-2D projection matrix, and $s \in \mathbb{R}$ and $\vect{t} \in \mathbb{R}^2$ denote isotropic scale and 2D translation, respectively.
The parameters $s$, and $\vect{t}$ are summarized as $\boldsymbol{c}$.

\qheading{Illumination model:}
For face reconstruction, the most frequently-employed illumination model %
is based on Spherical Harmonics (SH)~\cite{Ramamoorthi2001AnER}. 
By assuming that the light source is distant and the face's surface reflectance is Lambertian, the shaded face image is computed as:
\begin{equation}
    B(\albedocoeffs, \lighting, \normals_{uv})_{i,j} = \albedo(\albedocoeffs)_{i,j} \odot \sum_{k = 1}^{9}{\lighting_{k} H_{k}(\normals_{i,j})},
\end{equation}
where the albedo, $\albedo$, surface normals, $\normals$, and shaded texture, $B$, are represented in UV coordinates %
and where $B_{i,j} \in \mathbb{R}^3$, $\albedo_{i,j} \in \mathbb{R}^3$, and $\normals_{i,j} \in \mathbb{R}^3$ denote pixel $(i,j)$ in the UV coordinate system. %
The SH basis and coefficients are defined as $H_k: \mathbb{R}^3 \rightarrow \mathbb{R}$ and $\lighting = [\lighting_{1}^T, \cdots, \lighting_{9}^T]^T$, with $\lighting_{k} \in \mathbb{R}^3$, and $\odot$ denotes the Hadamard product. 

\qheading{Texture rendering:}
Given the geometry parameters ($\shapecoeff, \posecoeff, \expcoeff$), albedo ($\albedocoeffs$), lighting ($\lighting$) and camera information $\boldsymbol{c}$, we can generate the 2D image $\image_r$ by rendering as $\image_r = \mathcal{R}(\flamev, B, \cam)$, where $\mathcal{R}$ denotes the rendering function.
 
FLAME is able to generate the face geometry with various poses, shapes and expressions from a low-dimensional latent space. 
However, the representational power of the model is limited by the low mesh resolution and therefore mid-frequency details are mostly missing from FLAME's surface.
The next section introduces our expres\-sion-dependent displacement model that augments FLAME with mid-frequency details, and it demonstrates how to reconstruct this geometry from a single image and animate it.

\section{Method}

\begin{figure*}[t]
	\centering
	\includegraphics[width=0.98\textwidth, trim={0cm, 0cm, 0cm, 0cm}]{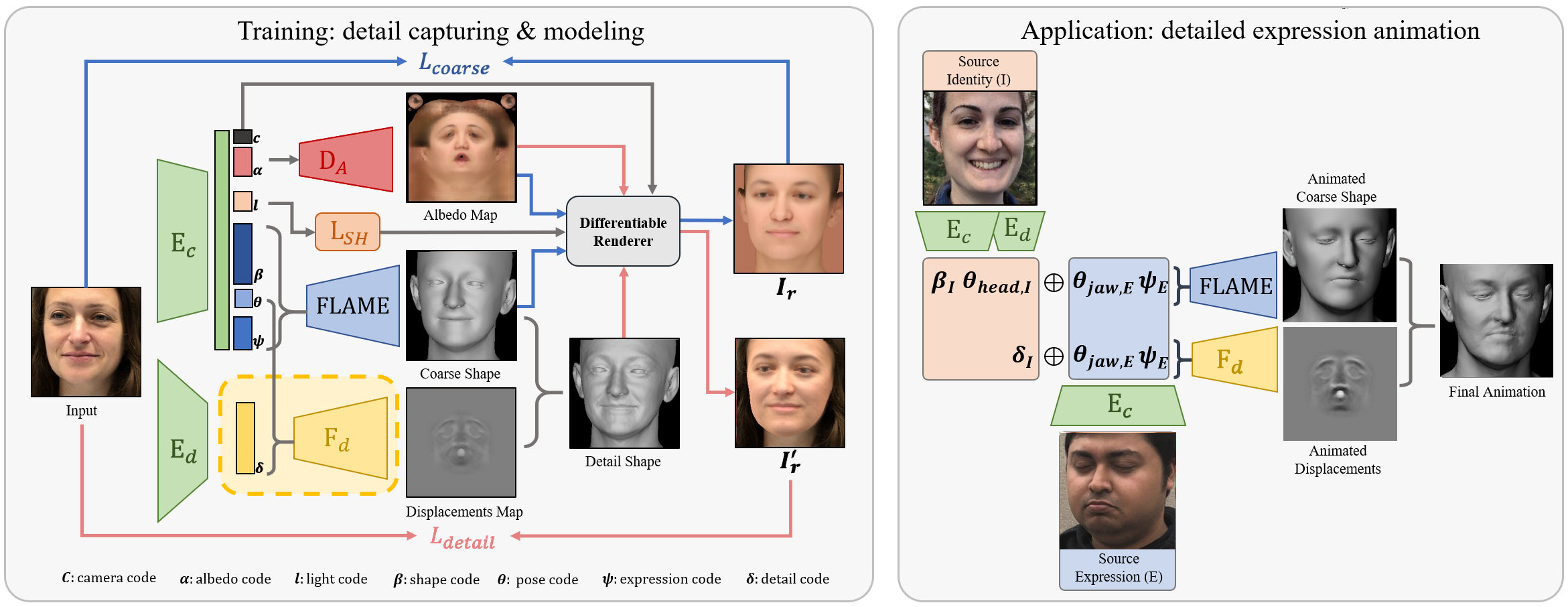} 
	\caption{\modelname training and animation. 
	During training (left box), \modelname estimates parameters to reconstruct face shape for each image with the aid of the shape consistency information (following the blue arrows) and, then, learns an expression-conditioned displacement model by leveraging detail consistency information (following the red arrows) from multiple images of the same individual (see Sec.~\ref{sec: disentanglement} for details). 
	While the analysis-by-synthesis pipeline is, by now, standard, the yellow box region contains our key novelty. This displacement consistency loss is further illustrated in Fig.~\ref{fig:detail_consistency}.
	Once trained, \modelname animates a face (right box) by combining the reconstructed source identity's shape, head pose, and detail code, with the reconstructed source expression's jaw pose and expression parameters to obtain an animated coarse shape and an animated displacement map.
	Finally, \modelname outputs an animated detail shape.
	Images are taken from NoW~\cite{Sanyal2019}.
	Note that NoW images are not used for training \modelname, but are just selected for illustration purposes.
}
	\label{fig:overview}
\end{figure*}

\begin{figure}[t]
	\centering
	\includegraphics[width=0.7\columnwidth, trim={2cm, 1cm, 2cm, 1cm}]{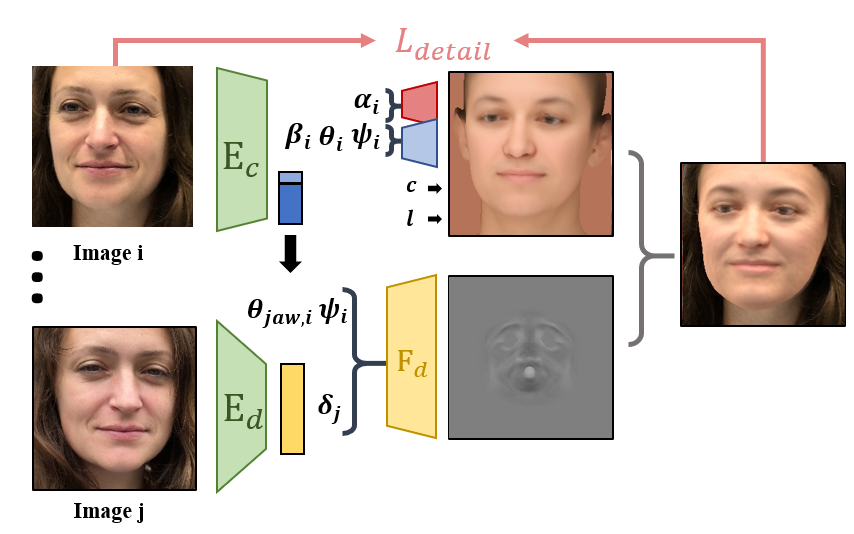}  
	\caption{Detail consistency loss. \modelname uses multiple images of the same person during training to disentangle static person-specific details from expression-dependent details. When properly factored, we should be able to take the detail code from one image of a person and use it to reconstruct another image of that person with a different expression.
	See Sec.~\ref{sec: disentanglement} for details.
	Images are taken from NoW~\cite{Sanyal2019}.
	Note that NoW images are not used for training, but are just selected for illustration purposes.
	}
	\label{fig:detail_consistency}
\end{figure}

\modelname learns to regress a parameterized face model with geometric detail solely from in-the-wild training images (Fig.~\ref{fig:overview} left).
Once trained, \modelname reconstructs the 3D head with detailed face geometry from a single face image, $\image$. 
The learned parametrization of the reconstructed details enables us to then animate the detail reconstruction by controlling FLAME's expression and jaw pose parameters (Fig.~\ref{fig:overview}, right).
This synthesizes new wrinkles while keeping person-specific details unchanged. 

\qheading{Key idea:}
The key idea of \modelname is grounded in the observation that an individual's face shows different details (i.e.~wrinkles), depending on their facial expressions but that other properties of their shape remain unchanged.
Consequently, facial details should be  separated into static person-specific details and dynamic expression-dependent details such as wrinkles~\cite{Li2009RobustSG}. 
However, disentangling static and dynamic facial details is a non-trivial task.
Static facial details are different across people, whereas dynamic expression dependent facial details even vary for the same person.
Thus, \modelname learns an expression-conditioned detail model to infer facial details from both the person-specific detail latent space and the expression space. 

The main difficulty in learning a detail displacement model is the lack of training data. 
Prior work uses specialized camera systems to scan people in a controlled environment to obtain detailed facial geometry. 
However, this approach is expensive and impractical for capturing large numbers of identities with varying expressions and diversity in ethnicity and age. 
Therefore we propose an approach to learn detail geometry from in-the-wild images.

\subsection{Coarse reconstruction}
We first learn a coarse reconstruction (i.e.~in FLAME's model space) in an analysis-by-synthesis way: given a 2D image $\image$ as input, we encode the image into a latent code, decode this to synthesize a 2D image $\image_r$, and minimize the difference between the synthesized image and the input.
As shown in Fig.~\ref{fig:overview}, we train an encoder $\coarseencoder$, which consists of a ResNet50~\cite{He2015DeepRL} network followed by a fully connected layer, to regress a low-dimensional latent code.
This latent code consists of FLAME parameters $\shapecoeff$, $\expcoeff$, $\posecoeff$ (i.e.~representing the coarse geometry), albedo coefficients $\albedocoeffs$, camera $\boldsymbol{c}$, and lighting parameters $\lighting$. 
More specifically, the coarse geometry uses the first 100 FLAME shape parameters ($\shapecoeff$), 50 expression parameters ($\expcoeff$), and 50 albedo parameters ($\albedocoeffs$). 
In total, $\coarseencoder$ predicts a $236$ dimensional latent code.

Given a dataset of $2D$ face images $\image_i$ with multiple images per subject, corresponding identity labels $c_i$, and $68$ $2D$ keypoints $\landmark_i$ per image, the coarse reconstruction branch is trained by minimizing
\begin{equation}
    L_{\mathit{coarse}} = L_{\mathit{lmk}} + L_{\mathit{eye}} + L_{\mathit{pho}} + L_{id} + L_{sc} + L_{\mathit{reg}},
\end{equation}
with landmark loss $L_{\mathit{lmk}}$, eye closure loss $L_{\mathit{eye}}$, photometric loss $L_{\mathit{pho}}$, identity loss $L_{id}$, shape consistency loss $L_{sc}$ and regularization $L_{\mathit{reg}}$.

\qheading{Landmark re-projection loss:}
The landmark loss measures the difference between ground-truth $2D$ face landmarks $\landmark_i$ and the corresponding landmarks on the FLAME model's surface $\flamev_i \in \mathbb{R}^3$, projected into the image by the estimated camera model. 
The landmark loss is defined as
\begin{equation}
    L_{\mathit{lmk}} = \sum \limits_{i=1}^{68} \norm{\landmark_i - s\Pi(\flamev_i)+\vect{t}}_1.
    \label{eq:landmark_loss}
\end{equation}

\qheading{Eye closure loss:}
The eye closure loss computes the relative offset of landmarks $\landmark_i$ and $\landmark_j$ on the upper and lower eyelid, and measures the difference to the offset of the corresponding landmarks on FLAME's surface $\flamev_i$ and $\flamev_j$ projected into the image.
Formally, the loss is given as
\begin{equation}
    L_{\mathit{eye}} = \sum \limits_{(i,j) \in E} \norm{\landmark_i - \landmark_j - s\Pi(\flamev_i - \flamev_j)}_1,
\end{equation}
where $E$ is the set of upper/lower eyelid landmark pairs.
While the landmark loss, $L_{\mathit{lmk}}$ (Eq. \ref{eq:landmark_loss}), penalizes the absolute landmark location differences, $L_{\mathit{eye}}$ penalizes the relative difference between eyelid landmarks.
Because the eye closure loss $L_{\mathit{eye}}$ is translation invariant, it is less susceptible to a misalignment between the projected 3D face and the image, compared to $L_{\mathit{lmk}}$. 
In contrast, simply increasing the landmark loss for the eye landmarks affects the overall face shape and can lead to unsatisfactory reconstructions.
See Fig.~\ref{fig:eye} for the effect of the eye-closure loss. 

\qheading{Photometric loss:}
The photometric loss computes the error between the input image $\image$ and the rendering $\image_{r}$ as
$$L_{\mathit{pho}} = \norm{V_{\image} \odot (\image - \image_{r})}_{1,1}.$$
Here, $V_{\image}$ is a face mask with value $1$ in the face skin region, and value $0$ elsewhere obtained by an existing face segmentation method~\cite{Nirkin2018}, and $\odot$ denotes the Hadamard product.
Computing the error in only the face region provides robustness to common occlusions by e.g.~hair, clothes, sunglasses, etc.
Without this, the predicted albedo will also consider the color of the occluder, which may be far from skin color, resulting in unnatural rendering (see Fig.~\ref{fig:eye}).

\qheading{Identity loss:}
Recent 3D face reconstruction methods demonstrate the effectiveness of utilizing an identity loss to produce more realistic face shapes~\cite{Deng2019,Gecer2019}. 
Motivated by this, we also use a pretrained face recognition network~\cite{Cao2018_VGGFace2}, to employ an identity loss during training. 

The face recognition network $f$ outputs feature embeddings of the rendered images and the input image, and the identity loss then measures the cosine similarity between the two embeddings.
Formally, the loss is defined as
\begin{equation}
    L_{id} = 1 - \frac{f(I) f(I_{r})}{\norm{f(I)}_2 \cdot \norm{f(I_{r})}_2}.
\end{equation}
By computing the error between embeddings, the loss encourages the rendered image to
capture fundamental properties of a person's identity, ensuring that the rendered image looks like the same person as the input subject.
Figure \ref{fig:eye} shows that the coarse shape results with $L_{id}$ look more like the input subject than those without.

\qheading{Shape consistency loss:}
Given two images $\image_i$ and $\image_j$ of the same subject (i.e.~$c_i = c_j$), the coarse encoder $\coarseencoder$ should output the same shape parameters (i.e.~$\shapecoeff_i = \shapecoeff_j$).
Previous work encourages shape consistency by enforcing the distance between $\shapecoeff_i$ and $\shapecoeff_j$ to be smaller by a margin than the distance to the shape coefficients corresponding to a different subject~\cite{Sanyal2019}. 
However, choosing this fixed margin is challenging in practice. 
Instead, we propose a different strategy by replacing $\shapecoeff_i$ with $\shapecoeff_j$ while keeping all other parameters unchanged.
Given that $\shapecoeff_i$ and $\shapecoeff_j$ represent the same subject, this new set of parameters must reconstruct $\image_i$ well. 
Formally, we minimize
\begin{equation}
    L_{sc} = L_{\mathit{coarse}}(I_i, \mathcal{R}(M(\shapecoeff_j, \posecoeff_i, \expcoeff_i), B(\albedocoeffs_i, \lighting_i, \normals_{uv,i}), \cam_i)).    
    \label{loss:sc}
\end{equation}
The goal is to make the rendered images look like the real person.
If the method has correctly estimated the shape of the face in two images of the same person, then swapping the shape parameters between these images should produce rendered images that are indistinguishable. 
Thus, we employ the photometric and identity loss on the rendered images from swapped shape parameters.

\qheading{Regularization:}
$L_{\mathit{reg}}$ regularizes shape $E_{\shapecoeff} = \norm{\shapecoeff}_2^2$, expression $E_{\expcoeff} = \norm{\expcoeff}_2^2$, and albedo $E_{\albedocoeffs} = \norm{\albedocoeffs}_2^2$.

\subsection{Detail reconstruction} 

The detail reconstruction augments the coarse FLAME geometry with a detailed UV displacement map $D \in [-0.01,0.01]^{\uvsize \times \uvsize}$ (see Fig.~\ref{fig:overview}).
Similar to the coarse reconstruction, we train an encoder $\detailencoder$ (with the same architecture as $\coarseencoder$) to encode $\image$ to a $128$-dimensional latent code $\zcode$, representing subject-specific details.
The latent code $\zcode$ is then concatenated with FLAME's expression $\expcoeff$ and jaw pose parameters $\posecoeff_{jaw}$, and decoded by $\edm$ to $D$. 

\qheading{Detail decoder: }
The detail decoder is defined as
\begin{equation}
    \displacements = \edm(\zcode, \expcoeff, \posecoeff_{jaw}),
\end{equation}
where the detail code $\zcode \in \mathbb{R}^{128}$ controls the static person-specific details.
We leverage the expression $\expcoeff \in \mathbb{R}^{50}$ and jaw pose parameters $\posecoeff_{jaw} \in \mathbb{R}^{3}$ from the coarse reconstruction branch to capture the dynamic expression wrinkle details.
For rendering, $\displacements$ is converted to a normal map. 

\qheading{Detail rendering:}
The detail displacement model allows us to generate images with mid-frequency surface details. 
To reconstruct the detailed geometry $\flamev^\prime$, we convert $\flamev$ and its surface normals $\normals$ to UV space, denoted as $\flamev_{uv} \in \mathbb{R}^{\uvsize \times \uvsize \times 3}$ and $\normals_{uv} \in \mathbb{R}^{\uvsize \times \uvsize \times 3}$, and combine them with $\displacements$ as
\begin{equation}
\flamev^\prime_{uv} = \flamev_{uv} + \displacements \odot N_{uv} .
\end{equation}
By calculating normals $\normals^\prime$ from $\flamev^\prime$, we obtain the detail rendering $\image_r^\prime$ by rendering $\flamev$ with the applied normal map as
\begin{equation}
    \image_r^\prime = \mathcal{R}(\flamev, B(\albedocoeffs, \lighting, \normals^\prime), \cam).
\end{equation}

The detail reconstruction is trained by minimizing
\begin{equation}
    L_{\mathit{detail}} =  L_{\mathit{phoD}} +L_{\mathit{mrf}} + L_{\mathit{sym}} + L_{dc} + L_{\mathit{regD}},
\end{equation}
with photometric detail loss $L_{\mathit{phoD}}$, ID-MRF loss $L_{\mathit{mrf}}$, soft symmetry loss $L_{\mathit{sym}}$, and detail regularization $L_{\mathit{regD}}$.
Since our estimated albedo is generated by a linear model with $50$ basis vectors, the rendered coarse face image only recovers low frequency information such as skin tone and basic facial attributes. 
High frequency details in the the rendered image result mainly from the displacement map, and hence, since $ L_{\mathit{detail}}$ compares the rendered detailed image with the real image, $\edm$ is forced to model detailed geometric information. 

\qheading{Detail photometric losses:}
With the applied detail displacement map, the rendered images $I_r^\prime$ contain some geometric details.
Equivalent to the coarse rendering, we use a photometric loss $L_{\mathit{phoD}} = \norm{V_{\image} \odot (I - I_r^\prime)}_{1,1}$, where, recall, $V_{\image}$ is a mask representing the visible skin pixels.

\qheading{ID-MRF loss:}
We adopt an Implicit Diversified Markov Random Field (ID-MRF) loss~\cite{Wang2018} to reconstruct geometric details. 
Given the input image and the detail rendering, the ID-MRF loss extracts feature patches from different layers of a pre-trained network, and then minimizes the difference between corresponding nearest neighbor feature patches from both images. 
Larsen et al.~\shortcite{Larsen2016AutoencodingBP} and Isola et al.~\shortcite{Isola2017ImagetoImageTW} point out that L1 losses are not able to recover the high frequency information in the data. 
Consequently, these two methods use a discriminator to obtain high-frequency detail. Unfortunately, this may result in an unstable adversarial training process.  
Instead, the ID-MRF loss regularizes the generated content to the original input at the local patch level; this encourages \modelname to capture high-frequency details.

Following Wang et al.~\shortcite{Wang2018}, the loss is computed on layers $conv3\_2$ and $conv4\_2$ of VGG19~\cite{Simonyan2014VeryDC} as
\begin{equation}
L_{\mathit{mrf}} = 2 L_M(conv4\_2) +  L_M(conv 3\_2),
\end{equation}
where $L_M(layer_{th})$ denotes the ID-MRF loss that is employed on the feature patches extracted from $\image_r^\prime$ and $\image$ with layer $layer_{th}$ of VGG19.
As with the photometric losses, we compute $L_{\mathit{mrf}}$ only for the face skin region in UV space.

\qheading{Soft symmetry loss:}
To add robustness to self-occlusions, we add a soft symmetry loss to regularize non-visible face parts.
Specifically, we minimize
\begin{equation}
 L_{sym} = \norm{V_{uv} \odot (\displacements - \mathit{flip}(\displacements))}_{1,1},
\end{equation}
where $V_{uv}$ denotes the face skin mask in UV space, and $\mathit{flip}$ is the horizontal flip operation. %
Without $L_{\mathit{sym}}$, for extreme poses, boundary artifacts become visible in occluded regions (Fig.~\ref{fig:abl}).

\qheading{Detail regularization:} 
The detail displacements are regularized by $L_{\mathit{regD}} = \norm{D}_{1,1}$ to reduce noise.

\subsection{Detail disentanglement} 
\label{sec: disentanglement}
Optimizing $L_{\mathit{detail}}$ enables us to reconstruct faces with mid-frequen\-cy details.
Making these detail reconstructions animatable, however, requires us to disentangle person specific details (i.e.~moles, pores, eyebrows, and expression-independent wrinkles) controlled by $\zcode$ from expression-dependent wrinkles (i.e.~wrinkles that change for varying facial expression) controlled by FLAME's expression and jaw pose parameters, $\expcoeff$ and $\posecoeff_{\mathit{jaw}}$.
Our key observation is that the same person in two images should have both similar coarse geometry {\em and} personalized details. 

Specifically, for the rendered detail image, {\em exchanging the detail codes between two images of the same subject should have no effect on the rendered image.}
This concept is illustrated in Fig.~\ref{fig:detail_consistency}.
Here we take the the jaw and expression parameters from image $i$, extract the detail code from image $j$, and combine these to estimate the wrinkle detail.
When we swap detail codes between different images of the same person, the produced results must remain realistic.

\qheading{Detail consistency loss:}
Given two images $\image_i$ and $\image_j$ of the same subject (i.e.~$c_i = c_j$), the loss is defined as
\begin{equation}
\begin{split}
L_{dc} = L_{\mathit{detail}}(I_i, \mathcal{R}(M(\shapecoeff_i, \posecoeff_i, \expcoeff_i), \albedo(\albedocoeffs_i), \\ \edm(\zcode_j, \expcoeff_i, \posecoeff_{\mathit{jaw}, i}), \lighting_i, \cam_i)),    
\end{split}
\label{loss:dc}
\end{equation}
where $\shapecoeff_i$, $\posecoeff_i$, $\expcoeff_i$, $\posecoeff_{\mathit{jaw}, i}$, $\albedocoeffs_i$, $\lighting_i$, and $\boldsymbol{c}_i$ are the parameters of $I_i$, while $\zcode_j$ is the detail code of $I_j$ (see Fig.~\ref{fig:detail_consistency}).
The detail consistency loss is essential for the disentanglement of identity-dependent and expression-dependent details.  
Without the detail consistency loss, the person-specific detail code, $\delta$, captures identity and expression dependent  details, and therefore, reconstructed details cannot be re-posed by varying the FLAME jaw pose and expression. 
We show the necessity and effectiveness of $L_{dc}$ in Sec.~\ref{sec:ablation}.

\section{Implementation Details}

\qheading{Data:}
We train \modelname on three publicly available datasets: \mbox{VGGFace2} \cite{Cao2018_VGGFace2}, BUPT-Balancedface~\cite{Wang_2019_ICCV} and VoxCeleb2 \cite{Chung18b}.
VGGFace2~\cite{Cao2018_VGGFace2} contains images of over $8k$ subjects, with an average of more than $350$ images per subject. 
BUPT-Balancedface~\cite{Wang_2019_ICCV} offers $7k$ subjects per ethnicity (i.e.~Caucasian, Indian, Asian and African), and VoxCeleb2~\cite{Chung18b} contains $145k$ videos of $6k$ subjects.
In total, \modelname is trained on 2 Million images. 

All datasets provide an identity label for each image.
We use FAN~\cite{Bulat2017} to predict $68$ 2D landmarks $\landmark_i$ on each face. 
To improve the robustness of the predicted landmarks, we run FAN for each image twice with different face crops, and discard all images with non-matching landmarks.
See \supmat~for details on data selection and data cleaning.

\qheading{Implementation details:}
\modelname is implemented in PyTorch~\cite{Paszke2019PyTorchAI}, using the differentiable rasterizer from Pytorch3D~\cite{ravi2020pytorch3d} for rendering.
We use Adam~\cite{Kingma2015AdamAM} as optimizer with a learning rate of $1e-4$.  
The input image size is $224^2$ and the UV space size is $\uvsize = 256$. 
See \supmat~for details.

\section{Evaluation}

\begin{figure*}[t]
	\centering
	\includegraphics[width=0.7\textwidth, trim={2cm, 1cm, 2cm, 1cm}]{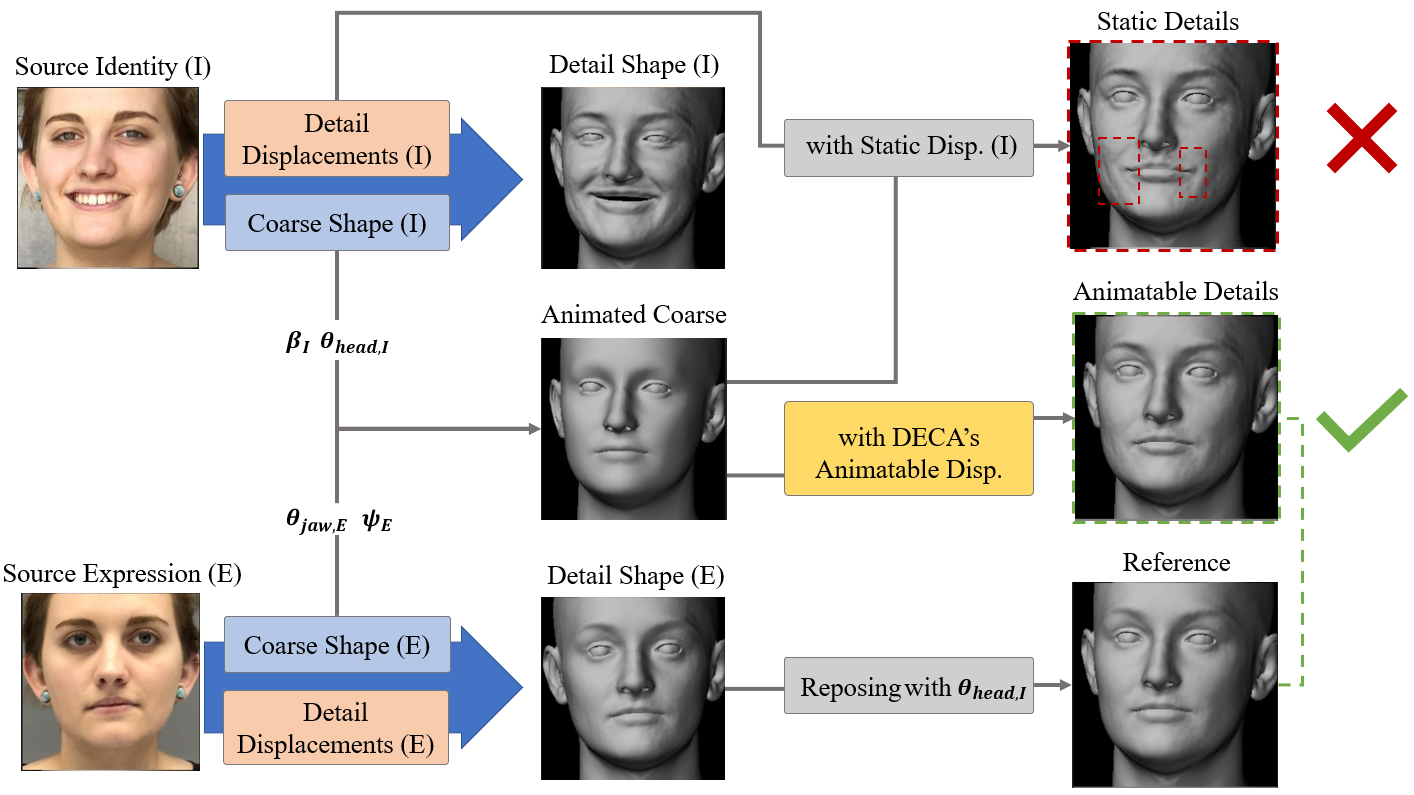}  
	\caption{Effect of \modelname's animatable details. Given images of source identity I and source expression E (left), \modelname reconstructs the detail shapes (middle) and animates the detail shape of I with the expression of E (right, middle). This synthesized \modelname expression appears nearly identical to the reconstructed same subject's reference detail shape (right, bottom). Using the reconstructed details of I instead (i.e.~static details) and animating the coarse shape only, results in visible artifacts (right, top). See Sec.~\ref{sec:qual_eval} for details.
	Input images are taken from NoW~\cite{Sanyal2019}.		
	}
	\label{fig:animation2}
\end{figure*}

\subsection{Qualitative evaluation}
\label{sec:qual_eval}

\qheading{Reconstruction:}
Given a single face image, \modelname reconstructs the 3D face shape with mid-frequency geometric details.
The second row of Fig.~\ref{fig:teaser} shows that the coarse shape (i.e.~in FLAME space) well represents the overall face shape, and the learned \modelname detail model reconstructs subject-specific details and wrinkles of the input identity (Fig.~\ref{fig:teaser}, row three), while being robust to partial occlusions.

\begin{figure}[t]
    \offinterlineskip
    \centering
    \includegraphics[width=0.142\columnwidth]{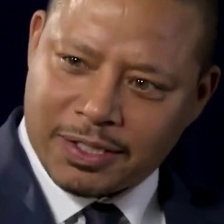}%
    \includegraphics[width=0.142\columnwidth]{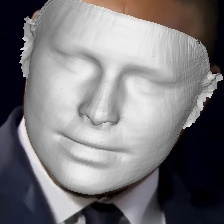}%
    \includegraphics[width=0.142\columnwidth]{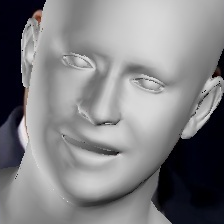}%
    \includegraphics[width=0.142\columnwidth]{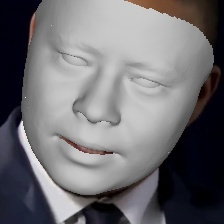}%
    \includegraphics[width=0.142\columnwidth]{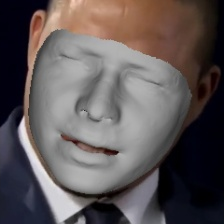}%
    \includegraphics[width=0.142\columnwidth]{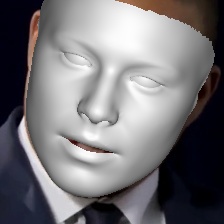}%
    \includegraphics[width=0.142\columnwidth]{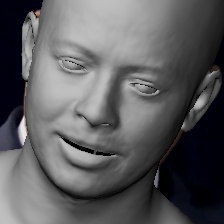}\\  
    \includegraphics[width=0.142\columnwidth]{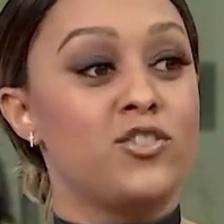}%
    \includegraphics[width=0.142\columnwidth]{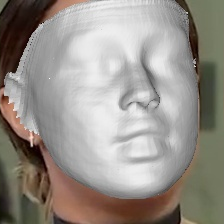}%
    \includegraphics[width=0.142\columnwidth]{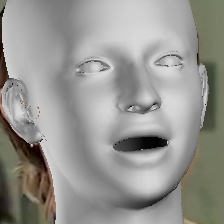}%
    \includegraphics[width=0.142\columnwidth]{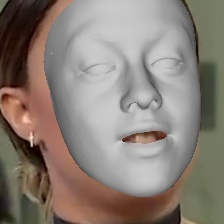}%
    \includegraphics[width=0.142\columnwidth]{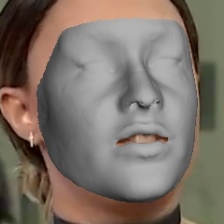}%
    \includegraphics[width=0.142\columnwidth]{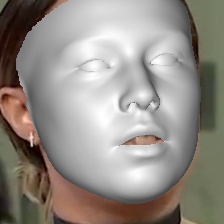}%
    \includegraphics[width=0.142\columnwidth]{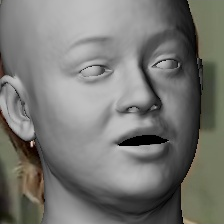}\\  
    \includegraphics[width=0.142\columnwidth]{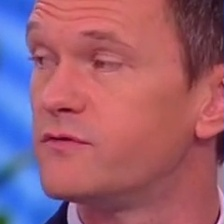}%
    \includegraphics[width=0.142\columnwidth]{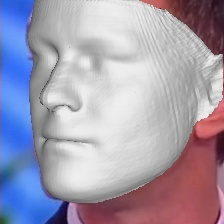}%
    \includegraphics[width=0.142\columnwidth]{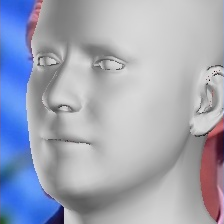}%
    \includegraphics[width=0.142\columnwidth]{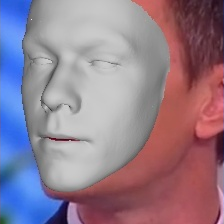}%
    \includegraphics[width=0.142\columnwidth]{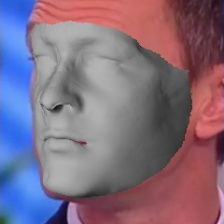}%
    \includegraphics[width=0.142\columnwidth]{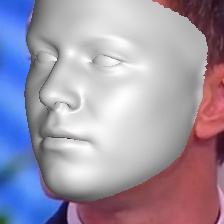}%
    \includegraphics[width=0.142\columnwidth]{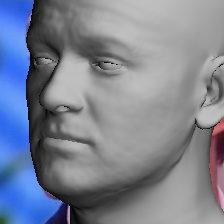}\\  
    \includegraphics[width=0.142\columnwidth]{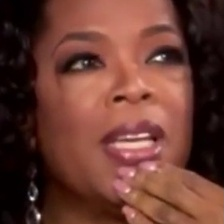}%
    \includegraphics[width=0.142\columnwidth]{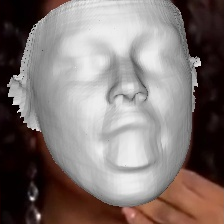}%
    \includegraphics[width=0.142\columnwidth]{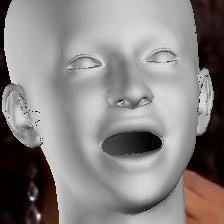}%
    \includegraphics[width=0.142\columnwidth]{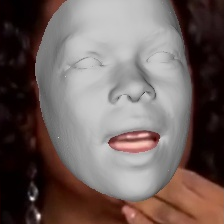}%
    \includegraphics[width=0.142\columnwidth]{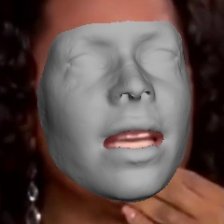}%
    \includegraphics[width=0.142\columnwidth]{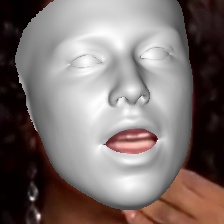}%
    \includegraphics[width=0.142\columnwidth]{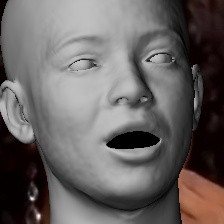}\\  
    \includegraphics[width=0.142\columnwidth]{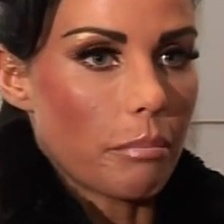}%
    \includegraphics[width=0.142\columnwidth]{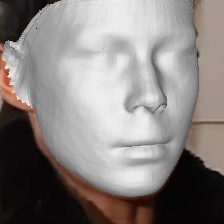}%
    \includegraphics[width=0.142\columnwidth]{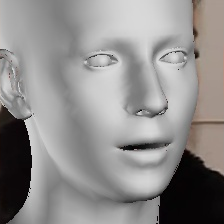}%
    \includegraphics[width=0.142\columnwidth]{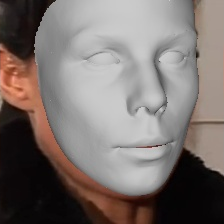}%
    \includegraphics[width=0.142\columnwidth]{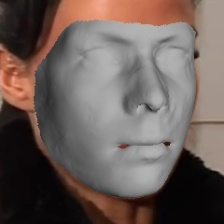}%
    \includegraphics[width=0.142\columnwidth]{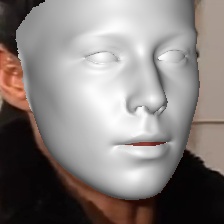}%
    \includegraphics[width=0.142\columnwidth]{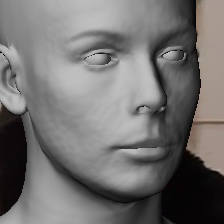}\\  
    \includegraphics[width=0.142\columnwidth]{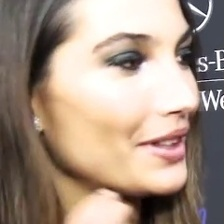}%
    \includegraphics[width=0.142\columnwidth]{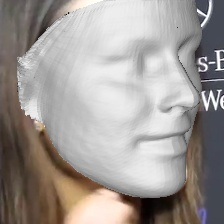}%
    \includegraphics[width=0.142\columnwidth]{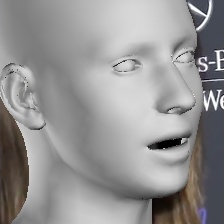}%
    \includegraphics[width=0.142\columnwidth]{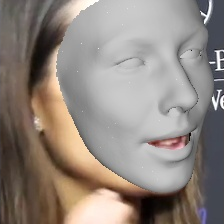}%
    \includegraphics[width=0.142\columnwidth]{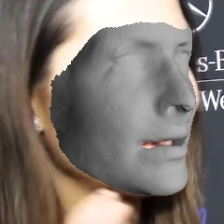}%
    \includegraphics[width=0.142\columnwidth]{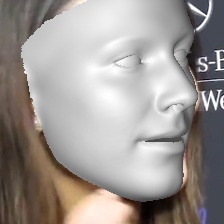}%
    \includegraphics[width=0.142\columnwidth]{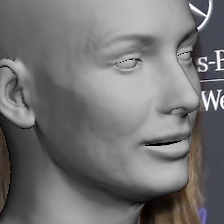}  
    \caption{Comparison to other {\bf coarse reconstruction} methods, from left to right: PRNet~\cite{Feng2018}, RingNet~\cite{Sanyal2019} Deng et al.~\shortcite{Deng2019}, FML~\cite{Tewari2019}, 3DDFA-V2~\cite{guo2020towards}, \modelname (ours).
	Input images are taken from VoxCeleb2~\cite{VoxCeleb2}.
    }
    \label{fig:qualitative1}
\end{figure}

Figure \ref{fig:qualitative1} qualitatively compares \modelname results with state-of-the-art coarse face reconstruction methods, namely PRNet~\cite{Feng2018}, RingNet~\cite{Sanyal2019}, Deng et al.~\shortcite{Deng2019}, FML~\cite{Tewari2019} and 3DDFA-V2~\cite{guo2020towards}.
Compared to these methods, \modelname better reconstructs the overall face shape with details like the nasolabial fold (rows 1, 2, 3, 4, and 6) and forehead wrinkles (row 3).
\modelname better reconstructs the mouth shape and the eye region than all other methods.
\modelname further reconstructs a full head while PRNet~\cite{Feng2018}, Deng et al.~\shortcite{Deng2019}, FML~\cite{Tewari2019} and 3DDFA-V2~\cite{guo2020towards} reconstruct tightly cropped faces. 
While RingNet~\cite{Sanyal2019}, like \modelname, is based on FLAME~\cite{FLAME2017}, \modelname better reconstructs the face shape and the facial expression. 

\begin{figure}[t]
    \offinterlineskip
        \includegraphics[width=0.99\columnwidth]{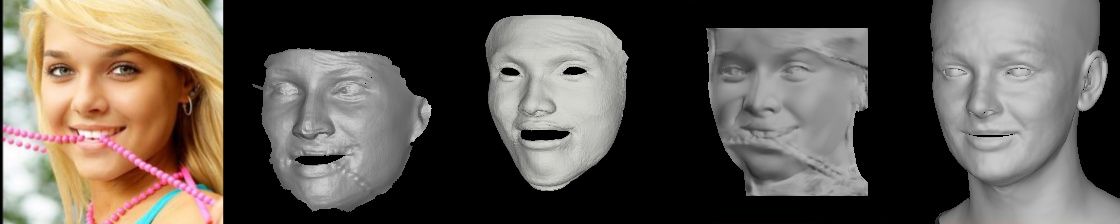}\\ 
        \includegraphics[width=0.99\columnwidth]{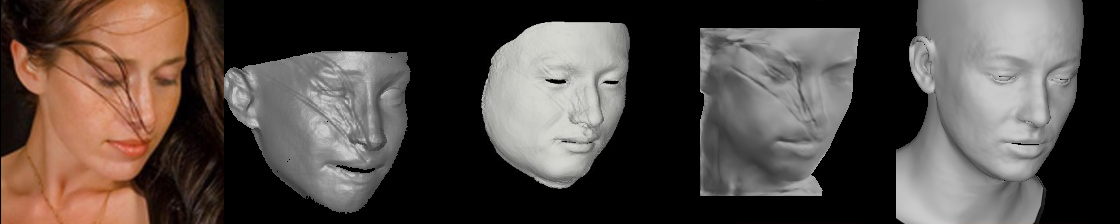}\\ 
        \includegraphics[width=0.99\columnwidth]{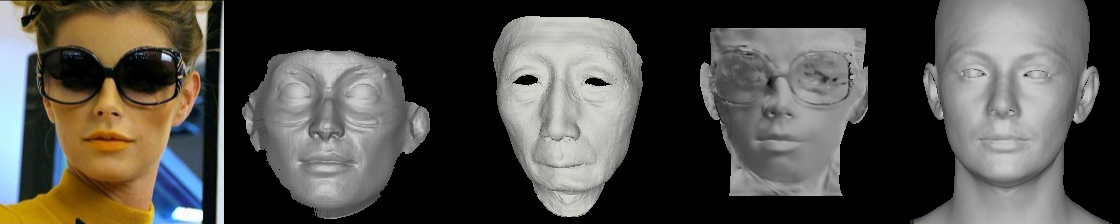}\\ 
        \includegraphics[width=0.99\columnwidth]{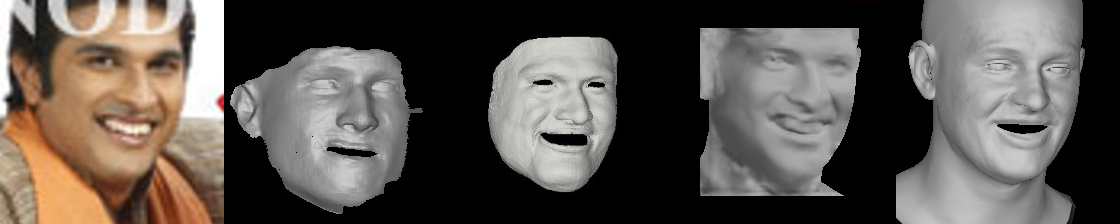}\\ 
        \includegraphics[width=0.99\columnwidth]{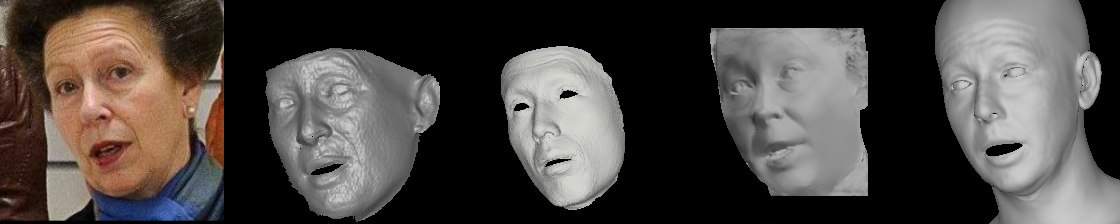}\\ 
        \includegraphics[width=0.99\columnwidth]{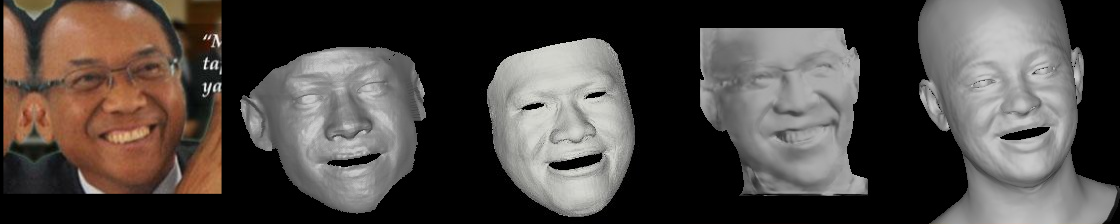} 
    \caption{Comparison to other detailed face reconstruction methods, from left to right: Extreme3D~\cite{AnhTran2018}, FaceScape~\cite{yang2020facescape}, Cross-modal~\cite{Abrevaya2020}, \modelname (ours). See Sup.~Mat.~for many more examples.
	Input images are taken from AFLW2000~\cite{Zhu2015} (rows 1-3) and VGGFace2~\cite{Cao2018_VGGFace2} (rows 4-6).
	}
    \label{fig:qualitative2}
\end{figure}

Figure~\ref{fig:qualitative2} compares \modelname visually to existing detailed face reconstruction methods, namely Extreme3D~\cite{AnhTran2018}, Cross-modal~\cite{Abrevaya2020}, and FaceScape~\cite{yang2020facescape}.
Extreme3D~\cite{AnhTran2018} and Cross-modal~\cite{Abrevaya2020} reconstruct more details than \modelname but at the cost of being less robust to occlusions (rows 1, 2, 3).
Unlike \modelname, Extreme3D and Cross-modal only reconstruct static details. 
However, using static details instead of \modelname's animatable details leads to visible artifacts when animating the face (see Fig.~\ref{fig:animation2}).
While FaceScape~\cite{yang2020facescape} provides animatable details, unlike \modelname, the method is trained on high-resolution scans while \modelname is solely trained on in-the-wild images.
Also, with occlusion, FaceScape produces artifacts (rows 1, 2) or effectively fails (row 3).

In summary, \modelname produces high-quality reconstructions, outperforming previous work in terms of robustness, while enabling animation of the detailed reconstruction.
To demonstrate the quality of \modelname and the robustness to variations in head pose, expression, occlusions, image resolution, lighting conditions, etc., we show results for 200 randomly selected ALFW2000~\cite{Zhu2015} images in the \supmat~along with more qualitative coarse and detail reconstruction comparisons to the state-of-the-art.

\begin{figure}[t]
	\centering
	\includegraphics[width=0.99\columnwidth]{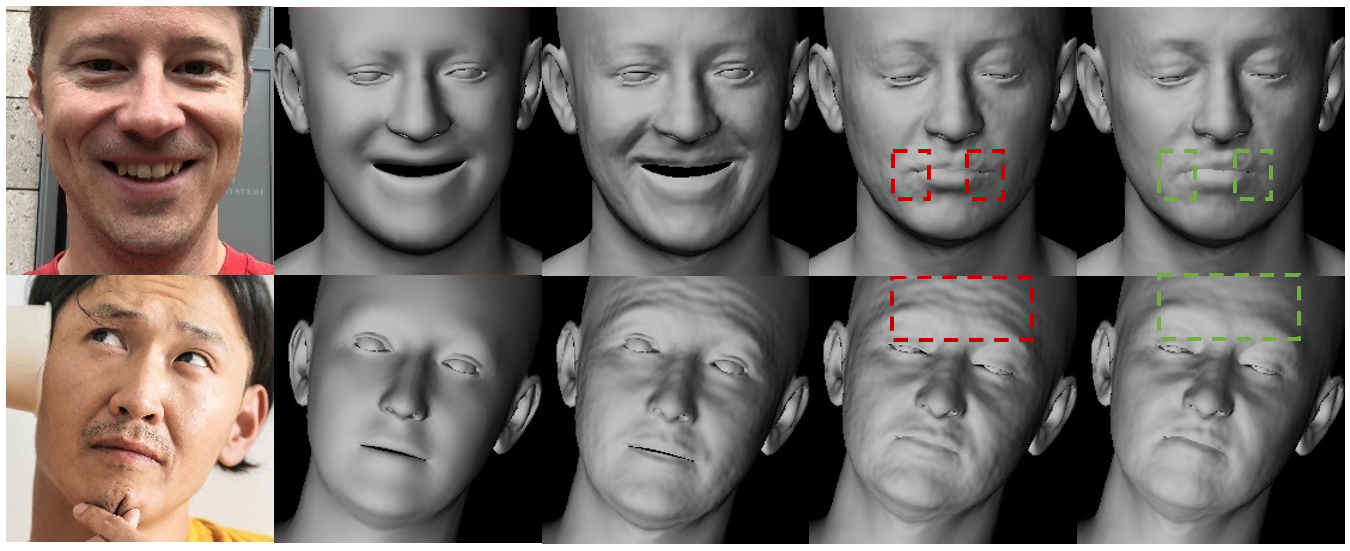}
	\caption{Effect of \modelname's animatable details. Given a single image (left), \modelname reconstructs a course mesh (second column) and a detailed mesh (third column). Using static details and animating (i.e.~reposing) the coarse FLAME shape only (fourth column) results in visible artifacts as highlighted by the red boxes. 
	Instead, reposing with \modelname's animatable details (right) results in a more realistic mesh with geometric details. 
	The reposing uses the source expression shown in Fig.~\ref{fig:teaser} (bottom).
	Input images are taken from NoW~\cite{Sanyal2019} (top), and Pexels~\shortcite{Pexels2021} (bottom).
	}
	\label{fig:animation1}
\end{figure}

\qheading{Detail animation:}
\modelname models detail displacements as a function of subject-specific detail parameters $\zcode$ and FLAME's jaw pose $\posecoeff_{\mathit{jaw}}$ and expression parameters $\expcoeff$ as illustrated in Fig.~\ref{fig:overview} (right).
This formulation allows us to animate detailed facial geometry such that wrinkles are specific to the source shape and expression as shown in Fig.~\ref{fig:teaser}.
Using static details instead of \modelname's animatable details (i.e.~by using the reconstructed details as a static displacement map) and animating only the coarse shape by changing the FLAME parameters results in visible artifacts as shown in Fig.~\ref{fig:animation2} (top), while animatable details (middle) look similar to the reference shape (bottom) of the same identity. 
Figure~\ref{fig:animation1} shows more examples where using static details results in artifacts at the mouth corner or the forehead region, while  \modelname's animated results look plausible.

\subsection{Quantitative evaluation}
\begin{figure*}[t]
    \begin{tabular}{c@{}c@{}c}
        \includegraphics[width=0.33\textwidth]{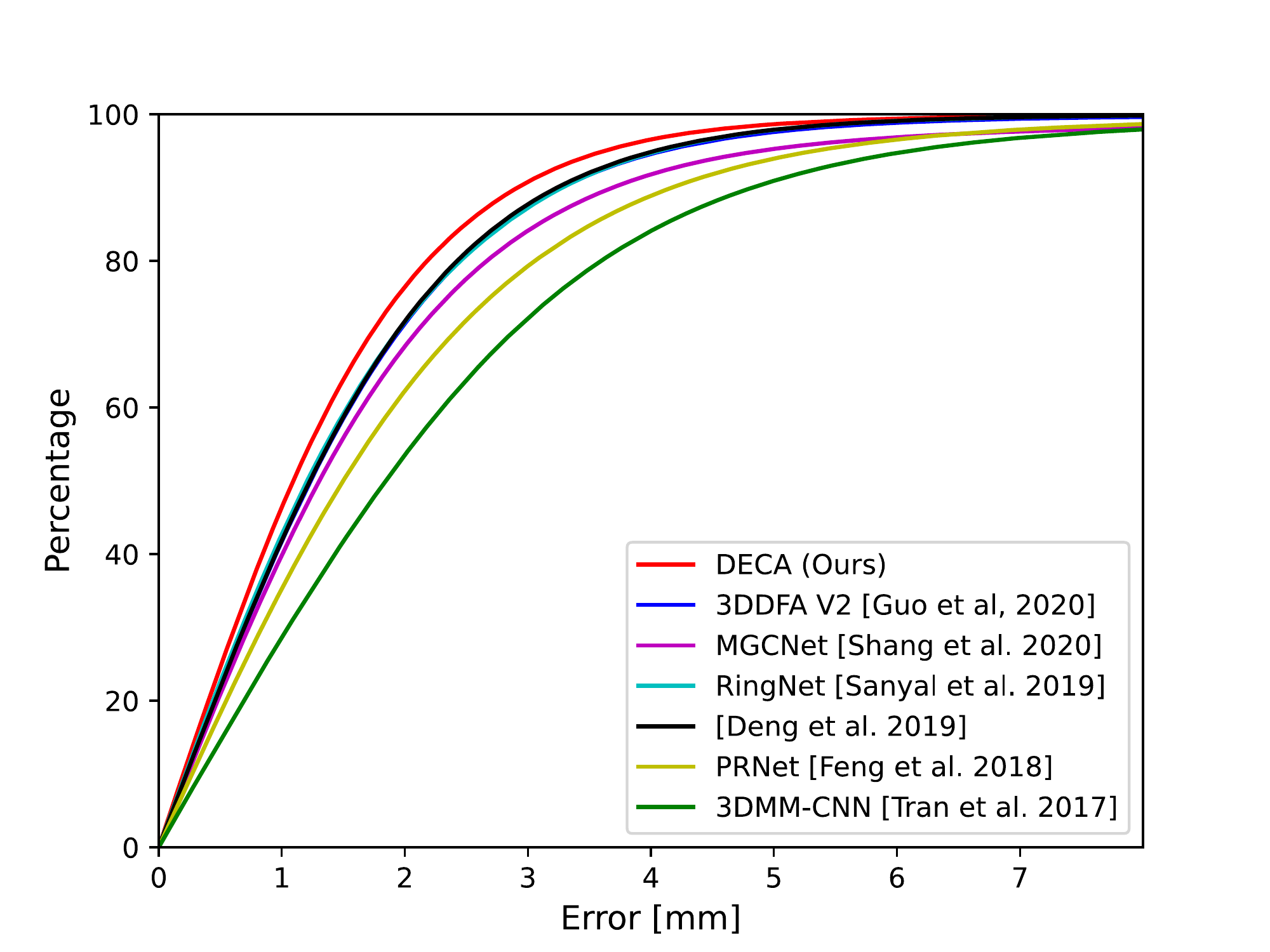} &  
        \includegraphics[width=0.33\textwidth]{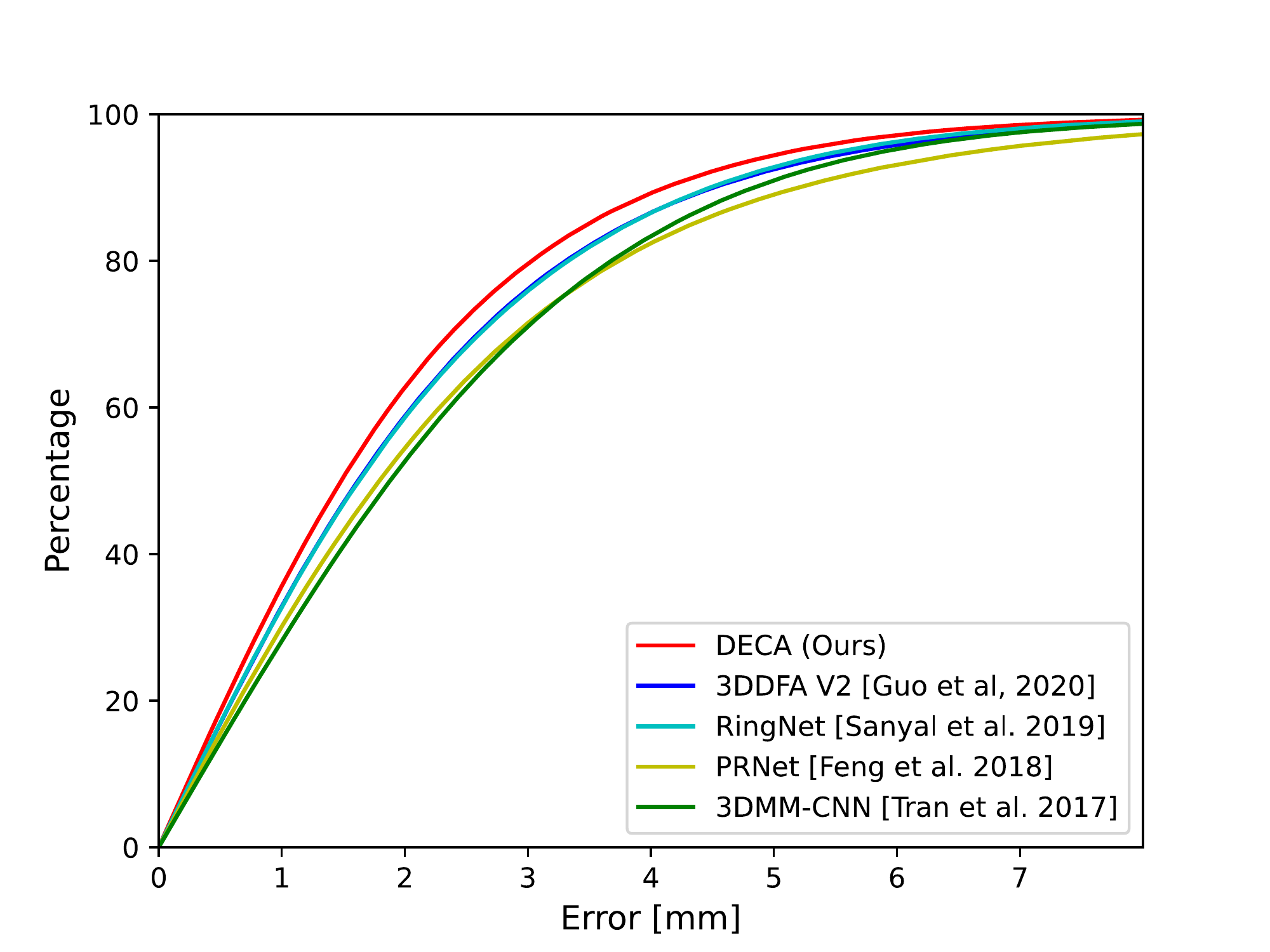} & 
        \includegraphics[width=0.33\textwidth]{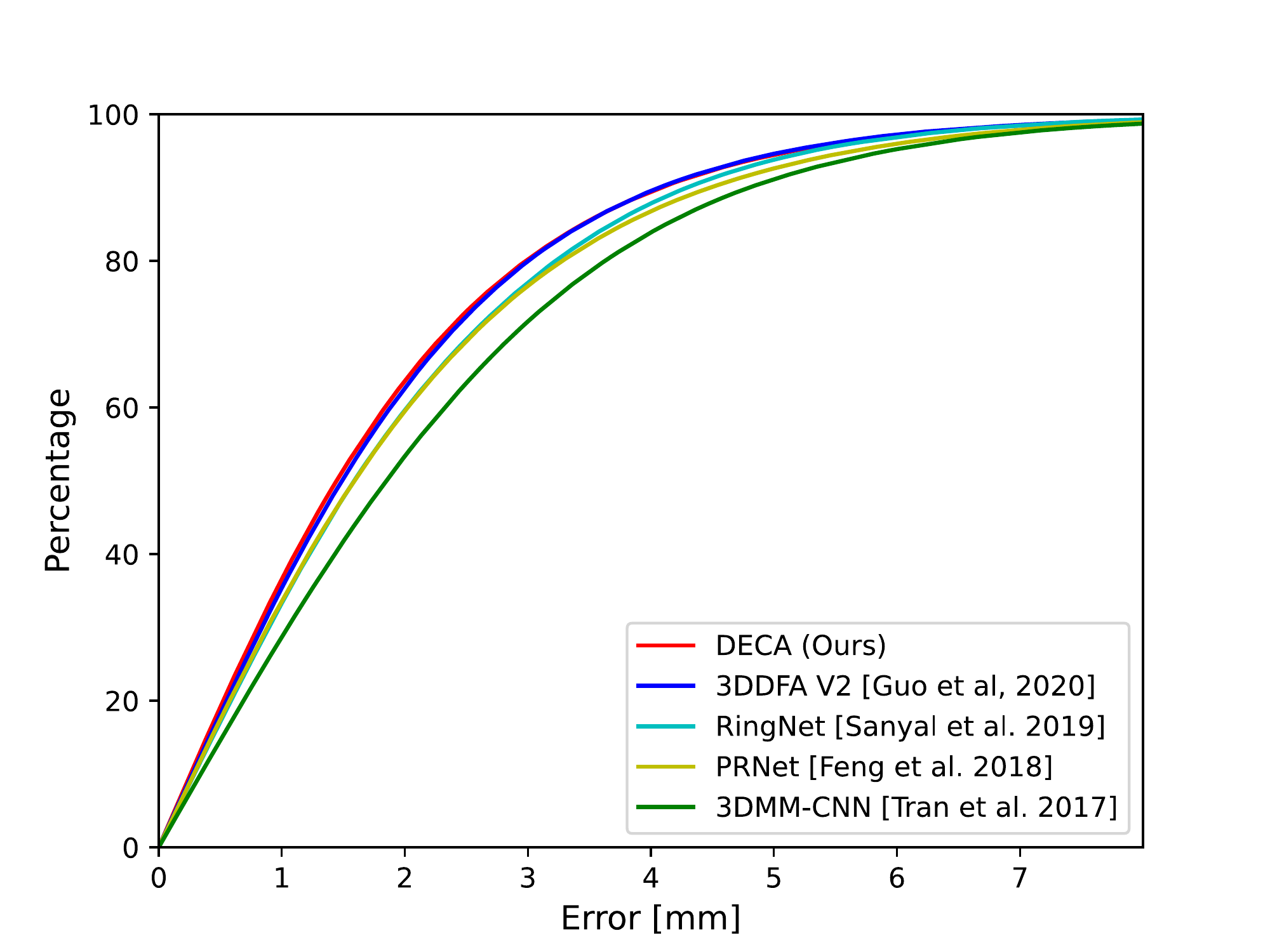} \\
         NoW~\cite{Sanyal2019} &
         Feng et al.~\shortcite{Feng2018evaluation} LQ &
         Feng et al.~\shortcite{Feng2018evaluation} HQ
    \end{tabular}
	\caption{Quantitative comparison to state-of-the-art on two 3D face reconstruction benchmarks, namely the NoW~\cite{Sanyal2019} challenge (left) and the Feng et al.~\shortcite{Feng2018evaluation} benchmark for low-quality (middle) and high-quality (right) images.}
	\label{fig:cumulative_error}
\end{figure*}
We compare \modelname with publicly available methods, namely 3DDFA-V2~\cite{guo2020towards}, Deng et al.~\shortcite{Deng2019}, RingNet~\cite{Sanyal2019}, PRNet~\cite{Feng2018}, 3DMM-CNN~\cite{AnhTran2017} and Extreme3D~\cite{AnhTran2018}.
Note that there is no benchmark face dataset with ground truth shape detail. 
Consequently, our quantitative analysis focuses on the accuracy of the coarse shape.
Note that \modelname achieves SOTA performance on 3D reconstruction without any paired 3D data in training.

\qheading{NoW benchmark:} 
The NoW challenge~\cite{Sanyal2019} consists of 2054 face images of 100 subjects, split into a validation set (20 subjects) and a test set (80 subjects), with a reference 3D face scan per subject. 
The images consist of indoor and outdoor images, neutral expression and expressive face images, partially occluded faces, and varying viewing angles ranging from frontal view to profile view, and selfie images.
The challenge provides a standard evaluation protocol that measures the distance from all reference scan vertices to the closest point in the reconstructed mesh surface, after rigidly aligning scans and reconstructions. 
For details, see~\cite{NoW2019}.

We found that the tightly cropped face meshes predicted by Deng et al.~\shortcite{Deng2019} are smaller than the NoW reference scans, which would result in a high reconstruction error in the missing region.
For a fair comparison to the method of Deng et al.~\shortcite{Deng2019}, we use the Basel Face Model (BFM)~\cite{Paysan2009} parameters they output, reconstruct the complete BFM mesh, and get the NoW evaluation for these complete meshes. 
As shown in Tab.~\ref{tab:NoW_table} and the cumulative error plot in Figure~\ref{fig:cumulative_error} (left), \modelname gives state-of-the-art results on NoW, providing the reconstruction error with the lowest mean, median, and standard deviation.

To quantify the influence of the geometric details, we separately evaluate the coarse and the detail shape (i.e. w/o and w/ details) on the NoW {\em validation set}. 
The reconstruction errors are, median: $1.18 / 1.1$9 (coarse / detailed), mean: $1.46 / 1.47$ (coarse / detailed), std: $1.25 / 1.25$ (coarse / detailed).
This indicates that while the detail shape improves visual quality when compared to the coarse shape, the quantitative performance is slightly worse. 

To test for gender bias in the results, we report errors separately for female (f) and male (m) NoW test subjects.
We find that recovered female shapes are slightly more accurate.
Reconstruction errors are, median: $1.03 / 1.16$ (f/m), mean: $1.32 / 1.45$ (f/m), and std: $1.16 / 1.20$ (f/m).
The cumulative error plots in Fig.~1 of the \supmat demonstrate that \modelname gives state-of-the-art performance for both genders. 

\begin{table}
    \centering
{\footnotesize 
	\begin{tabular}{l|c|c|c}
		\hline
		\multicolumn{1}{c|}{\textbf{Method}} & \multicolumn{1}{c|}{\textbf{\begin{tabular}[c]{@{}c@{}}Median (mm)\end{tabular}}} & \multicolumn{1}{c|}{\textbf{\begin{tabular}[c]{@{}c@{}}Mean (mm)\end{tabular}}} & \multicolumn{1}{c}{\textbf{\begin{tabular}[c]{@{}c@{}}Std (mm)\end{tabular}}} \\ 
		\hline
		3DMM-CNN~\cite{AnhTran2017}   & 1.84   & 2.33  & 2.05    \\ 
		PRNet~\cite{Feng2018}         & 1.50   & 1.98  & 1.88    \\ 
		Deng et al.19~\shortcite{Deng2019}      & 1.23   & 1.54  & 1.29    \\ 
		RingNet~\cite{Sanyal2019}  & 1.21   & 1.54  & 1.31    \\ 
		3DDFA-V2~\cite{guo2020towards}  & 1.23   & 1.57  & 1.39    \\ 
		MGCNet~\cite{Shang2020}  & 1.31 & 1.87  & 2.63    \\ 
		\modelname (ours)          & \textbf{1.09}   & \textbf{1.38}    & \textbf{1.18}  \\ 
		\hline
	\end{tabular}
}	
	\caption{Reconstruction error on the NoW~\cite{Sanyal2019} benchmark.}
    \label{tab:NoW_table}
\end{table}

\qheading{Feng et al. benchmark:}
The Feng et al.~\shortcite{Feng2018evaluation}~challenge contains 2000 face images of 135 subjects, and a reference 3D face scan for each subject. 
The benchmark consists of 1344 low-quality (LQ) images extracted from videos, and 656 high-quality (HQ) images taken in controlled scenarios.
A  protocol similar to NoW is used for evaluation, which measures the distance between all reference scan vertices to the closest points on the reconstructed mesh surface, after rigidly aligning scan and reconstruction. 
As shown in Tab.~\ref{tab:benchmarkdtirling} and the cumulative error plot in Fig.~\ref{fig:cumulative_error} (middle \& right), \modelname provides state-of-the-art performance.

\begin{table}[]
    \centering
{\footnotesize 
	\begin{tabular}{l|c|c|c|c|c|c}
		\hline
		\multicolumn{1}{c|}{\textbf{Method}} & 
		\multicolumn{2}{c|}{\textbf{\begin{tabular}[c]{@{}c@{}}Median (mm)\end{tabular}}} & \multicolumn{2}{c|}{\textbf{\begin{tabular}[c]{@{}c@{}}Mean (mm)\end{tabular}}} & \multicolumn{2}{c}{\textbf{\begin{tabular}[c]{@{}c@{}}Std (mm)\end{tabular}}} \\ \cline{2-7} 
		& \textbf{LQ}  & \textbf{HQ}  & \textbf{LQ}  & \textbf{HQ}  & \textbf{LQ}  & \textbf{HQ}  \\ \hline
		3DMM-CNN~\cite{AnhTran2017}  & 1.88  & 1.85  & 2.32   & 2.29   & 1.89  & 1.88  \\ 
		Extreme3D~\cite{AnhTran2018} & 2.40  & 2.37  & 3.49   & 3.58   & 6.15  & 6.75  \\ 
		PRNet~\cite{Feng2018}        & 1.79  & 1.59  & 2.38  & 2.06  & 2.19  & 1.79 \\ 
		RingNet~\cite{Sanyal2019}  & 1.63  & 1.59  & 2.08  & 2.02  & 1.79  & 1.69  \\ 
		3DDFA-V2~\cite{guo2020towards}  & 1.62  & 1.49  & 2.10  & 1.91  & 1.87  & \textbf{1.64}  \\ 
		\modelname (ours)          & \textbf{1.48}   & \textbf{1.45}    & \textbf{1.91}  & \textbf{1.89}   & \textbf{1.66}    & 1.68 \\ 
		\hline
	\end{tabular}
}
	\caption{Feng et al.~\shortcite{Feng2018evaluation} benchmark performance.}
	\label{tab:benchmarkdtirling}
\end{table}

\begin{figure}[t]
    \centerline{
		\includegraphics[width=0.75\columnwidth, trim={0cm, 1.0cm, 0cm, 0cm}]{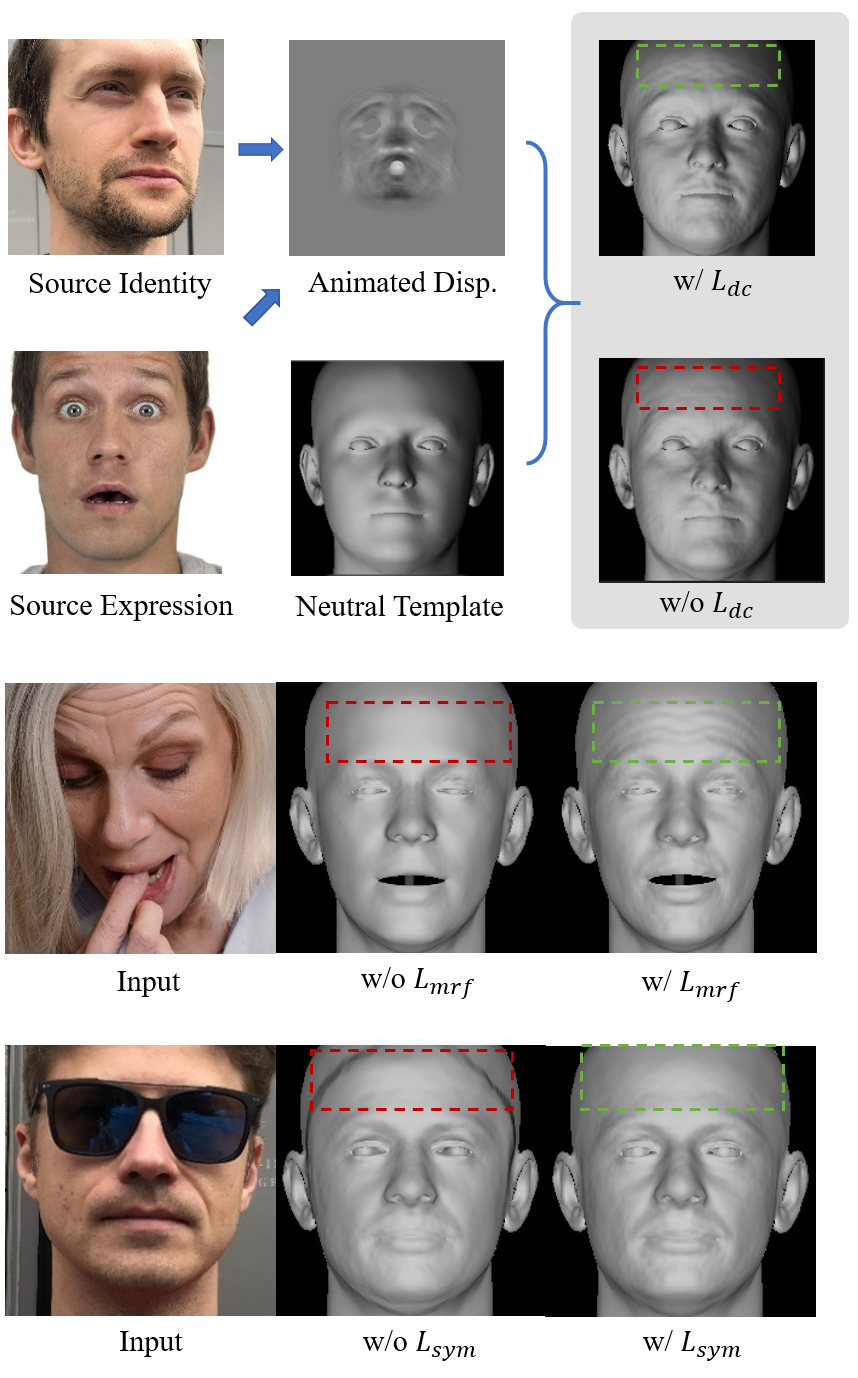}        
    }
    \caption{Ablation experiments. Top: Effects of $L_{dc}$ on the animation of the source identity with the source expression visualized on a neutral expression template mesh. Without $L_{dc}$, no wrinkles appear in the forehead despite the ``surprise" source expression. Middle: Effect of $L_{mrf}$ on the detail reconstruction. Without $L_{mrf}$, fewer details are reconstructed. 
    Bottom: Effect of $L_{sym}$ on the reconstructed details. Without $L_{sym}$, boundary artifacts become visible.
	Input images are taken from NoW~\cite{Sanyal2019} (rows 1 \& 4), Chicago~\cite{Chicago} (row 2), and Pexels~\shortcite{Pexels2021} (row 3).
    }
	\label{fig:abl}
\end{figure}

\begin{figure}[t]
    \offinterlineskip
		\includegraphics[width=0.54\columnwidth]{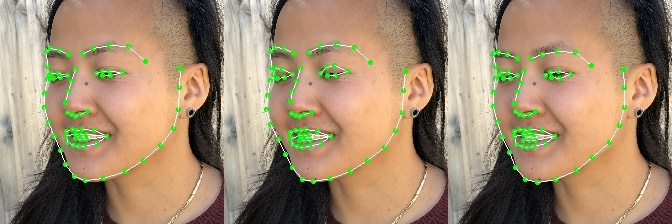}
		\includegraphics[width=0.36\columnwidth]{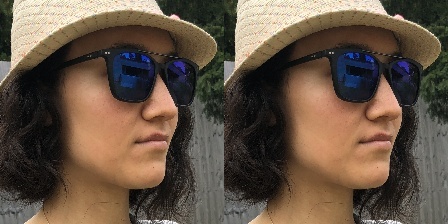}
        \\
		\includegraphics[width=0.54\columnwidth]{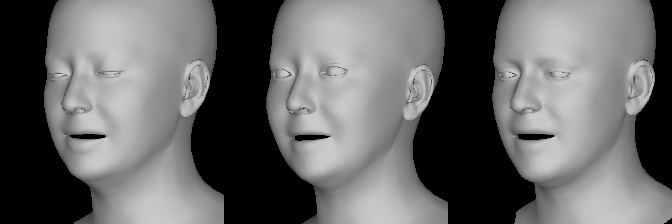}
		\includegraphics[width=0.36\columnwidth]{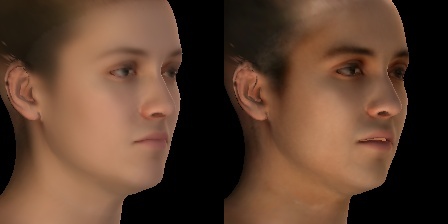}
    \caption{
    More ablation experiments. Left: estimated landmarks and reconstructed coarse shape from \modelname (first column) and \modelname without $L_{\mathit{eye}}$ (second column), and without $L_{id}$ (third column). 
    When trained without $L_{\mathit{eye}}$, \modelname is not able to capture closed-eye expressions. 
    Using $L_{id}$ helps reconstruct coarse shape. 
    Right: rendered image from \modelname and \modelname without segmentation. 
    Without using the skin mask in the photometric loss, the estimated result bakes in the color of the occluder (e.g.~sunglasses, hats) into the albedo.
    Input images are taken from NoW~\cite{Sanyal2019}.    
    }
	\label{fig:eye}
\end{figure}

\subsection{Ablation experiment}
\label{sec:ablation}

\qheading{Detail consistency loss:}
To evaluate the importance of our novel detail consistency loss $L_{dc}$ (Eq.~\ref{loss:dc}), we train \modelname with and without $L_{dc}$.
Figure~\ref{fig:abl} (left) shows the \modelname details for detail code $\zcode_I$ from the source identity, and expression $\expcoeff_E$ and jaw pose parameters $\posecoeff_{\mathit{jaw},E}$ from the source expression. 
For \modelname trained with $L_{dc}$ (top), wrinkles appear in the forehead as a result of the raised eyebrows of the source expression, while for \modelname trained without $L_{dc}$ (bottom), no such wrinkles appear. 
This indicates that without $L_{dc}$, person-specific details and expression-dependent wrinkles are not well disentangled. 
See \supmat~for more disentanglement results. 

\qheading{ID-MRF loss:}
Figure~\ref{fig:abl} (right) shows the effect of $L_{\mathit{mrf}}$ on the detail reconstruction. 
Without $L_{\mathit{mrf}}$ (middle), wrinkle details (e.g. in the forehead) are not reconstructed, resulting in an overly smooth result.
With $L_{\mathit{mrf}}$ (right), \modelname captures the details. 

\qheading{Other losses:}
We also evaluate the effect of the eye-closure loss $L_{eye}$, segmentation on the photometric loss, and the identity loss $L_{id}$. 
Fig.~\ref{fig:eye} provides a qualitative comparison of the \modelname coarse model with/without using these losses. 
Quantitatively, we also evaluate \modelname with and without $L_{id}$ on the NoW validation set; the former gives a mean error of 1.46mm, while the latter is worse with an error of 1.59mm. 

\section{Limitations and Future Work}
While \modelname achieves SOTA results for reconstructed face shape and provides novel animatable details, there are several limitations.
First, the rendering quality for \modelname detailed meshes is mainly limited by the albedo model, which is derived from BFM. 
\modelname requires an albedo space without baked in shading, specularities, and shadows in order to disentangle facial albedo from geometric details. 
Future work should focus on learning a high-quality albedo model with a sufficiently large variety of skin colors, texture details, and no illumination effects.
Second, existing methods, like \modelname, do not explicitly model facial hair.
This pushes skin tone into the lighting model and causes facial hair to be explained by shape deformations. 
A different approach is needed to properly model this.
Third, while robust, our method can still fail due to extreme head pose and lighting. 
While we are tolerant to common occlusions in existing face datasets (Fig.~\ref{fig:qualitative2} and examples in Sup.~Mat.), we do not address extreme occlusion, e.g.~where the hand covers large portions of the face.
This suggests the need for more diverse training data.

Further, the training set contains many low-res images, which help with robustness but can introduce noisy details. 
Existing high-res datasets (e.g.~\cite{Karras2018,Karras2019}) are less varied, thus training \modelname from these datasets results in a model that is less robust to general in-the-wild images, but captures more detail.
Additionally, the limited size of high-resolution datasets makes it difficult to disentangle expression- and identity-dependent details.
To further research on this topic, we also release a model trained using high-resolution images only (i.e.~\modelname-HR). 
Using \modelname-HR increases the visual quality and reduces noise in the reconstructed details at the cost of being less robust (i.e.~to low image resolutions, extreme head poses, extreme expressions, etc.). %

\modelname uses a weak perspective camera model.  
To use \modelname to recover head geometry from ``selfies'', we would need to extend the method to include the focal length.
For some applications, the focal length may be directly available from the camera.
However, inferring 3D geometry and focal length from a single image under perspective projection for in-the-wild images is unsolved and likely requires explicit supervision during training (cf. \cite{zhao2019learning}).

Finally, in future work, we want to extend the model over time, both for tracking and to learn more personalized models of individuals from video where we could enforce continuity of intrinsic wrinkles over time.

\section{Conclusion}

We have presented \modelname, which enables detailed expression capture and animation from single images by learning an animatable detail model from a dataset of in-the-wild images. 
In total, \modelname is trained from about 2M in-the-wild face images without 2D-to-3D supervision.
\modelname reaches state-of-the-art shape reconstruction performance enabled by a shape consistency loss. 
A novel detail consistency loss helps \modelname to disentangle expression-dependent wrinkles from person-specific details. 
The low-dimensional detail latent space makes the fine-scale reconstruction robust to noise and occlusions, and the novel loss leads to disentanglement of identity and expression-dependent wrinkle details. 
This enables applications like animation, shape change, wrinkle transfer, etc.
\modelname is publicly available for research purposes. 
Due to the reconstruction accuracy, the reliability, and the speed, \modelname is useful for applications like face reenactment or virtual avatar creation.

\begin{acks}
We thank S. Sanyal for providing us the RingNet PyTorch implementation, support with paper writing, and fruitful discussions, M. Kocabas, N. Athanasiou, V. Fernández Abrevaya, and R. Danecek for the helpful suggestions, and T. McConnell and S. Sorce for the video voice over. 
This work was partially supported by the Max Planck ETH Center for Learning Systems.

\qheading{Disclosure:}
MJB has received research gift funds from Intel, Nvidia, Adobe, Facebook, and Amazon. While MJB is a
part-time employee of Amazon, his research was performed solely at, and funded solely by, MPI. 
MJB has financial interests in Amazon, Datagen Technologies, and Meshcapade GmbH.
\end{acks}

%\balance
%\bibliographystyle{ACM-Reference-Format}
%\bibliography{bibliography}

\begin{thebibliography}{108}

%%% ====================================================================
%%% NOTE TO THE USER: you can override these defaults by providing
%%% customized versions of any of these macros before the \bibliography
%%% command.  Each of them MUST provide its own final punctuation,
%%% except for \shownote{}, \showDOI{}, and \showURL{}.  The latter two
%%% do not use final punctuation, in order to avoid confusing it with
%%% the Web address.
%%%
%%% To suppress output of a particular field, define its macro to expand
%%% to an empty string, or better, \unskip, like this:
%%%
%%% \newcommand{\showDOI}[1]{\unskip}   % LaTeX syntax
%%%
%%% \def \showDOI #1{\unskip}           % plain TeX syntax
%%%
%%% ====================================================================

\ifx \showCODEN    \undefined \def \showCODEN     #1{\unskip}     \fi
\ifx \showDOI      \undefined \def \showDOI       #1{#1}\fi
\ifx \showISBNx    \undefined \def \showISBNx     #1{\unskip}     \fi
\ifx \showISBNxiii \undefined \def \showISBNxiii  #1{\unskip}     \fi
\ifx \showISSN     \undefined \def \showISSN      #1{\unskip}     \fi
\ifx \showLCCN     \undefined \def \showLCCN      #1{\unskip}     \fi
\ifx \shownote     \undefined \def \shownote      #1{#1}          \fi
\ifx \showarticletitle \undefined \def \showarticletitle #1{#1}   \fi
\ifx \showURL      \undefined \def \showURL       {\relax}        \fi
% The following commands are used for tagged output and should be
% invisible to TeX
\providecommand\bibfield[2]{#2}
\providecommand\bibinfo[2]{#2}
\providecommand\natexlab[1]{#1}
\providecommand\showeprint[2][]{arXiv:#2}

\bibitem[\protect\citeauthoryear{Abrevaya, Boukhayma, Torr, and Boyer}{Abrevaya
  et~al\mbox{.}}{2020}]%
        {Abrevaya2020}
\bibfield{author}{\bibinfo{person}{Victoria~Fern{\'a}ndez Abrevaya},
  \bibinfo{person}{Adnane Boukhayma}, \bibinfo{person}{Philip~HS Torr}, {and}
  \bibinfo{person}{Edmond Boyer}.} \bibinfo{year}{2020}\natexlab{}.
\newblock \showarticletitle{Cross-modal Deep Face Normals with Deactivable Skip
  Connections}. In \bibinfo{booktitle}{\emph{IEEE Conference on Computer Vision
  and Pattern Recognition (CVPR)}}. \bibinfo{pages}{4979--4989}.
\newblock


\bibitem[\protect\citeauthoryear{Aldrian and Smith}{Aldrian and Smith}{2013}]%
        {AldrianSmith2013}
\bibfield{author}{\bibinfo{person}{Oswald Aldrian} {and}
  \bibinfo{person}{William~AP Smith}.} \bibinfo{year}{2013}\natexlab{}.
\newblock \showarticletitle{Inverse Rendering of Faces with a {3D} Morphable
  Model}.
\newblock \bibinfo{journal}{\emph{IEEE Transactions on Pattern Analysis and
  Machine Intelligence (PAMI)}} \bibinfo{volume}{35}, \bibinfo{number}{5}
  (\bibinfo{year}{2013}), \bibinfo{pages}{1080--1093}.
\newblock


\bibitem[\protect\citeauthoryear{Bas, Smith, Bolkart, and Wuhrer}{Bas
  et~al\mbox{.}}{2017}]%
        {Bas2017fitting}
\bibfield{author}{\bibinfo{person}{Anil Bas}, \bibinfo{person}{William A.~P.
  Smith}, \bibinfo{person}{Timo Bolkart}, {and} \bibinfo{person}{Stefanie
  Wuhrer}.} \bibinfo{year}{2017}\natexlab{}.
\newblock \showarticletitle{Fitting a {3D} Morphable Model to Edges: A
  Comparison Between Hard and Soft Correspondences}. In
  \bibinfo{booktitle}{\emph{Asian Conference on Computer Vision Workshops}}.
  \bibinfo{pages}{377--391}.
\newblock


\bibitem[\protect\citeauthoryear{Beeler, Bickel, Beardsley, Sumner, and
  Gross}{Beeler et~al\mbox{.}}{2010}]%
        {Beeler2010}
\bibfield{author}{\bibinfo{person}{Thabo Beeler}, \bibinfo{person}{Bernd
  Bickel}, \bibinfo{person}{Paul Beardsley}, \bibinfo{person}{Bob Sumner},
  {and} \bibinfo{person}{Markus Gross}.} \bibinfo{year}{2010}\natexlab{}.
\newblock \showarticletitle{High-quality single-shot capture of facial
  geometry}.
\newblock \bibinfo{journal}{\emph{ACM Transactions on Graphics (TOG)}}
  \bibinfo{volume}{29}, \bibinfo{number}{4} (\bibinfo{year}{2010}),
  \bibinfo{pages}{40}.
\newblock


\bibitem[\protect\citeauthoryear{Bickel, Lang, Botsch, Otaduy, and
  Gross}{Bickel et~al\mbox{.}}{2008}]%
        {Bickel2008}
\bibfield{author}{\bibinfo{person}{Bernd Bickel}, \bibinfo{person}{Manuel
  Lang}, \bibinfo{person}{Mario Botsch}, \bibinfo{person}{Miguel~A. Otaduy},
  {and} \bibinfo{person}{Markus~H. Gross}.} \bibinfo{year}{2008}\natexlab{}.
\newblock \showarticletitle{Pose-Space Animation and Transfer of Facial
  Details}. In \bibinfo{booktitle}{\emph{Eurographics/SIGGRAPH Symposium on
  Computer Animation (SCA)}}, \bibfield{editor}{\bibinfo{person}{Markus~H.
  Gross} {and} \bibinfo{person}{Doug~L. James}} (Eds.).
  \bibinfo{pages}{57--66}.
\newblock


\bibitem[\protect\citeauthoryear{Blanz, Romdhani, and Vetter}{Blanz
  et~al\mbox{.}}{2002}]%
        {Blanz2002}
\bibfield{author}{\bibinfo{person}{Volker Blanz}, \bibinfo{person}{Sami
  Romdhani}, {and} \bibinfo{person}{Thomas Vetter}.}
  \bibinfo{year}{2002}\natexlab{}.
\newblock \showarticletitle{Face identification across different poses and
  illuminations with a {3D} morphable model}. In
  \bibinfo{booktitle}{\emph{International Conference on Automatic Face \&
  Gesture Recognition (FG)}}. \bibinfo{pages}{202--207}.
\newblock


\bibitem[\protect\citeauthoryear{Blanz and Vetter}{Blanz and Vetter}{1999}]%
        {BlanzVetter1999}
\bibfield{author}{\bibinfo{person}{Volker Blanz} {and} \bibinfo{person}{Thomas
  Vetter}.} \bibinfo{year}{1999}\natexlab{}.
\newblock \showarticletitle{A morphable model for the synthesis of {3D} faces}.
  In \bibinfo{booktitle}{\emph{SIGGRAPH}}. \bibinfo{pages}{187--194}.
\newblock


\bibitem[\protect\citeauthoryear{Brunton, Salazar, Bolkart, and Wuhrer}{Brunton
  et~al\mbox{.}}{2014}]%
        {Brunton2014}
\bibfield{author}{\bibinfo{person}{Alan Brunton}, \bibinfo{person}{Augusto
  Salazar}, \bibinfo{person}{Timo Bolkart}, {and} \bibinfo{person}{Stefanie
  Wuhrer}.} \bibinfo{year}{2014}\natexlab{}.
\newblock \showarticletitle{Review of statistical shape spaces for {3D} data
  with comparative analysis for human faces}.
\newblock \bibinfo{journal}{\emph{Computer Vision and Image Understanding
  (CVIU)}} \bibinfo{volume}{128}, \bibinfo{number}{0} (\bibinfo{year}{2014}),
  \bibinfo{pages}{1--17}.
\newblock


\bibitem[\protect\citeauthoryear{Bulat and Tzimiropoulos}{Bulat and
  Tzimiropoulos}{2017}]%
        {Bulat2017}
\bibfield{author}{\bibinfo{person}{Adrian Bulat} {and}
  \bibinfo{person}{Georgios Tzimiropoulos}.} \bibinfo{year}{2017}\natexlab{}.
\newblock \showarticletitle{How far are we from solving the {2D} \& {3D} face
  alignment problem? (and a dataset of 230,000 {3D} facial landmarks)}. In
  \bibinfo{booktitle}{\emph{IEEE International Conference on Computer Vision
  (ICCV)}}. \bibinfo{pages}{1021--1030}.
\newblock


\bibitem[\protect\citeauthoryear{Cao, Bradley, Zhou, and Beeler}{Cao
  et~al\mbox{.}}{2015}]%
        {Cao2015}
\bibfield{author}{\bibinfo{person}{Chen Cao}, \bibinfo{person}{Derek Bradley},
  \bibinfo{person}{Kun Zhou}, {and} \bibinfo{person}{Thabo Beeler}.}
  \bibinfo{year}{2015}\natexlab{}.
\newblock \showarticletitle{Real-time high-fidelity facial performance
  capture}.
\newblock \bibinfo{journal}{\emph{ACM Transactions on Graphics (TOG)}}
  \bibinfo{volume}{34}, \bibinfo{number}{4} (\bibinfo{year}{2015}),
  \bibinfo{pages}{1--9}.
\newblock


\bibitem[\protect\citeauthoryear{Cao, Shen, Xie, Parkhi, and Zisserman}{Cao
  et~al\mbox{.}}{2018b}]%
        {Cao2018_VGGFace2}
\bibfield{author}{\bibinfo{person}{Qiong Cao}, \bibinfo{person}{Li Shen},
  \bibinfo{person}{Weidi Xie}, \bibinfo{person}{Omkar~M Parkhi}, {and}
  \bibinfo{person}{Andrew Zisserman}.} \bibinfo{year}{2018}\natexlab{b}.
\newblock \showarticletitle{{VGGFace2}: A dataset for recognising faces across
  pose and age}. In \bibinfo{booktitle}{\emph{International Conference on
  Automatic Face \& Gesture Recognition (FG)}}. \bibinfo{pages}{67--74}.
\newblock


\bibitem[\protect\citeauthoryear{Cao, Chen, Chen, Chen, Li, and Yu}{Cao
  et~al\mbox{.}}{2018a}]%
        {Cao2018}
\bibfield{author}{\bibinfo{person}{Xuan Cao}, \bibinfo{person}{Zhang Chen},
  \bibinfo{person}{Anpei Chen}, \bibinfo{person}{Xin Chen},
  \bibinfo{person}{Shiying Li}, {and} \bibinfo{person}{Jingyi Yu}.}
  \bibinfo{year}{2018}\natexlab{a}.
\newblock \showarticletitle{Sparse Photometric {3D} Face Reconstruction Guided
  by Morphable Models}. In \bibinfo{booktitle}{\emph{IEEE Conference on
  Computer Vision and Pattern Recognition (CVPR)}}.
  \bibinfo{pages}{4635--4644}.
\newblock


\bibitem[\protect\citeauthoryear{Chang, Tran, Hassner, Masi, Nevatia, and
  Medioni}{Chang et~al\mbox{.}}{2018}]%
        {Chang2018}
\bibfield{author}{\bibinfo{person}{Feng-Ju Chang}, \bibinfo{person}{Anh~Tuan
  Tran}, \bibinfo{person}{Tal Hassner}, \bibinfo{person}{Iacopo Masi},
  \bibinfo{person}{Ram Nevatia}, {and} \bibinfo{person}{Gerard Medioni}.}
  \bibinfo{year}{2018}\natexlab{}.
\newblock \showarticletitle{ExpNet: Landmark-free, deep, {3D} facial
  expressions}. In \bibinfo{booktitle}{\emph{International Conference on
  Automatic Face \& Gesture Recognition (FG)}}. \bibinfo{pages}{122--129}.
\newblock


\bibitem[\protect\citeauthoryear{Chaudhuri, Vesdapunt, Shapiro, and
  Wang}{Chaudhuri et~al\mbox{.}}{2020}]%
        {chaudhuri2020personalized}
\bibfield{author}{\bibinfo{person}{Bindita Chaudhuri},
  \bibinfo{person}{Noranart Vesdapunt}, \bibinfo{person}{Linda~G. Shapiro},
  {and} \bibinfo{person}{Baoyuan Wang}.} \bibinfo{year}{2020}\natexlab{}.
\newblock \showarticletitle{Personalized Face Modeling for Improved Face
  Reconstruction and Motion Retargeting}. In \bibinfo{booktitle}{\emph{European
  Conference on Computer Vision (ECCV)}}. \bibinfo{pages}{142--160}.
\newblock


\bibitem[\protect\citeauthoryear{Chen, Chen, Zhang, Mitchell, and Yu}{Chen
  et~al\mbox{.}}{2019}]%
        {Chen2019}
\bibfield{author}{\bibinfo{person}{Anpei Chen}, \bibinfo{person}{Zhang Chen},
  \bibinfo{person}{Guli Zhang}, \bibinfo{person}{Kenny Mitchell}, {and}
  \bibinfo{person}{Jingyi Yu}.} \bibinfo{year}{2019}\natexlab{}.
\newblock \showarticletitle{Photo-Realistic Facial Details Synthesis from
  Single Image}. In \bibinfo{booktitle}{\emph{IEEE International Conference on
  Computer Vision (ICCV)}}. \bibinfo{pages}{9429--9439}.
\newblock


\bibitem[\protect\citeauthoryear{Chung, Nagrani, and Zisserman}{Chung
  et~al\mbox{.}}{2018a}]%
        {Chung18b}
\bibfield{author}{\bibinfo{person}{J.~S. Chung}, \bibinfo{person}{A. Nagrani},
  {and} \bibinfo{person}{A. Zisserman}.} \bibinfo{year}{2018}\natexlab{a}.
\newblock \showarticletitle{VoxCeleb2: Deep Speaker Recognition}. In
  \bibinfo{booktitle}{\emph{INTERSPEECH}}.
\newblock


\bibitem[\protect\citeauthoryear{Chung, Nagrani, and Zisserman}{Chung
  et~al\mbox{.}}{2018b}]%
        {VoxCeleb2}
\bibfield{author}{\bibinfo{person}{Joon~Son Chung}, \bibinfo{person}{Arsha
  Nagrani}, {and} \bibinfo{person}{Andrew Zisserman}.}
  \bibinfo{year}{2018}\natexlab{b}.
\newblock \showarticletitle{VoxCeleb2: Deep Speaker Recognition}. In
  \bibinfo{booktitle}{\emph{Annual Conference of the International Speech
  Communication Association (Interspeech)}},
  \bibfield{editor}{\bibinfo{person}{B.~Yegnanarayana}} (Ed.).
  \bibinfo{publisher}{{ISCA}}, \bibinfo{pages}{1086--1090}.
\newblock


\bibitem[\protect\citeauthoryear{Cudeiro, Bolkart, Laidlaw, Ranjan, and
  Black}{Cudeiro et~al\mbox{.}}{2019}]%
        {VOCA2019}
\bibfield{author}{\bibinfo{person}{Daniel Cudeiro}, \bibinfo{person}{Timo
  Bolkart}, \bibinfo{person}{Cassidy Laidlaw}, \bibinfo{person}{Anurag Ranjan},
  {and} \bibinfo{person}{Michael Black}.} \bibinfo{year}{2019}\natexlab{}.
\newblock \showarticletitle{Capture, Learning, and Synthesis of {3D} Speaking
  Styles}. In \bibinfo{booktitle}{\emph{IEEE Conference on Computer Vision and
  Pattern Recognition (CVPR)}}. \bibinfo{pages}{10101--10111}.
\newblock


\bibitem[\protect\citeauthoryear{Deng, Yang, Xu, Chen, Jia, and Tong}{Deng
  et~al\mbox{.}}{2019}]%
        {Deng2019}
\bibfield{author}{\bibinfo{person}{Yu Deng}, \bibinfo{person}{Jiaolong Yang},
  \bibinfo{person}{Sicheng Xu}, \bibinfo{person}{Dong Chen},
  \bibinfo{person}{Yunde Jia}, {and} \bibinfo{person}{Xin Tong}.}
  \bibinfo{year}{2019}\natexlab{}.
\newblock \showarticletitle{Accurate {3D} Face Reconstruction with
  Weakly-Supervised Learning: From Single Image to Image Set}. In
  \bibinfo{booktitle}{\emph{Computer Vision and Pattern Recognition
  Workshops}}. \bibinfo{pages}{285--295}.
\newblock


\bibitem[\protect\citeauthoryear{Dou, Shah, and Kakadiaris}{Dou
  et~al\mbox{.}}{2017}]%
        {Dou2017}
\bibfield{author}{\bibinfo{person}{Pengfei Dou}, \bibinfo{person}{Shishir~K
  Shah}, {and} \bibinfo{person}{Ioannis~A Kakadiaris}.}
  \bibinfo{year}{2017}\natexlab{}.
\newblock \showarticletitle{End-to-end {3D} face reconstruction with deep
  neural networks}. In \bibinfo{booktitle}{\emph{IEEE Conference on Computer
  Vision and Pattern Recognition (CVPR)}}. \bibinfo{pages}{5908--5917}.
\newblock


\bibitem[\protect\citeauthoryear{Egger, Smith, Tewari, Wuhrer, Zollh{\"{o}}fer,
  Beeler, Bernard, Bolkart, Kortylewski, Romdhani, Theobalt, Blanz, and
  Vetter}{Egger et~al\mbox{.}}{2020}]%
        {Egger2020}
\bibfield{author}{\bibinfo{person}{Bernhard Egger}, \bibinfo{person}{William
  A.~P. Smith}, \bibinfo{person}{Ayush Tewari}, \bibinfo{person}{Stefanie
  Wuhrer}, \bibinfo{person}{Michael Zollh{\"{o}}fer}, \bibinfo{person}{Thabo
  Beeler}, \bibinfo{person}{Florian Bernard}, \bibinfo{person}{Timo Bolkart},
  \bibinfo{person}{Adam Kortylewski}, \bibinfo{person}{Sami Romdhani},
  \bibinfo{person}{Christian Theobalt}, \bibinfo{person}{Volker Blanz}, {and}
  \bibinfo{person}{Thomas Vetter}.} \bibinfo{year}{2020}\natexlab{}.
\newblock \showarticletitle{{3D} Morphable Face Models - Past, Present, and
  Future}.
\newblock \bibinfo{journal}{\emph{ACM Transactions on Graphics (TOG)}}
  \bibinfo{volume}{39}, \bibinfo{number}{5} (\bibinfo{year}{2020}),
  \bibinfo{pages}{157:1--157:38}.
\newblock


\bibitem[\protect\citeauthoryear{Feng, Wu, Shao, Wang, and Zhou}{Feng
  et~al\mbox{.}}{2018b}]%
        {Feng2018}
\bibfield{author}{\bibinfo{person}{Yao Feng}, \bibinfo{person}{Fan Wu},
  \bibinfo{person}{Xiaohu Shao}, \bibinfo{person}{Yanfeng Wang}, {and}
  \bibinfo{person}{Xi Zhou}.} \bibinfo{year}{2018}\natexlab{b}.
\newblock \showarticletitle{Joint {3D} Face Reconstruction and Dense Alignment
  with Position Map Regression Network}. In \bibinfo{booktitle}{\emph{European
  Conference on Computer Vision (ECCV)}}. \bibinfo{pages}{534--551}.
\newblock


\bibitem[\protect\citeauthoryear{Feng, Huber, Kittler, Hancock, Wu, Zhao,
  Koppen, and R{\"a}tsch}{Feng et~al\mbox{.}}{2018a}]%
        {Feng2018evaluation}
\bibfield{author}{\bibinfo{person}{Zhen-Hua Feng}, \bibinfo{person}{Patrik
  Huber}, \bibinfo{person}{Josef Kittler}, \bibinfo{person}{Peter Hancock},
  \bibinfo{person}{Xiao-Jun Wu}, \bibinfo{person}{Qijun Zhao},
  \bibinfo{person}{Paul Koppen}, {and} \bibinfo{person}{Matthias R{\"a}tsch}.}
  \bibinfo{year}{2018}\natexlab{a}.
\newblock \showarticletitle{Evaluation of dense {3D} reconstruction from {2D}
  face images in the wild}. In \bibinfo{booktitle}{\emph{International
  Conference on Automatic Face \& Gesture Recognition (FG)}}.
\newblock


\bibitem[\protect\citeauthoryear{{Flickr image}}{{Flickr image}}{2021}]%
        {FlickrImg}
\bibfield{author}{\bibinfo{person}{{Flickr image}}.}
  \bibinfo{year}{2021}\natexlab{}.
\newblock
  \bibinfo{howpublished}{\url{https://www.flickr.com/photos/gageskidmore/14602415448/}}.
\newblock


\bibitem[\protect\citeauthoryear{Fyffe, Jones, Alexander, Ichikari, and
  Debevec}{Fyffe et~al\mbox{.}}{2014}]%
        {Fyffe2014}
\bibfield{author}{\bibinfo{person}{Graham Fyffe}, \bibinfo{person}{Andrew
  Jones}, \bibinfo{person}{Oleg Alexander}, \bibinfo{person}{Ryosuke Ichikari},
  {and} \bibinfo{person}{Paul~E. Debevec}.} \bibinfo{year}{2014}\natexlab{}.
\newblock \showarticletitle{Driving High-Resolution Facial Scans with Video
  Performance Capture}.
\newblock \bibinfo{journal}{\emph{ACM Transactions on Graphics (TOG)}}
  \bibinfo{volume}{34}, \bibinfo{number}{1} (\bibinfo{year}{2014}),
  \bibinfo{pages}{8:1--8:14}.
\newblock


\bibitem[\protect\citeauthoryear{Garrido, Zollh{\"o}fer, Casas, Valgaerts,
  Varanasi, P{\'e}rez, and Theobalt}{Garrido et~al\mbox{.}}{2016}]%
        {Garrido2016}
\bibfield{author}{\bibinfo{person}{Pablo Garrido}, \bibinfo{person}{Michael
  Zollh{\"o}fer}, \bibinfo{person}{Dan Casas}, \bibinfo{person}{Levi
  Valgaerts}, \bibinfo{person}{Kiran Varanasi}, \bibinfo{person}{Patrick
  P{\'e}rez}, {and} \bibinfo{person}{Christian Theobalt}.}
  \bibinfo{year}{2016}\natexlab{}.
\newblock \showarticletitle{Reconstruction of personalized {3D} face rigs from
  monocular video}.
\newblock \bibinfo{journal}{\emph{ACM Transactions on Graphics (TOG)}}
  \bibinfo{volume}{35}, \bibinfo{number}{3} (\bibinfo{year}{2016}),
  \bibinfo{pages}{28}.
\newblock


\bibitem[\protect\citeauthoryear{Gecer, Ploumpis, Kotsia, and Zafeiriou}{Gecer
  et~al\mbox{.}}{2019}]%
        {Gecer2019}
\bibfield{author}{\bibinfo{person}{Baris Gecer}, \bibinfo{person}{Stylianos
  Ploumpis}, \bibinfo{person}{Irene Kotsia}, {and} \bibinfo{person}{Stefanos
  Zafeiriou}.} \bibinfo{year}{2019}\natexlab{}.
\newblock \showarticletitle{{GANFIT:} Generative Adversarial Network Fitting
  for High Fidelity {3D} Face Reconstruction}. In
  \bibinfo{booktitle}{\emph{IEEE Conference on Computer Vision and Pattern
  Recognition (CVPR)}}. \bibinfo{pages}{1155--1164}.
\newblock


\bibitem[\protect\citeauthoryear{Genova, Cole, Maschinot, Sarna, Vlasic, and
  Freeman}{Genova et~al\mbox{.}}{2018}]%
        {Genova2018}
\bibfield{author}{\bibinfo{person}{Kyle Genova}, \bibinfo{person}{Forrester
  Cole}, \bibinfo{person}{Aaron Maschinot}, \bibinfo{person}{Aaron Sarna},
  \bibinfo{person}{Daniel Vlasic}, {and} \bibinfo{person}{William~T. Freeman}.}
  \bibinfo{year}{2018}\natexlab{}.
\newblock \showarticletitle{Unsupervised Training for {3D} Morphable Model
  Regression}. In \bibinfo{booktitle}{\emph{IEEE Conference on Computer Vision
  and Pattern Recognition (CVPR)}}. \bibinfo{pages}{8377--8386}.
\newblock


\bibitem[\protect\citeauthoryear{Gerig, Morel-Forster, Blumer, Egger, Luthi,
  Sch{\"o}nborn, and Vetter}{Gerig et~al\mbox{.}}{2018}]%
        {Gerig2018}
\bibfield{author}{\bibinfo{person}{Thomas Gerig}, \bibinfo{person}{Andreas
  Morel-Forster}, \bibinfo{person}{Clemens Blumer}, \bibinfo{person}{Bernhard
  Egger}, \bibinfo{person}{Marcel Luthi}, \bibinfo{person}{Sandro
  Sch{\"o}nborn}, {and} \bibinfo{person}{Thomas Vetter}.}
  \bibinfo{year}{2018}\natexlab{}.
\newblock \showarticletitle{Morphable face models-an open framework}. In
  \bibinfo{booktitle}{\emph{International Conference on Automatic Face \&
  Gesture Recognition (FG)}}. \bibinfo{pages}{75--82}.
\newblock


\bibitem[\protect\citeauthoryear{Ghosh, Gupta, Uziel, Ranjan, Black, and
  Bolkart}{Ghosh et~al\mbox{.}}{2020}]%
        {Ghosh2020}
\bibfield{author}{\bibinfo{person}{Partha Ghosh}, \bibinfo{person}{Pravir~Singh
  Gupta}, \bibinfo{person}{Roy Uziel}, \bibinfo{person}{Anurag Ranjan},
  \bibinfo{person}{Michael~J. Black}, {and} \bibinfo{person}{Timo Bolkart}.}
  \bibinfo{year}{2020}\natexlab{}.
\newblock \showarticletitle{{GIF:} Generative Interpretable Faces}. In
  \bibinfo{booktitle}{\emph{International Conference on 3D Vision (3DV)}}.
  \bibinfo{pages}{868--878}.
\newblock


\bibitem[\protect\citeauthoryear{Golovinskiy, Matusik, Pfister, Rusinkiewicz,
  and Funkhouser}{Golovinskiy et~al\mbox{.}}{2006}]%
        {Golovinskiy2006}
\bibfield{author}{\bibinfo{person}{Aleksey Golovinskiy},
  \bibinfo{person}{Wojciech Matusik}, \bibinfo{person}{Hanspeter Pfister},
  \bibinfo{person}{Szymon Rusinkiewicz}, {and} \bibinfo{person}{Thomas~A.
  Funkhouser}.} \bibinfo{year}{2006}\natexlab{}.
\newblock \showarticletitle{A statistical model for synthesis of detailed
  facial geometry}.
\newblock \bibinfo{journal}{\emph{ACM Transactions on Graphics (TOG)}}
  \bibinfo{volume}{25}, \bibinfo{number}{3} (\bibinfo{year}{2006}),
  \bibinfo{pages}{1025--1034}.
\newblock


\bibitem[\protect\citeauthoryear{G{\"u}ler, Trigeorgis, Antonakos, Snape,
  Zafeiriou, and Kokkinos}{G{\"u}ler et~al\mbox{.}}{2017}]%
        {Guler2017}
\bibfield{author}{\bibinfo{person}{Riza~Alp G{\"u}ler}, \bibinfo{person}{George
  Trigeorgis}, \bibinfo{person}{Epameinondas Antonakos},
  \bibinfo{person}{Patrick Snape}, \bibinfo{person}{Stefanos Zafeiriou}, {and}
  \bibinfo{person}{Iasonas Kokkinos}.} \bibinfo{year}{2017}\natexlab{}.
\newblock \showarticletitle{{DenseReg}: Fully convolutional dense shape
  regression in-the-wild}. In \bibinfo{booktitle}{\emph{IEEE Conference on
  Computer Vision and Pattern Recognition (CVPR)}}.
  \bibinfo{pages}{6799--6808}.
\newblock


\bibitem[\protect\citeauthoryear{Guo, Zhu, Yang, Yang, Lei, and Li}{Guo
  et~al\mbox{.}}{2020}]%
        {guo2020towards}
\bibfield{author}{\bibinfo{person}{Jianzhu Guo}, \bibinfo{person}{Xiangyu Zhu},
  \bibinfo{person}{Yang Yang}, \bibinfo{person}{Fan Yang},
  \bibinfo{person}{Zhen Lei}, {and} \bibinfo{person}{Stan~Z Li}.}
  \bibinfo{year}{2020}\natexlab{}.
\newblock \showarticletitle{Towards Fast, Accurate and Stable {3D} Dense Face
  Alignment}. In \bibinfo{booktitle}{\emph{European Conference on Computer
  Vision (ECCV)}}. \bibinfo{pages}{152--168}.
\newblock


\bibitem[\protect\citeauthoryear{Guo, Cai, Jiang, Zheng, et~al\mbox{.}}{Guo
  et~al\mbox{.}}{2018}]%
        {Guo2018}
\bibfield{author}{\bibinfo{person}{Yudong Guo}, \bibinfo{person}{Jianfei Cai},
  \bibinfo{person}{Boyi Jiang}, \bibinfo{person}{Jianmin Zheng},
  {et~al\mbox{.}}} \bibinfo{year}{2018}\natexlab{}.
\newblock \showarticletitle{{CNN}-based real-time dense face reconstruction
  with inverse-rendered photo-realistic face images}.
\newblock \bibinfo{journal}{\emph{IEEE Transactions on Pattern Analysis and
  Machine Intelligence (PAMI)}} \bibinfo{volume}{41}, \bibinfo{number}{6}
  (\bibinfo{year}{2018}), \bibinfo{pages}{1294--1307}.
\newblock


\bibitem[\protect\citeauthoryear{He, Zhang, Ren, and Sun}{He
  et~al\mbox{.}}{2016}]%
        {He2015DeepRL}
\bibfield{author}{\bibinfo{person}{Kaiming He}, \bibinfo{person}{Xiangyu
  Zhang}, \bibinfo{person}{Shaoqing Ren}, {and} \bibinfo{person}{Jian Sun}.}
  \bibinfo{year}{2016}\natexlab{}.
\newblock \showarticletitle{Deep Residual Learning for Image Recognition}. In
  \bibinfo{booktitle}{\emph{IEEE Conference on Computer Vision and Pattern
  Recognition (CVPR)}}. \bibinfo{pages}{770--778}.
\newblock


\bibitem[\protect\citeauthoryear{Hu, Saito, Wei, Nagano, Seo, Fursund, Sadeghi,
  Sun, Chen, and Li}{Hu et~al\mbox{.}}{2017}]%
        {Hu2017}
\bibfield{author}{\bibinfo{person}{Liwen Hu}, \bibinfo{person}{Shunsuke Saito},
  \bibinfo{person}{Lingyu Wei}, \bibinfo{person}{Koki Nagano},
  \bibinfo{person}{Jaewoo Seo}, \bibinfo{person}{Jens Fursund},
  \bibinfo{person}{Iman Sadeghi}, \bibinfo{person}{Carrie Sun},
  \bibinfo{person}{Yen-Chun Chen}, {and} \bibinfo{person}{Hao Li}.}
  \bibinfo{year}{2017}\natexlab{}.
\newblock \showarticletitle{Avatar Digitization from a Single Image for
  Real-time Rendering}.
\newblock \bibinfo{journal}{\emph{ACM Transactions on Graphics (TOG)}}
  \bibinfo{volume}{36}, \bibinfo{number}{6} (\bibinfo{year}{2017}),
  \bibinfo{pages}{195:1--195:14}.
\newblock


\bibitem[\protect\citeauthoryear{Ichim, Bouaziz, and Pauly}{Ichim
  et~al\mbox{.}}{2015}]%
        {Ichim2015}
\bibfield{author}{\bibinfo{person}{Alexandru~Eugen Ichim},
  \bibinfo{person}{Sofien Bouaziz}, {and} \bibinfo{person}{Mark Pauly}.}
  \bibinfo{year}{2015}\natexlab{}.
\newblock \showarticletitle{Dynamic {3D} avatar creation from hand-held video
  input}.
\newblock \bibinfo{journal}{\emph{ACM Transactions on Graphics (TOG)}}
  \bibinfo{volume}{34}, \bibinfo{number}{4} (\bibinfo{year}{2015}),
  \bibinfo{pages}{45}.
\newblock


\bibitem[\protect\citeauthoryear{Isola, Zhu, Zhou, and Efros}{Isola
  et~al\mbox{.}}{2017}]%
        {Isola2017ImagetoImageTW}
\bibfield{author}{\bibinfo{person}{Phillip Isola}, \bibinfo{person}{Jun-Yan
  Zhu}, \bibinfo{person}{Tinghui Zhou}, {and} \bibinfo{person}{Alexei~A.
  Efros}.} \bibinfo{year}{2017}\natexlab{}.
\newblock \showarticletitle{Image-to-Image Translation with Conditional
  Adversarial Networks}. In \bibinfo{booktitle}{\emph{IEEE Conference on
  Computer Vision and Pattern Recognition (CVPR)}}.
  \bibinfo{pages}{5967--5976}.
\newblock


\bibitem[\protect\citeauthoryear{Jackson, Bulat, Argyriou, and
  Tzimiropoulos}{Jackson et~al\mbox{.}}{2017}]%
        {Jackson2017}
\bibfield{author}{\bibinfo{person}{Aaron~S Jackson}, \bibinfo{person}{Adrian
  Bulat}, \bibinfo{person}{Vasileios Argyriou}, {and} \bibinfo{person}{Georgios
  Tzimiropoulos}.} \bibinfo{year}{2017}\natexlab{}.
\newblock \showarticletitle{Large pose {3D} face reconstruction from a single
  image via direct volumetric {CNN} regression}. In
  \bibinfo{booktitle}{\emph{IEEE International Conference on Computer Vision
  (ICCV)}}. \bibinfo{pages}{1031--1039}.
\newblock


\bibitem[\protect\citeauthoryear{Jeni, Cohn, and Kanade}{Jeni
  et~al\mbox{.}}{2015}]%
        {Jeni2015}
\bibfield{author}{\bibinfo{person}{L{\'a}szl{\'o}~A Jeni},
  \bibinfo{person}{Jeffrey~F Cohn}, {and} \bibinfo{person}{Takeo Kanade}.}
  \bibinfo{year}{2015}\natexlab{}.
\newblock \showarticletitle{Dense {3D} face alignment from {2D} videos in
  real-time}. In \bibinfo{booktitle}{\emph{International Conference on
  Automatic Face \& Gesture Recognition (FG)}}, Vol.~\bibinfo{volume}{1}.
  \bibinfo{pages}{1--8}.
\newblock


\bibitem[\protect\citeauthoryear{Jiang, Zhang, Deng, Li, and Liu}{Jiang
  et~al\mbox{.}}{2018}]%
        {Jiang2018}
\bibfield{author}{\bibinfo{person}{Luo Jiang}, \bibinfo{person}{Juyong Zhang},
  \bibinfo{person}{Bailin Deng}, \bibinfo{person}{Hao Li}, {and}
  \bibinfo{person}{Ligang Liu}.} \bibinfo{year}{2018}\natexlab{}.
\newblock \showarticletitle{{3D} face reconstruction with geometry details from
  a single image}.
\newblock \bibinfo{journal}{\emph{Transactions on Image Processing}}
  \bibinfo{volume}{27}, \bibinfo{number}{10} (\bibinfo{year}{2018}),
  \bibinfo{pages}{4756--4770}.
\newblock


\bibitem[\protect\citeauthoryear{Karras, Aila, Laine, Herva, and
  Lehtinen}{Karras et~al\mbox{.}}{2017}]%
        {Karras2017}
\bibfield{author}{\bibinfo{person}{Tero Karras}, \bibinfo{person}{Timo Aila},
  \bibinfo{person}{Samuli Laine}, \bibinfo{person}{Antti Herva}, {and}
  \bibinfo{person}{Jaakko Lehtinen}.} \bibinfo{year}{2017}\natexlab{}.
\newblock \showarticletitle{Audio-driven facial animation by joint end-to-end
  learning of pose and emotion}.
\newblock \bibinfo{journal}{\emph{ACM Transactions on Graphics, (Proc.
  SIGGRAPH)}} \bibinfo{volume}{36}, \bibinfo{number}{4} (\bibinfo{year}{2017}),
  \bibinfo{pages}{94:1--94:12}.
\newblock


\bibitem[\protect\citeauthoryear{Karras, Aila, Laine, and Lehtinen}{Karras
  et~al\mbox{.}}{2018}]%
        {Karras2018}
\bibfield{author}{\bibinfo{person}{Tero Karras}, \bibinfo{person}{Timo Aila},
  \bibinfo{person}{Samuli Laine}, {and} \bibinfo{person}{Jaakko Lehtinen}.}
  \bibinfo{year}{2018}\natexlab{}.
\newblock \showarticletitle{Progressive growing of {GANs} for improved quality,
  stability, and variation}. In \bibinfo{booktitle}{\emph{International
  Conference on Learning Representations (ICLR)}}.
\newblock


\bibitem[\protect\citeauthoryear{Karras, Laine, and Aila}{Karras
  et~al\mbox{.}}{2019}]%
        {Karras2019}
\bibfield{author}{\bibinfo{person}{Tero Karras}, \bibinfo{person}{Samuli
  Laine}, {and} \bibinfo{person}{Timo Aila}.} \bibinfo{year}{2019}\natexlab{}.
\newblock \showarticletitle{A style-based generator architecture for generative
  adversarial networks}. In \bibinfo{booktitle}{\emph{IEEE Conference on
  Computer Vision and Pattern Recognition (CVPR)}}.
  \bibinfo{pages}{4401--4410}.
\newblock


\bibitem[\protect\citeauthoryear{Kemelmacher-Shlizerman and
  Seitz}{Kemelmacher-Shlizerman and Seitz}{2011}]%
        {KemelmacherSeitz2011}
\bibfield{author}{\bibinfo{person}{Ira Kemelmacher-Shlizerman} {and}
  \bibinfo{person}{Steven~M Seitz}.} \bibinfo{year}{2011}\natexlab{}.
\newblock \showarticletitle{Face reconstruction in the wild}. In
  \bibinfo{booktitle}{\emph{IEEE International Conference on Computer Vision
  (ICCV)}}. \bibinfo{pages}{1746--1753}.
\newblock


\bibitem[\protect\citeauthoryear{Kim, Garrido, Tewari, Xu, Thies, Nie{\ss}ner,
  P{\'{e}}rez, Richardt, Zollh{\"{o}}fer, and Theobalt}{Kim
  et~al\mbox{.}}{2018a}]%
        {Kim2018_DeepVideo}
\bibfield{author}{\bibinfo{person}{Hyeongwoo Kim}, \bibinfo{person}{Pablo
  Garrido}, \bibinfo{person}{Ayush Tewari}, \bibinfo{person}{Weipeng Xu},
  \bibinfo{person}{Justus Thies}, \bibinfo{person}{Matthias Nie{\ss}ner},
  \bibinfo{person}{Patrick P{\'{e}}rez}, \bibinfo{person}{Christian Richardt},
  \bibinfo{person}{Michael Zollh{\"{o}}fer}, {and} \bibinfo{person}{Christian
  Theobalt}.} \bibinfo{year}{2018}\natexlab{a}.
\newblock \showarticletitle{Deep video portraits}.
\newblock \bibinfo{journal}{\emph{ACM Transactions on Graphics (TOG)}}
  \bibinfo{volume}{37}, \bibinfo{number}{4} (\bibinfo{year}{2018}),
  \bibinfo{pages}{163:1--163:14}.
\newblock


\bibitem[\protect\citeauthoryear{Kim, Zollh{\"o}fer, Tewari, Thies, Richardt,
  and Theobalt}{Kim et~al\mbox{.}}{2018b}]%
        {Kim2018}
\bibfield{author}{\bibinfo{person}{Hyeongwoo Kim}, \bibinfo{person}{Michael
  Zollh{\"o}fer}, \bibinfo{person}{Ayush Tewari}, \bibinfo{person}{Justus
  Thies}, \bibinfo{person}{Christian Richardt}, {and}
  \bibinfo{person}{Christian Theobalt}.} \bibinfo{year}{2018}\natexlab{b}.
\newblock \showarticletitle{{InverseFaceNet:} Deep Monocular Inverse Face
  Rendering}. In \bibinfo{booktitle}{\emph{IEEE Conference on Computer Vision
  and Pattern Recognition (CVPR)}}. \bibinfo{pages}{4625--4634}.
\newblock


\bibitem[\protect\citeauthoryear{Kingma and Ba}{Kingma and Ba}{2015}]%
        {Kingma2015AdamAM}
\bibfield{author}{\bibinfo{person}{Diederik~P. Kingma} {and}
  \bibinfo{person}{Jimmy Ba}.} \bibinfo{year}{2015}\natexlab{}.
\newblock \showarticletitle{Adam: {A} Method for Stochastic Optimization}. In
  \bibinfo{booktitle}{\emph{International Conference on Learning
  Representations (ICLR)}}.
\newblock


\bibitem[\protect\citeauthoryear{K{\"{o}}stinger, Wohlhart, Roth, and
  Bischof}{K{\"{o}}stinger et~al\mbox{.}}{2011}]%
        {AFLW2011}
\bibfield{author}{\bibinfo{person}{Martin K{\"{o}}stinger},
  \bibinfo{person}{Paul Wohlhart}, \bibinfo{person}{Peter~M. Roth}, {and}
  \bibinfo{person}{Horst Bischof}.} \bibinfo{year}{2011}\natexlab{}.
\newblock \showarticletitle{Annotated Facial Landmarks in the Wild: {A}
  large-scale, real-world database for facial landmark localization}. In
  \bibinfo{booktitle}{\emph{IEEE International Conference on Computer Vision
  Workshops (ICCV-W)}}. \bibinfo{pages}{2144--2151}.
\newblock


\bibitem[\protect\citeauthoryear{Larsen, S{\o}nderby, Larochelle, and
  Winther}{Larsen et~al\mbox{.}}{2016}]%
        {Larsen2016AutoencodingBP}
\bibfield{author}{\bibinfo{person}{Anders Boesen~Lindbo Larsen},
  \bibinfo{person}{S{\o}ren~Kaae S{\o}nderby}, \bibinfo{person}{Hugo
  Larochelle}, {and} \bibinfo{person}{Ole Winther}.}
  \bibinfo{year}{2016}\natexlab{}.
\newblock \showarticletitle{Autoencoding beyond pixels using a learned
  similarity metric}. In \bibinfo{booktitle}{\emph{International Conference on
  Machine Learning (ICML)}}, Vol.~\bibinfo{volume}{48}.
  \bibinfo{pages}{1558--1566}.
\newblock


\bibitem[\protect\citeauthoryear{Lattas, Moschoglou, Gecer, Ploumpis,
  Triantafyllou, Ghosh, and Zafeiriou}{Lattas et~al\mbox{.}}{2020}]%
        {Lattas2020}
\bibfield{author}{\bibinfo{person}{Alexandros Lattas},
  \bibinfo{person}{Stylianos Moschoglou}, \bibinfo{person}{Baris Gecer},
  \bibinfo{person}{Stylianos Ploumpis}, \bibinfo{person}{Vasileios
  Triantafyllou}, \bibinfo{person}{Abhijeet Ghosh}, {and}
  \bibinfo{person}{Stefanos Zafeiriou}.} \bibinfo{year}{2020}\natexlab{}.
\newblock \showarticletitle{{AvatarMe}: Realistically Renderable {3D} Facial
  Reconstruction "In-the-Wild"}. In \bibinfo{booktitle}{\emph{IEEE Conference
  on Computer Vision and Pattern Recognition (CVPR)}}.
  \bibinfo{pages}{757--766}.
\newblock


\bibitem[\protect\citeauthoryear{Li, Adams, Guibas, and Pauly}{Li
  et~al\mbox{.}}{2009}]%
        {Li2009RobustSG}
\bibfield{author}{\bibinfo{person}{Hao Li}, \bibinfo{person}{Bart Adams},
  \bibinfo{person}{Leonidas~J. Guibas}, {and} \bibinfo{person}{Mark Pauly}.}
  \bibinfo{year}{2009}\natexlab{}.
\newblock \showarticletitle{Robust single-view geometry and motion
  reconstruction}.
\newblock \bibinfo{journal}{\emph{ACM Transactions on Graphics (TOG)}}
  \bibinfo{volume}{28} (\bibinfo{year}{2009}), \bibinfo{pages}{175}.
\newblock


\bibitem[\protect\citeauthoryear{Li, Yu, Ye, and Bregler}{Li
  et~al\mbox{.}}{2013}]%
        {Li2013}
\bibfield{author}{\bibinfo{person}{Hao Li}, \bibinfo{person}{Jihun Yu},
  \bibinfo{person}{Yuting Ye}, {and} \bibinfo{person}{Chris Bregler}.}
  \bibinfo{year}{2013}\natexlab{}.
\newblock \showarticletitle{Realtime facial animation with on-the-fly
  correctives}.
\newblock \bibinfo{journal}{\emph{ACM Transactions on Graphics (TOG)}}
  \bibinfo{volume}{32}, \bibinfo{number}{4} (\bibinfo{year}{2013}),
  \bibinfo{pages}{42--1}.
\newblock


\bibitem[\protect\citeauthoryear{Li, Bolkart, Black, Li, and Romero}{Li
  et~al\mbox{.}}{2017}]%
        {FLAME2017}
\bibfield{author}{\bibinfo{person}{Tianye Li}, \bibinfo{person}{Timo Bolkart},
  \bibinfo{person}{Michael.~J. Black}, \bibinfo{person}{Hao Li}, {and}
  \bibinfo{person}{Javier Romero}.} \bibinfo{year}{2017}\natexlab{}.
\newblock \showarticletitle{Learning a model of facial shape and expression
  from {4D} scans}.
\newblock \bibinfo{journal}{\emph{ACM Transactions on Graphics, (Proc. SIGGRAPH
  Asia)}} \bibinfo{volume}{36}, \bibinfo{number}{6} (\bibinfo{year}{2017}),
  \bibinfo{pages}{194:1--194:17}.
\newblock


\bibitem[\protect\citeauthoryear{Li, Ma, Fan, and Mitchell}{Li
  et~al\mbox{.}}{2018}]%
        {Li2018}
\bibfield{author}{\bibinfo{person}{Yue Li}, \bibinfo{person}{Liqian Ma},
  \bibinfo{person}{Haoqiang Fan}, {and} \bibinfo{person}{Kenny Mitchell}.}
  \bibinfo{year}{2018}\natexlab{}.
\newblock \showarticletitle{Feature-preserving detailed {3D} face
  reconstruction from a single image}. In \bibinfo{booktitle}{\emph{European
  Conference on Visual Media Production}}. \bibinfo{pages}{1--9}.
\newblock


\bibitem[\protect\citeauthoryear{Ma, Correll, and Wittenbrink}{Ma
  et~al\mbox{.}}{2015}]%
        {Chicago}
\bibfield{author}{\bibinfo{person}{Debbie~S. Ma}, \bibinfo{person}{Joshua
  Correll}, {and} \bibinfo{person}{Bernd Wittenbrink}.}
  \bibinfo{year}{2015}\natexlab{}.
\newblock \showarticletitle{The Chicago face database: A free stimulus set of
  faces and norming datan}.
\newblock \bibinfo{journal}{\emph{Behavior Research Methods volume}}
  \bibinfo{volume}{47} (\bibinfo{year}{2015}), \bibinfo{pages}{1122--1135}.
\newblock


\bibitem[\protect\citeauthoryear{Ma, Jones, Chiang, Hawkins, Frederiksen,
  Peers, Vukovic, Ouhyoung, and Debevec}{Ma et~al\mbox{.}}{2008}]%
        {Ma2008}
\bibfield{author}{\bibinfo{person}{Wan{-}Chun Ma}, \bibinfo{person}{Andrew
  Jones}, \bibinfo{person}{Jen{-}Yuan Chiang}, \bibinfo{person}{Tim Hawkins},
  \bibinfo{person}{Sune Frederiksen}, \bibinfo{person}{Pieter Peers},
  \bibinfo{person}{Marko Vukovic}, \bibinfo{person}{Ming Ouhyoung}, {and}
  \bibinfo{person}{Paul~E. Debevec}.} \bibinfo{year}{2008}\natexlab{}.
\newblock \showarticletitle{Facial performance synthesis using
  deformation-driven polynomial displacement maps}.
\newblock \bibinfo{journal}{\emph{ACM Transactions on Graphics (TOG)}}
  \bibinfo{volume}{27}, \bibinfo{number}{5} (\bibinfo{year}{2008}),
  \bibinfo{pages}{121}.
\newblock


\bibitem[\protect\citeauthoryear{Morales, Piella, and Sukno}{Morales
  et~al\mbox{.}}{2021}]%
        {Morales2021}
\bibfield{author}{\bibinfo{person}{Araceli Morales}, \bibinfo{person}{Gemma
  Piella}, {and} \bibinfo{person}{Federico~M Sukno}.}
  \bibinfo{year}{2021}\natexlab{}.
\newblock \showarticletitle{Survey on {3D} face reconstruction from
  uncalibrated images}.
\newblock \bibinfo{journal}{\emph{Computer Science Review}}
  \bibinfo{volume}{40} (\bibinfo{year}{2021}), \bibinfo{pages}{100400}.
\newblock


\bibitem[\protect\citeauthoryear{Nagano, Seo, Xing, Wei, Li, Saito, Agarwal,
  Fursund, and Li}{Nagano et~al\mbox{.}}{2018}]%
        {Nagano2018}
\bibfield{author}{\bibinfo{person}{Koki Nagano}, \bibinfo{person}{Jaewoo Seo},
  \bibinfo{person}{Jun Xing}, \bibinfo{person}{Lingyu Wei},
  \bibinfo{person}{Zimo Li}, \bibinfo{person}{Shunsuke Saito},
  \bibinfo{person}{Aviral Agarwal}, \bibinfo{person}{Jens Fursund}, {and}
  \bibinfo{person}{Hao Li}.} \bibinfo{year}{2018}\natexlab{}.
\newblock \showarticletitle{{paGAN}: real-time avatars using dynamic textures}.
\newblock \bibinfo{journal}{\emph{ACM Transactions on Graphics (TOG)}}
  \bibinfo{volume}{37}, \bibinfo{number}{6} (\bibinfo{year}{2018}),
  \bibinfo{pages}{258:1--258:12}.
\newblock


\bibitem[\protect\citeauthoryear{Nirkin, Masi, Tuan, Hassner, and
  Medioni}{Nirkin et~al\mbox{.}}{2018}]%
        {Nirkin2018}
\bibfield{author}{\bibinfo{person}{Yuval Nirkin}, \bibinfo{person}{Iacopo
  Masi}, \bibinfo{person}{Anh~Tran Tuan}, \bibinfo{person}{Tal Hassner}, {and}
  \bibinfo{person}{Gerard Medioni}.} \bibinfo{year}{2018}\natexlab{}.
\newblock \showarticletitle{On face segmentation, face swapping, and face
  perception}. In \bibinfo{booktitle}{\emph{International Conference on
  Automatic Face \& Gesture Recognition (FG)}}. \bibinfo{pages}{98--105}.
\newblock


\bibitem[\protect\citeauthoryear{{NoW challenge}}{{NoW challenge}}{2019}]%
        {NoW2019}
\bibfield{author}{\bibinfo{person}{{NoW challenge}}.}
  \bibinfo{year}{2019}\natexlab{}.
\newblock
  \bibinfo{howpublished}{\url{https://ringnet.is.tue.mpg.de/challenge}}.
\newblock


\bibitem[\protect\citeauthoryear{Parke}{Parke}{1974}]%
        {Parke1974}
\bibfield{author}{\bibinfo{person}{Frederick~Ira Parke}.}
  \bibinfo{year}{1974}\natexlab{}.
\newblock \bibinfo{booktitle}{\emph{A parametric model for human faces}}.
\newblock \bibinfo{type}{{T}echnical {R}eport}.
  \bibinfo{institution}{University of Utah}.
\newblock


\bibitem[\protect\citeauthoryear{Paszke, Gross, Massa, Lerer, Bradbury, Chanan,
  Killeen, Lin, Gimelshein, Antiga, Desmaison, K{\"o}pf, Yang, DeVito, Raison,
  Tejani, Chilamkurthy, Steiner, Fang, Bai, and Chintala}{Paszke
  et~al\mbox{.}}{2019}]%
        {Paszke2019PyTorchAI}
\bibfield{author}{\bibinfo{person}{Adam Paszke}, \bibinfo{person}{Sam Gross},
  \bibinfo{person}{Francisco Massa}, \bibinfo{person}{Adam Lerer},
  \bibinfo{person}{James Bradbury}, \bibinfo{person}{Gregory Chanan},
  \bibinfo{person}{Trevor Killeen}, \bibinfo{person}{Zeming Lin},
  \bibinfo{person}{Natalia Gimelshein}, \bibinfo{person}{Luca Antiga},
  \bibinfo{person}{Alban Desmaison}, \bibinfo{person}{Andreas K{\"o}pf},
  \bibinfo{person}{Edward Yang}, \bibinfo{person}{Zach DeVito},
  \bibinfo{person}{Martin Raison}, \bibinfo{person}{Alykhan Tejani},
  \bibinfo{person}{Sasank Chilamkurthy}, \bibinfo{person}{Benoit Steiner},
  \bibinfo{person}{Lu Fang}, \bibinfo{person}{Junjie Bai}, {and}
  \bibinfo{person}{Soumith Chintala}.} \bibinfo{year}{2019}\natexlab{}.
\newblock \showarticletitle{PyTorch: An Imperative Style, High-Performance Deep
  Learning Library}. In \bibinfo{booktitle}{\emph{Advances in Neural
  Information Processing Systems (NeurIPS)}}.
\newblock


\bibitem[\protect\citeauthoryear{Paysan, Knothe, Amberg, Romdhani, and
  Vetter}{Paysan et~al\mbox{.}}{2009}]%
        {Paysan2009}
\bibfield{author}{\bibinfo{person}{Pascal Paysan}, \bibinfo{person}{Reinhard
  Knothe}, \bibinfo{person}{Brian Amberg}, \bibinfo{person}{Sami Romdhani},
  {and} \bibinfo{person}{Thomas Vetter}.} \bibinfo{year}{2009}\natexlab{}.
\newblock \showarticletitle{A {3D} face model for pose and illumination
  invariant face recognition}. In \bibinfo{booktitle}{\emph{International
  Conference on Advanced Video and Signal Based Surveillance}}.
  \bibinfo{pages}{296--301}.
\newblock


\bibitem[\protect\citeauthoryear{{Pexels}}{{Pexels}}{2021}]%
        {Pexels2021}
\bibfield{author}{\bibinfo{person}{{Pexels}}.} \bibinfo{year}{2021}\natexlab{}.
\newblock \bibinfo{howpublished}{\url{https://www.pexels.com}}.
\newblock


\bibitem[\protect\citeauthoryear{Pighin, Hecker, Lischinski, Szeliski, and
  Salesin}{Pighin et~al\mbox{.}}{1998}]%
        {Pighin1998}
\bibfield{author}{\bibinfo{person}{Fr{\'e}d{\'e}ric Pighin},
  \bibinfo{person}{Jamie Hecker}, \bibinfo{person}{Dani Lischinski},
  \bibinfo{person}{Richard Szeliski}, {and} \bibinfo{person}{David~H.
  Salesin}.} \bibinfo{year}{1998}\natexlab{}.
\newblock \showarticletitle{Synthesizing Realistic Facial Expressions from
  Photographs}. In \bibinfo{booktitle}{\emph{SIGGRAPH}}.
  \bibinfo{pages}{75--84}.
\newblock


\bibitem[\protect\citeauthoryear{Ploumpis, Ververas, O'Sullivan, Moschoglou,
  Wang, Pears, Smith, Gecer, and Zafeiriou}{Ploumpis et~al\mbox{.}}{2020}]%
        {Ploumpis2020}
\bibfield{author}{\bibinfo{person}{Stylianos Ploumpis},
  \bibinfo{person}{Evangelos Ververas}, \bibinfo{person}{Eimear O'Sullivan},
  \bibinfo{person}{Stylianos Moschoglou}, \bibinfo{person}{Haoyang Wang},
  \bibinfo{person}{Nick Pears}, \bibinfo{person}{William Smith},
  \bibinfo{person}{Baris Gecer}, {and} \bibinfo{person}{Stefanos~P Zafeiriou}.}
  \bibinfo{year}{2020}\natexlab{}.
\newblock \showarticletitle{Towards a complete {3D} morphable model of the
  human head}.
\newblock \bibinfo{journal}{\emph{IEEE Transactions on Pattern Analysis and
  Machine Intelligence (PAMI)}} (\bibinfo{year}{2020}).
\newblock


\bibitem[\protect\citeauthoryear{Ramamoorthi and Hanrahan}{Ramamoorthi and
  Hanrahan}{2001}]%
        {Ramamoorthi2001AnER}
\bibfield{author}{\bibinfo{person}{R. Ramamoorthi} {and} \bibinfo{person}{P.
  Hanrahan}.} \bibinfo{year}{2001}\natexlab{}.
\newblock \showarticletitle{An efficient representation for irradiance
  environment maps}.
\newblock \bibinfo{journal}{\emph{Proceedings of the 28th annual conference on
  Computer graphics and interactive techniques}} (\bibinfo{year}{2001}).
\newblock


\bibitem[\protect\citeauthoryear{Ravi, Reizenstein, Novotny, Gordon, Lo,
  Johnson, and Gkioxari}{Ravi et~al\mbox{.}}{2020}]%
        {ravi2020pytorch3d}
\bibfield{author}{\bibinfo{person}{Nikhila Ravi}, \bibinfo{person}{Jeremy
  Reizenstein}, \bibinfo{person}{David Novotny}, \bibinfo{person}{Taylor
  Gordon}, \bibinfo{person}{Wan-Yen Lo}, \bibinfo{person}{Justin Johnson},
  {and} \bibinfo{person}{Georgia Gkioxari}.} \bibinfo{year}{2020}\natexlab{}.
\newblock \bibinfo{title}{PyTorch3D}.
\newblock
  \bibinfo{howpublished}{\url{https://github.com/facebookresearch/pytorch3d}}.
\newblock


\bibitem[\protect\citeauthoryear{Richard, Zollh{\"{o}}fer, Wen, la~Torre, and
  Sheikh}{Richard et~al\mbox{.}}{2021}]%
        {Richard2021}
\bibfield{author}{\bibinfo{person}{Alexander Richard}, \bibinfo{person}{Michael
  Zollh{\"{o}}fer}, \bibinfo{person}{Yandong Wen}, \bibinfo{person}{Fernando~De
  la Torre}, {and} \bibinfo{person}{Yaser Sheikh}.}
  \bibinfo{year}{2021}\natexlab{}.
\newblock \showarticletitle{MeshTalk: {3D} Face Animation from Speech using
  Cross-Modality Disentanglement}.
\newblock \bibinfo{journal}{\emph{CoRR}}  \bibinfo{volume}{abs/2104.08223}
  (\bibinfo{year}{2021}).
\newblock


\bibitem[\protect\citeauthoryear{Richardson, Sela, and Kimmel}{Richardson
  et~al\mbox{.}}{2016}]%
        {Richardson2016}
\bibfield{author}{\bibinfo{person}{E. Richardson}, \bibinfo{person}{M. Sela},
  {and} \bibinfo{person}{R. Kimmel}.} \bibinfo{year}{2016}\natexlab{}.
\newblock \showarticletitle{{3D} Face Reconstruction by Learning from Synthetic
  Data}. In \bibinfo{booktitle}{\emph{International Conference on 3D Vision
  (3DV)}}. \bibinfo{pages}{460--469}.
\newblock


\bibitem[\protect\citeauthoryear{Richardson, Sela, Or-El, and
  Kimmel}{Richardson et~al\mbox{.}}{2017}]%
        {Richardson2017}
\bibfield{author}{\bibinfo{person}{Elad Richardson}, \bibinfo{person}{Matan
  Sela}, \bibinfo{person}{Roy Or-El}, {and} \bibinfo{person}{Ron Kimmel}.}
  \bibinfo{year}{2017}\natexlab{}.
\newblock \showarticletitle{Learning Detailed Face Reconstruction From a Single
  Image}. In \bibinfo{booktitle}{\emph{IEEE Conference on Computer Vision and
  Pattern Recognition (CVPR)}}. \bibinfo{pages}{1259--1268}.
\newblock


\bibitem[\protect\citeauthoryear{Riviere, Gotardo, Bradley, Ghosh, and
  Beeler}{Riviere et~al\mbox{.}}{2020}]%
        {Riviere2020}
\bibfield{author}{\bibinfo{person}{J{\'{e}}r{\'{e}}my Riviere},
  \bibinfo{person}{Paulo F.~U. Gotardo}, \bibinfo{person}{Derek Bradley},
  \bibinfo{person}{Abhijeet Ghosh}, {and} \bibinfo{person}{Thabo Beeler}.}
  \bibinfo{year}{2020}\natexlab{}.
\newblock \showarticletitle{Single-shot high-quality facial geometry and skin
  appearance capture}.
\newblock \bibinfo{journal}{\emph{ACM Transactions on Graphics, (Proc.
  SIGGRAPH)}} \bibinfo{volume}{39}, \bibinfo{number}{4} (\bibinfo{year}{2020}),
  \bibinfo{pages}{81}.
\newblock


\bibitem[\protect\citeauthoryear{Romdhani, Blanz, and Vetter}{Romdhani
  et~al\mbox{.}}{2002}]%
        {Romdhani2002}
\bibfield{author}{\bibinfo{person}{Sami Romdhani}, \bibinfo{person}{Volker
  Blanz}, {and} \bibinfo{person}{Thomas Vetter}.}
  \bibinfo{year}{2002}\natexlab{}.
\newblock \showarticletitle{Face identification by fitting a {3D} morphable
  model using linear shape and texture error functions}. In
  \bibinfo{booktitle}{\emph{European Conference on Computer Vision (ECCV)}}.
  \bibinfo{pages}{3--19}.
\newblock


\bibitem[\protect\citeauthoryear{Romdhani and Vetter}{Romdhani and
  Vetter}{2005}]%
        {RomdhaniVetter2005}
\bibfield{author}{\bibinfo{person}{Sami Romdhani} {and} \bibinfo{person}{Thomas
  Vetter}.} \bibinfo{year}{2005}\natexlab{}.
\newblock \showarticletitle{Estimating {3D} shape and texture using pixel
  intensity, edges, specular highlights, texture constraints and a prior}. In
  \bibinfo{booktitle}{\emph{IEEE Conference on Computer Vision and Pattern
  Recognition (CVPR)}}, Vol.~\bibinfo{volume}{2}. \bibinfo{pages}{986--993}.
\newblock


\bibitem[\protect\citeauthoryear{Roth, Tong, and Liu}{Roth
  et~al\mbox{.}}{2016}]%
        {Roth2016}
\bibfield{author}{\bibinfo{person}{Joseph Roth}, \bibinfo{person}{Yiying Tong},
  {and} \bibinfo{person}{Xiaoming Liu}.} \bibinfo{year}{2016}\natexlab{}.
\newblock \showarticletitle{Adaptive {3D} face reconstruction from
  unconstrained photo collections}. In \bibinfo{booktitle}{\emph{IEEE
  Conference on Computer Vision and Pattern Recognition (CVPR)}}.
  \bibinfo{pages}{4197--4206}.
\newblock


\bibitem[\protect\citeauthoryear{Saito, Wei, Hu, Nagano, and Li}{Saito
  et~al\mbox{.}}{2017}]%
        {Saito2017}
\bibfield{author}{\bibinfo{person}{Shunsuke Saito}, \bibinfo{person}{Lingyu
  Wei}, \bibinfo{person}{Liwen Hu}, \bibinfo{person}{Koki Nagano}, {and}
  \bibinfo{person}{Hao Li}.} \bibinfo{year}{2017}\natexlab{}.
\newblock \showarticletitle{Photorealistic Facial Texture Inference Using Deep
  Neural Networks}. In \bibinfo{booktitle}{\emph{IEEE Conference on Computer
  Vision and Pattern Recognition (CVPR)}}. \bibinfo{pages}{5144--5153}.
\newblock


\bibitem[\protect\citeauthoryear{Sanyal, Bolkart, Feng, and Black}{Sanyal
  et~al\mbox{.}}{2019}]%
        {Sanyal2019}
\bibfield{author}{\bibinfo{person}{Soubhik Sanyal}, \bibinfo{person}{Timo
  Bolkart}, \bibinfo{person}{Haiwen Feng}, {and} \bibinfo{person}{Michael
  Black}.} \bibinfo{year}{2019}\natexlab{}.
\newblock \showarticletitle{Learning to Regress {3D} Face Shape and Expression
  from an Image without {3D} Supervision}. In \bibinfo{booktitle}{\emph{IEEE
  Conference on Computer Vision and Pattern Recognition (CVPR)}}.
  \bibinfo{pages}{7763--7772}.
\newblock


\bibitem[\protect\citeauthoryear{Scherbaum, Ritschel, Hullin, Thorm{\"a}hlen,
  Blanz, and Seidel}{Scherbaum et~al\mbox{.}}{2011}]%
        {Scherbaum2011}
\bibfield{author}{\bibinfo{person}{Kristina Scherbaum}, \bibinfo{person}{Tobias
  Ritschel}, \bibinfo{person}{Matthias Hullin}, \bibinfo{person}{Thorsten
  Thorm{\"a}hlen}, \bibinfo{person}{Volker Blanz}, {and}
  \bibinfo{person}{Hans-Peter Seidel}.} \bibinfo{year}{2011}\natexlab{}.
\newblock \showarticletitle{Computer-suggested facial makeup}.
\newblock \bibinfo{journal}{\emph{Computer Graphics Forum}}
  \bibinfo{volume}{30}, \bibinfo{number}{2} (\bibinfo{year}{2011}),
  \bibinfo{pages}{485--492}.
\newblock


\bibitem[\protect\citeauthoryear{Sela, Richardson, and Kimmel}{Sela
  et~al\mbox{.}}{2017}]%
        {Sela2017}
\bibfield{author}{\bibinfo{person}{Matan Sela}, \bibinfo{person}{Elad
  Richardson}, {and} \bibinfo{person}{Ron Kimmel}.}
  \bibinfo{year}{2017}\natexlab{}.
\newblock \showarticletitle{Unrestricted facial geometry reconstruction using
  image-to-image translation}. In \bibinfo{booktitle}{\emph{IEEE International
  Conference on Computer Vision (ICCV)}}. \bibinfo{pages}{1576--1585}.
\newblock


\bibitem[\protect\citeauthoryear{Sengupta, Kanazawa, Castillo, and
  Jacobs}{Sengupta et~al\mbox{.}}{2018}]%
        {Sengupta2018}
\bibfield{author}{\bibinfo{person}{Soumyadip Sengupta}, \bibinfo{person}{Angjoo
  Kanazawa}, \bibinfo{person}{Carlos~D. Castillo}, {and}
  \bibinfo{person}{David~W. Jacobs}.} \bibinfo{year}{2018}\natexlab{}.
\newblock \showarticletitle{{SfSNet:} Learning Shape, Reflectance and
  Illuminance of Faces in the Wild}. In \bibinfo{booktitle}{\emph{IEEE
  Conference on Computer Vision and Pattern Recognition (CVPR)}}.
  \bibinfo{pages}{6296--6305}.
\newblock


\bibitem[\protect\citeauthoryear{Shang, Shen, Li, Zhou, Zhen, Fang, and
  Quan}{Shang et~al\mbox{.}}{2020}]%
        {Shang2020}
\bibfield{author}{\bibinfo{person}{Jiaxiang Shang}, \bibinfo{person}{Tianwei
  Shen}, \bibinfo{person}{Shiwei Li}, \bibinfo{person}{Lei Zhou},
  \bibinfo{person}{Mingmin Zhen}, \bibinfo{person}{Tian Fang}, {and}
  \bibinfo{person}{Long Quan}.} \bibinfo{year}{2020}\natexlab{}.
\newblock \showarticletitle{Self-Supervised Monocular {3D} Face Reconstruction
  by Occlusion-Aware Multi-view Geometry Consistency}. In
  \bibinfo{booktitle}{\emph{European Conference on Computer Vision (ECCV)}},
  Vol.~\bibinfo{volume}{12360}. \bibinfo{pages}{53--70}.
\newblock


\bibitem[\protect\citeauthoryear{Shi, Wu, Tong, and Chai}{Shi
  et~al\mbox{.}}{2014}]%
        {Shi2014}
\bibfield{author}{\bibinfo{person}{Fuhao Shi}, \bibinfo{person}{Hsiang-Tao Wu},
  \bibinfo{person}{Xin Tong}, {and} \bibinfo{person}{Jinxiang Chai}.}
  \bibinfo{year}{2014}\natexlab{}.
\newblock \showarticletitle{Automatic acquisition of high-fidelity facial
  performances using monocular videos}.
\newblock \bibinfo{journal}{\emph{ACM Transactions on Graphics (TOG)}}
  \bibinfo{volume}{33}, \bibinfo{number}{6} (\bibinfo{year}{2014}),
  \bibinfo{pages}{222}.
\newblock


\bibitem[\protect\citeauthoryear{Shin, {\"O}ztireli, Kim, Beeler, Gross, and
  Choi}{Shin et~al\mbox{.}}{2014}]%
        {Shin2014}
\bibfield{author}{\bibinfo{person}{Il-Kyu Shin}, \bibinfo{person}{A~Cengiz
  {\"O}ztireli}, \bibinfo{person}{Hyeon-Joong Kim}, \bibinfo{person}{Thabo
  Beeler}, \bibinfo{person}{Markus Gross}, {and} \bibinfo{person}{Soo-Mi
  Choi}.} \bibinfo{year}{2014}\natexlab{}.
\newblock \showarticletitle{Extraction and transfer of facial expression
  wrinkles for facial performance enhancement}. In
  \bibinfo{booktitle}{\emph{Pacific Conference on Computer Graphics and
  Applications}}. \bibinfo{pages}{113--118}.
\newblock


\bibitem[\protect\citeauthoryear{Simonyan and Zisserman}{Simonyan and
  Zisserman}{2014}]%
        {Simonyan2014VeryDC}
\bibfield{author}{\bibinfo{person}{Karen Simonyan} {and}
  \bibinfo{person}{Andrew Zisserman}.} \bibinfo{year}{2014}\natexlab{}.
\newblock \showarticletitle{Very Deep Convolutional Networks for Large-Scale
  Image Recognition}.
\newblock \bibinfo{journal}{\emph{CoRR}}  \bibinfo{volume}{abs/1409.1556}
  (\bibinfo{year}{2014}).
\newblock


\bibitem[\protect\citeauthoryear{Slossberg, Shamai, and Kimmel}{Slossberg
  et~al\mbox{.}}{2018}]%
        {Slossberg2018}
\bibfield{author}{\bibinfo{person}{Ron Slossberg}, \bibinfo{person}{Gil
  Shamai}, {and} \bibinfo{person}{Ron Kimmel}.}
  \bibinfo{year}{2018}\natexlab{}.
\newblock \showarticletitle{High quality facial surface and texture synthesis
  via generative adversarial networks}. In \bibinfo{booktitle}{\emph{European
  Conference on Computer Vision Workshops (ECCV-W)}}.
\newblock


\bibitem[\protect\citeauthoryear{Suwajanakorn, Kemelmacher-Shlizerman, and
  Seitz}{Suwajanakorn et~al\mbox{.}}{2014}]%
        {Suwajanakorn2014}
\bibfield{author}{\bibinfo{person}{Supasorn Suwajanakorn}, \bibinfo{person}{Ira
  Kemelmacher-Shlizerman}, {and} \bibinfo{person}{Steven~M Seitz}.}
  \bibinfo{year}{2014}\natexlab{}.
\newblock \showarticletitle{Total moving face reconstruction}. In
  \bibinfo{booktitle}{\emph{European Conference on Computer Vision (ECCV)}}.
  \bibinfo{pages}{796--812}.
\newblock


\bibitem[\protect\citeauthoryear{Tewari, Bernard, Garrido, Bharaj, Elgharib,
  Seidel, P{\'e}rez, Zollh{\"o}fer, and Theobalt}{Tewari et~al\mbox{.}}{2019}]%
        {Tewari2019}
\bibfield{author}{\bibinfo{person}{Ayush Tewari}, \bibinfo{person}{Florian
  Bernard}, \bibinfo{person}{Pablo Garrido}, \bibinfo{person}{Gaurav Bharaj},
  \bibinfo{person}{Mohamed Elgharib}, \bibinfo{person}{Hans-Peter Seidel},
  \bibinfo{person}{Patrick P{\'e}rez}, \bibinfo{person}{Michael Zollh{\"o}fer},
  {and} \bibinfo{person}{Christian Theobalt}.} \bibinfo{year}{2019}\natexlab{}.
\newblock \showarticletitle{{FML:} Face Model Learning from Videos}. In
  \bibinfo{booktitle}{\emph{IEEE Conference on Computer Vision and Pattern
  Recognition (CVPR)}}. \bibinfo{pages}{10812--10822}.
\newblock


\bibitem[\protect\citeauthoryear{Tewari, Elgharib, Bharaj, Bernard, Seidel,
  P{\'{e}}rez, Zollh{\"{o}}fer, and Theobalt}{Tewari et~al\mbox{.}}{2020}]%
        {Tewari2020}
\bibfield{author}{\bibinfo{person}{Ayush Tewari}, \bibinfo{person}{Mohamed
  Elgharib}, \bibinfo{person}{Gaurav Bharaj}, \bibinfo{person}{Florian
  Bernard}, \bibinfo{person}{Hans{-}Peter Seidel}, \bibinfo{person}{Patrick
  P{\'{e}}rez}, \bibinfo{person}{Michael Zollh{\"{o}}fer}, {and}
  \bibinfo{person}{Christian Theobalt}.} \bibinfo{year}{2020}\natexlab{}.
\newblock \showarticletitle{{StyleRig}: Rigging StyleGAN for {3D} Control Over
  Portrait Images}. In \bibinfo{booktitle}{\emph{IEEE Conference on Computer
  Vision and Pattern Recognition (CVPR)}}. \bibinfo{pages}{6141--6150}.
\newblock


\bibitem[\protect\citeauthoryear{Tewari, Zollh{\"o}fer, Garrido, Bernard, Kim,
  P{\'e}rez, and Theobalt}{Tewari et~al\mbox{.}}{2018}]%
        {Tewari2018}
\bibfield{author}{\bibinfo{person}{Ayush Tewari}, \bibinfo{person}{Michael
  Zollh{\"o}fer}, \bibinfo{person}{Pablo Garrido}, \bibinfo{person}{Florian
  Bernard}, \bibinfo{person}{Hyeongwoo Kim}, \bibinfo{person}{Patrick
  P{\'e}rez}, {and} \bibinfo{person}{Christian Theobalt}.}
  \bibinfo{year}{2018}\natexlab{}.
\newblock \showarticletitle{Self-supervised multi-level face model learning for
  monocular reconstruction at over 250 {Hz}}. In \bibinfo{booktitle}{\emph{IEEE
  Conference on Computer Vision and Pattern Recognition (CVPR)}}.
  \bibinfo{pages}{2549--2559}.
\newblock


\bibitem[\protect\citeauthoryear{Tewari, Zollh{\"o}fer, Kim, Garrido, Bernard,
  Perez, and Theobalt}{Tewari et~al\mbox{.}}{2017}]%
        {Tewari2017}
\bibfield{author}{\bibinfo{person}{Ayush Tewari}, \bibinfo{person}{Michael
  Zollh{\"o}fer}, \bibinfo{person}{Hyeongwoo Kim}, \bibinfo{person}{Pablo
  Garrido}, \bibinfo{person}{Florian Bernard}, \bibinfo{person}{Patrick Perez},
  {and} \bibinfo{person}{Christian Theobalt}.} \bibinfo{year}{2017}\natexlab{}.
\newblock \showarticletitle{{MoFA:} Model-Based Deep Convolutional Face
  Autoencoder for Unsupervised Monocular Reconstruction}. In
  \bibinfo{booktitle}{\emph{IEEE International Conference on Computer Vision
  (ICCV)}}. \bibinfo{pages}{1274--1283}.
\newblock


\bibitem[\protect\citeauthoryear{Thies, Zollh{\"o}fer, Nie{\ss}ner, Valgaerts,
  Stamminger, and Theobalt}{Thies et~al\mbox{.}}{2015}]%
        {Thies2015}
\bibfield{author}{\bibinfo{person}{Justus Thies}, \bibinfo{person}{Michael
  Zollh{\"o}fer}, \bibinfo{person}{Matthias Nie{\ss}ner}, \bibinfo{person}{Levi
  Valgaerts}, \bibinfo{person}{Marc Stamminger}, {and}
  \bibinfo{person}{Christian Theobalt}.} \bibinfo{year}{2015}\natexlab{}.
\newblock \showarticletitle{Real-time expression transfer for facial
  reenactment}.
\newblock \bibinfo{journal}{\emph{ACM Transactions on Graphics (TOG)}}
  \bibinfo{volume}{34}, \bibinfo{number}{6} (\bibinfo{year}{2015}),
  \bibinfo{pages}{183--1}.
\newblock


\bibitem[\protect\citeauthoryear{Thies, Zollh{\"o}fer, Stamminger, Theobalt,
  and Nie{\ss}ner}{Thies et~al\mbox{.}}{2016}]%
        {Thies2016}
\bibfield{author}{\bibinfo{person}{Justus Thies}, \bibinfo{person}{Michael
  Zollh{\"o}fer}, \bibinfo{person}{Marc Stamminger}, \bibinfo{person}{Christian
  Theobalt}, {and} \bibinfo{person}{Matthias Nie{\ss}ner}.}
  \bibinfo{year}{2016}\natexlab{}.
\newblock \showarticletitle{{Face2Face}: Real-time face capture and reenactment
  of {RGB} videos}. In \bibinfo{booktitle}{\emph{IEEE Conference on Computer
  Vision and Pattern Recognition (CVPR)}}. \bibinfo{pages}{2387--2395}.
\newblock


\bibitem[\protect\citeauthoryear{Tran, Hassner, Masi, and Medioni}{Tran
  et~al\mbox{.}}{2017}]%
        {AnhTran2017}
\bibfield{author}{\bibinfo{person}{Anh~Tuan Tran}, \bibinfo{person}{Tal
  Hassner}, \bibinfo{person}{Iacopo Masi}, {and} \bibinfo{person}{Gerard
  Medioni}.} \bibinfo{year}{2017}\natexlab{}.
\newblock \showarticletitle{Regressing Robust and Discriminative {3D} Morphable
  Models With a Very Deep Neural Network}. In \bibinfo{booktitle}{\emph{IEEE
  Conference on Computer Vision and Pattern Recognition (CVPR)}}.
  \bibinfo{pages}{1599--1608}.
\newblock


\bibitem[\protect\citeauthoryear{Tran, Hassner, Masi, Paz, Nirkin, and
  Medioni}{Tran et~al\mbox{.}}{2018}]%
        {AnhTran2018}
\bibfield{author}{\bibinfo{person}{Anh~Tuan Tran}, \bibinfo{person}{Tal
  Hassner}, \bibinfo{person}{Iacopo Masi}, \bibinfo{person}{Eran Paz},
  \bibinfo{person}{Yuval Nirkin}, {and} \bibinfo{person}{G{\'e}rard Medioni}.}
  \bibinfo{year}{2018}\natexlab{}.
\newblock \showarticletitle{Extreme {3D} face reconstruction: Seeing through
  occlusions}. In \bibinfo{booktitle}{\emph{IEEE Conference on Computer Vision
  and Pattern Recognition (CVPR)}}. \bibinfo{pages}{3935--3944}.
\newblock


\bibitem[\protect\citeauthoryear{Tran, Liu, and Liu}{Tran
  et~al\mbox{.}}{2019}]%
        {LuanTran2019}
\bibfield{author}{\bibinfo{person}{Luan Tran}, \bibinfo{person}{Feng Liu},
  {and} \bibinfo{person}{Xiaoming Liu}.} \bibinfo{year}{2019}\natexlab{}.
\newblock \showarticletitle{Towards High-Fidelity Nonlinear {3D} Face Morphable
  Model}. In \bibinfo{booktitle}{\emph{IEEE Conference on Computer Vision and
  Pattern Recognition (CVPR)}}. \bibinfo{pages}{1126--1135}.
\newblock


\bibitem[\protect\citeauthoryear{Tu, Zhao, Jiang, Luo, Xie, Zhao, He, Ma, and
  Feng}{Tu et~al\mbox{.}}{2019}]%
        {Tu2019}
\bibfield{author}{\bibinfo{person}{Xiaoguang Tu}, \bibinfo{person}{Jian Zhao},
  \bibinfo{person}{Zihang Jiang}, \bibinfo{person}{Yao Luo},
  \bibinfo{person}{Mei Xie}, \bibinfo{person}{Yang Zhao},
  \bibinfo{person}{Linxiao He}, \bibinfo{person}{Zheng Ma}, {and}
  \bibinfo{person}{Jiashi Feng}.} \bibinfo{year}{2019}\natexlab{}.
\newblock \showarticletitle{Joint {3D} Face Reconstruction and Dense Face
  Alignment from A Single Image with {2D}-Assisted Self-Supervised Learning}.
\newblock \bibinfo{journal}{\emph{IEEE International Conference on Computer
  Vision (ICCV)}} (\bibinfo{year}{2019}).
\newblock


\bibitem[\protect\citeauthoryear{Vetter and Blanz}{Vetter and Blanz}{1998}]%
        {VetterBlanz1998}
\bibfield{author}{\bibinfo{person}{Thomas Vetter} {and} \bibinfo{person}{Volker
  Blanz}.} \bibinfo{year}{1998}\natexlab{}.
\newblock \showarticletitle{Estimating coloured {3D} face models from single
  images: An example based approach}. In \bibinfo{booktitle}{\emph{European
  Conference on Computer Vision (ECCV)}}. \bibinfo{pages}{499--513}.
\newblock


\bibitem[\protect\citeauthoryear{Wang, Deng, Hu, Tao, and Huang}{Wang
  et~al\mbox{.}}{2019}]%
        {Wang_2019_ICCV}
\bibfield{author}{\bibinfo{person}{Mei Wang}, \bibinfo{person}{Weihong Deng},
  \bibinfo{person}{Jiani Hu}, \bibinfo{person}{Xunqiang Tao}, {and}
  \bibinfo{person}{Yaohai Huang}.} \bibinfo{year}{2019}\natexlab{}.
\newblock \showarticletitle{Racial Faces in the Wild: Reducing Racial Bias by
  Information Maximization Adaptation Network}. In
  \bibinfo{booktitle}{\emph{IEEE International Conference on Computer Vision
  (ICCV)}}.
\newblock


\bibitem[\protect\citeauthoryear{Wang, Tao, Qi, Shen, and Jia}{Wang
  et~al\mbox{.}}{2018}]%
        {Wang2018}
\bibfield{author}{\bibinfo{person}{Yi Wang}, \bibinfo{person}{Xin Tao},
  \bibinfo{person}{Xiaojuan Qi}, \bibinfo{person}{Xiaoyong Shen}, {and}
  \bibinfo{person}{Jiaya Jia}.} \bibinfo{year}{2018}\natexlab{}.
\newblock \showarticletitle{Image inpainting via generative multi-column
  convolutional neural networks}. In \bibinfo{booktitle}{\emph{Advances in
  Neural Information Processing Systems (NeurIPS)}}. \bibinfo{pages}{331--340}.
\newblock


\bibitem[\protect\citeauthoryear{Wei, Liang, and Wei}{Wei
  et~al\mbox{.}}{2019}]%
        {Wei2019}
\bibfield{author}{\bibinfo{person}{Huawei Wei}, \bibinfo{person}{Shuang Liang},
  {and} \bibinfo{person}{Yichen Wei}.} \bibinfo{year}{2019}\natexlab{}.
\newblock \showarticletitle{{3D} Dense Face Alignment via Graph Convolution
  Networks}.
\newblock \bibinfo{journal}{\emph{arXiv preprint arXiv:1904.05562}}
  (\bibinfo{year}{2019}).
\newblock


\bibitem[\protect\citeauthoryear{Weise, Bouaziz, Li, and Pauly}{Weise
  et~al\mbox{.}}{2011}]%
        {Weise2011}
\bibfield{author}{\bibinfo{person}{Thibaut Weise}, \bibinfo{person}{Sofien
  Bouaziz}, \bibinfo{person}{Hao Li}, {and} \bibinfo{person}{Mark Pauly}.}
  \bibinfo{year}{2011}\natexlab{}.
\newblock \showarticletitle{Realtime performance-based facial animation}.
\newblock \bibinfo{journal}{\emph{ACM Transactions on Graphics, (Proc.
  SIGGRAPH)}} \bibinfo{volume}{30}, \bibinfo{number}{4} (\bibinfo{year}{2011}),
  \bibinfo{pages}{77}.
\newblock


\bibitem[\protect\citeauthoryear{Yamaguchi, Saito, Nagano, Zhao, Chen,
  Olszewski, Morishima, and Li}{Yamaguchi et~al\mbox{.}}{2018}]%
        {Yamaguchi2018}
\bibfield{author}{\bibinfo{person}{Shugo Yamaguchi}, \bibinfo{person}{Shunsuke
  Saito}, \bibinfo{person}{Koki Nagano}, \bibinfo{person}{Yajie Zhao},
  \bibinfo{person}{Weikai Chen}, \bibinfo{person}{Kyle Olszewski},
  \bibinfo{person}{Shigeo Morishima}, {and} \bibinfo{person}{Hao Li}.}
  \bibinfo{year}{2018}\natexlab{}.
\newblock \showarticletitle{High-fidelity Facial Reflectance and Geometry
  Inference from an Unconstrained Image}.
\newblock \bibinfo{journal}{\emph{ACM Transactions on Graphics (TOG)}}
  \bibinfo{volume}{37}, \bibinfo{number}{4} (\bibinfo{year}{2018}),
  \bibinfo{pages}{162:1--162:14}.
\newblock


\bibitem[\protect\citeauthoryear{Yang, Zhu, Wang, Huang, Shen, Yang, and
  Cao}{Yang et~al\mbox{.}}{2020}]%
        {yang2020facescape}
\bibfield{author}{\bibinfo{person}{Haotian Yang}, \bibinfo{person}{Hao Zhu},
  \bibinfo{person}{Yanru Wang}, \bibinfo{person}{Mingkai Huang},
  \bibinfo{person}{Qiu Shen}, \bibinfo{person}{Ruigang Yang}, {and}
  \bibinfo{person}{Xun Cao}.} \bibinfo{year}{2020}\natexlab{}.
\newblock \showarticletitle{{FaceScape}: a Large-scale High Quality {3D} Face
  Dataset and Detailed Riggable {3D} Face Prediction}. In
  \bibinfo{booktitle}{\emph{IEEE Conference on Computer Vision and Pattern
  Recognition (CVPR)}}. \bibinfo{pages}{601--610}.
\newblock


\bibitem[\protect\citeauthoryear{Zeng, Peng, and Qiao}{Zeng
  et~al\mbox{.}}{2019}]%
        {Zeng2019}
\bibfield{author}{\bibinfo{person}{Xiaoxing Zeng}, \bibinfo{person}{Xiaojiang
  Peng}, {and} \bibinfo{person}{Yu Qiao}.} \bibinfo{year}{2019}\natexlab{}.
\newblock \showarticletitle{{DF2Net}: A Dense-Fine-Finer Network for Detailed
  {3D} Face Reconstruction}. In \bibinfo{booktitle}{\emph{IEEE International
  Conference on Computer Vision (ICCV)}}.
\newblock


\bibitem[\protect\citeauthoryear{Zhao, Huang, Li, Chen, LeGendre, Ren, Shapiro,
  and Li}{Zhao et~al\mbox{.}}{2019}]%
        {zhao2019learning}
\bibfield{author}{\bibinfo{person}{Yajie Zhao}, \bibinfo{person}{Zeng Huang},
  \bibinfo{person}{Tianye Li}, \bibinfo{person}{Weikai Chen},
  \bibinfo{person}{Chloe LeGendre}, \bibinfo{person}{Xinglei Ren},
  \bibinfo{person}{Ari Shapiro}, {and} \bibinfo{person}{Hao Li}.}
  \bibinfo{year}{2019}\natexlab{}.
\newblock \showarticletitle{Learning perspective undistortion of portraits}. In
  \bibinfo{booktitle}{\emph{IEEE International Conference on Computer Vision
  (ICCV)}}. \bibinfo{pages}{7849--7859}.
\newblock


\bibitem[\protect\citeauthoryear{Zhu, Lei, Yan, Yi, and Li}{Zhu
  et~al\mbox{.}}{2015}]%
        {Zhu2015}
\bibfield{author}{\bibinfo{person}{Xiangyu Zhu}, \bibinfo{person}{Zhen Lei},
  \bibinfo{person}{Junjie Yan}, \bibinfo{person}{Dong Yi}, {and}
  \bibinfo{person}{Stan~Z Li}.} \bibinfo{year}{2015}\natexlab{}.
\newblock \showarticletitle{High-fidelity pose and expression normalization for
  face recognition in the wild}. In \bibinfo{booktitle}{\emph{IEEE Conference
  on Computer Vision and Pattern Recognition (CVPR)}}.
  \bibinfo{pages}{787--796}.
\newblock


\bibitem[\protect\citeauthoryear{Zollh{\"{o}}fer, Thies, Garrido, Bradley,
  Beeler, P{\'{e}}rez, Stamminger, Nie{\ss}ner, and Theobalt}{Zollh{\"{o}}fer
  et~al\mbox{.}}{2018}]%
        {Zollhoefer2018}
\bibfield{author}{\bibinfo{person}{Michael Zollh{\"{o}}fer},
  \bibinfo{person}{Justus Thies}, \bibinfo{person}{Pablo Garrido},
  \bibinfo{person}{Derek Bradley}, \bibinfo{person}{Thabo Beeler},
  \bibinfo{person}{Patrick P{\'{e}}rez}, \bibinfo{person}{Marc Stamminger},
  \bibinfo{person}{Matthias Nie{\ss}ner}, {and} \bibinfo{person}{Christian
  Theobalt}.} \bibinfo{year}{2018}\natexlab{}.
\newblock \showarticletitle{State of the Art on Monocular {3D} Face
  Reconstruction, Tracking, and Applications}.
\newblock \bibinfo{journal}{\emph{Computer Graphics Forum (Eurographics State
  of the Art Reports 2018)}} \bibinfo{volume}{37}, \bibinfo{number}{2}
  (\bibinfo{year}{2018}).
\newblock


\end{thebibliography}

{\small
%%% -*-BibTeX-*-
%%% Do NOT edit. File created by BibTeX with style
%%% ACM-Reference-Format-Journals [18-Jan-2012].

\balance
}

\newpage
\appendix
\section{Overview}

The supplemental material for our paper includes this document and a video.
The video provides an illustrated summary of the method as well as animation examples.
Here we provide implementation details and an extended qualitative evaluation.

\section{Implementation Details}

\qheading{Data:}
\modelname is trained on 2 Million images from \mbox{VGGFace2}~\cite{Cao2018_VGGFace2}, BUPT-Balancedface~\cite{Wang_2019_ICCV} and VoxCeleb2 \cite{Chung18b}.
From VGGFace2~\cite{Cao2018_VGGFace2}, we randomly select $950k$ images such that $750K$ images are of resolution higher than $224 \times 224$, and $200K$ are of lower resolution.
From BUPT-Balancedface~\cite{Wang_2019_ICCV} we randomly sample $550k$ with Asian or African ethnicity labels to reduce the ethnicity bias of VGGFace2. 
From VoxCeleb2~\cite{Chung18b} we choose $500k$ frames, with multiple samples from the same video clip per subject to obtain data with variation only in the facial expression and head pose.
We also sample $50k$ images from the VGGFace2~\cite{Cao2018_VGGFace2} test set for validation. 

\qheading{Data cleaning:}
We generate a different crop for the face image by shifting the provided bounding box by $5\%$ to the bottom right (i.e.~shift by $\boldsymbol{\epsilon}=\frac{1}{20}(b_w, b_h)^T$, where $b_w$ and $b_h$ denote the bounding box width and height).
Then we expand the original and the shifted bounding boxes by $10\%$ to the top, and by $20\%$ to the left, right, and bottom.
We run FAN~\cite{Bulat2017}, providing the expanded bounding boxes as input and discard all images with $\max \limits_{i}  \norm{\textbf{D}(\landmark_i^{2}-\boldsymbol{\epsilon}-\landmark_i^{1})} \geq 0.1$, where $\landmark_i^{2}$ and $\landmark_i^{1}$ are the $i$th landmarks for the original and the shifted bounding box, respectively, and $\textbf{D}$ denote the normalization matrix $diag(b_w, b_h)^{-1}$.

\qheading{Training details:}
We pre-train the coarse model (i.e.~$\coarseencoder$) for two epochs with a batch size of $64$ with $\lambda_{lmk} = 1e-4$, $\lambda_{eye} = 1.0$, $\lambda_{\shapecoeff} = 1e-4$, and $\lambda_{\expcoeff} = 1e-4$.
Then, we train the coarse model for $1.5$ epochs with a batch size of $32$, with $4$ images per subject with $\lambda_{pho} = 2.0$, $\lambda_{id} = 0.2$, $\lambda_{sc} = 1.0$, $\lambda_{lmk} = 1.0$, $\lambda_{eye} = 1.0$,  $\lambda_{\shapecoeff} = 1e-4$, and $\lambda_{\expcoeff} = 1e-4$.
The landmark loss uses different weights for individual landmarks, the mouth corners and the nose tip landmarks are weighted by a factor of $3$, other mouth and nose landmarks with a factor of $1.5$, and all remaining landmarks have a weight of $1.0$.
This is followed by training the detail model (i.e.~$\detailencoder$ and $\edm$) on VGGFace2 and VoxCeleb2 with a batch size of $6$, with $3$ images per subject, and parameters $\lambda_{phoD} = 2.0$, $\lambda_{mrf} = 5e-2$, $\lambda_{sym} = 5e-3$, $\lambda_{dc} = 1.0$, and $\lambda_{regD} = 5e-3$.
The coarse model is fixed while training the detail model.

\section{Evaluation}

\begin{figure*}[t]
	\begin{tabular}{c@{}c}
		\includegraphics[width=0.29\textwidth]{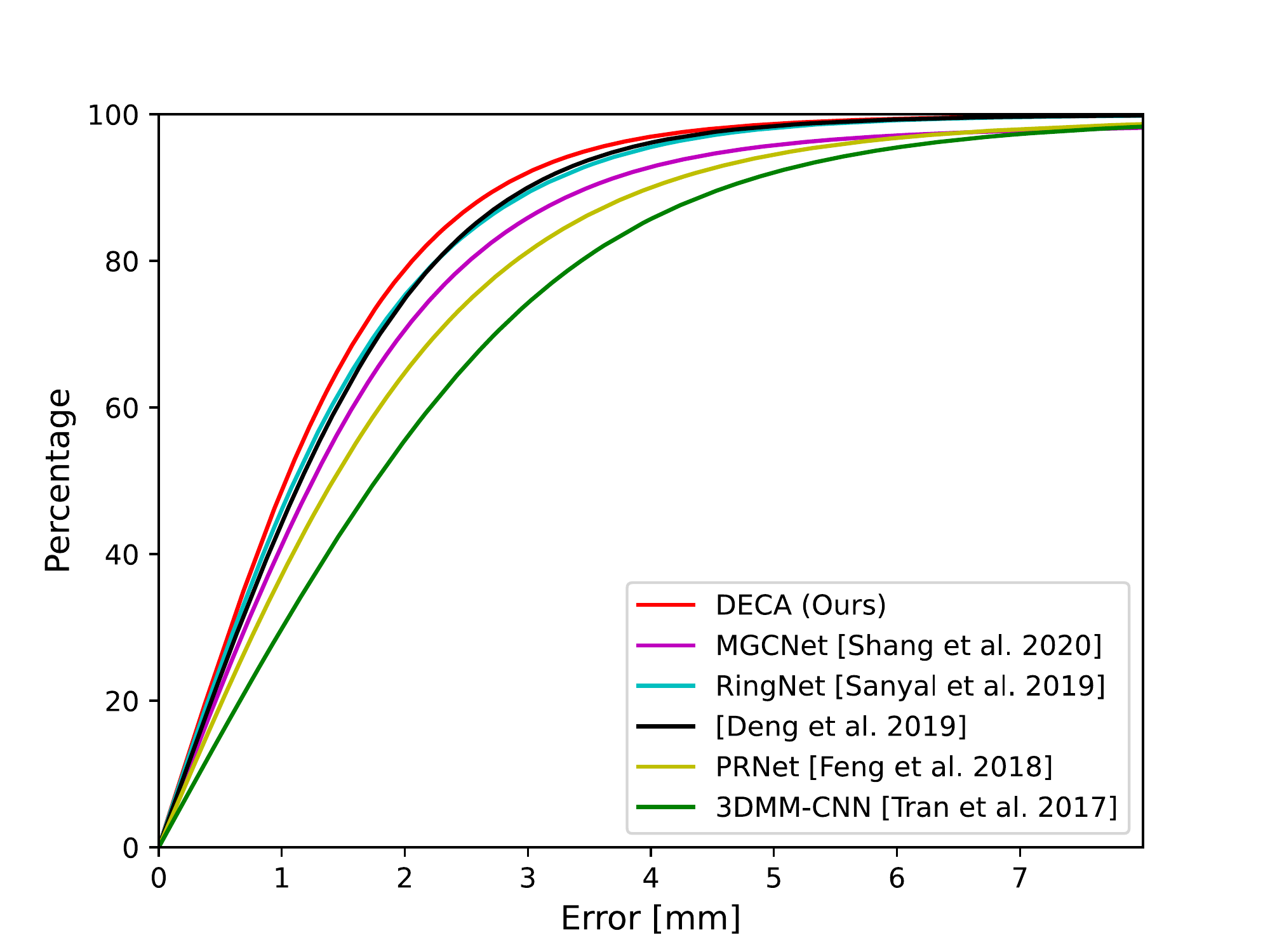} &  
		\includegraphics[width=0.29\textwidth]{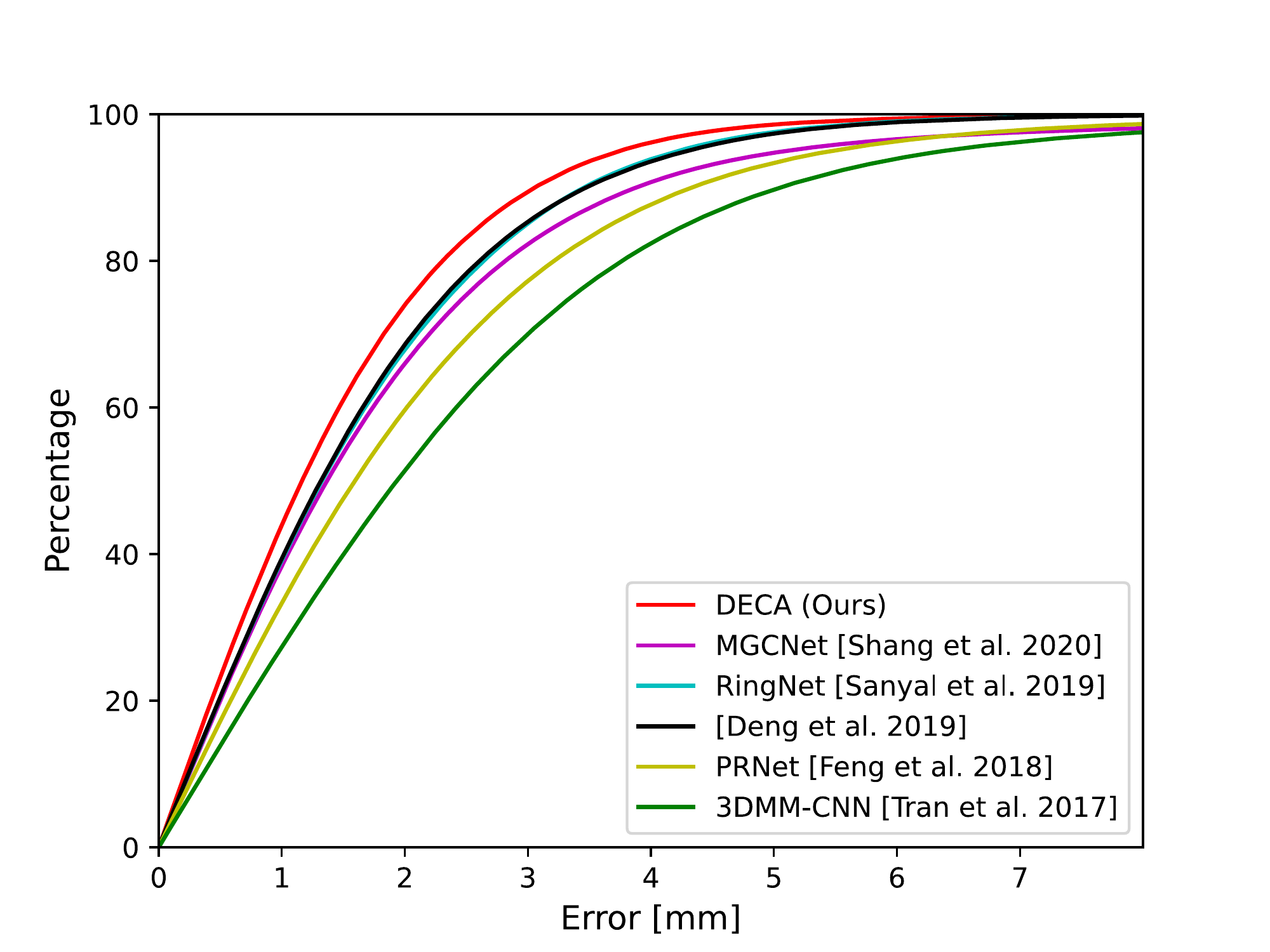} \\
		NoW (female)~\cite{Sanyal2019} & NoW (male)~\cite{Sanyal2019} 
	\end{tabular}
	\caption{Quantitative comparison to state-of-the-art on the NoW~\cite{Sanyal2019} challenge for female (left) and male (samples).
}
	\label{fig:cumulative_error_NoW_f_m}
\end{figure*}

\subsection{Qualitative comparisons}
\begin{figure*}[t]
    \offinterlineskip
    \includegraphics[width=0.97\textwidth]{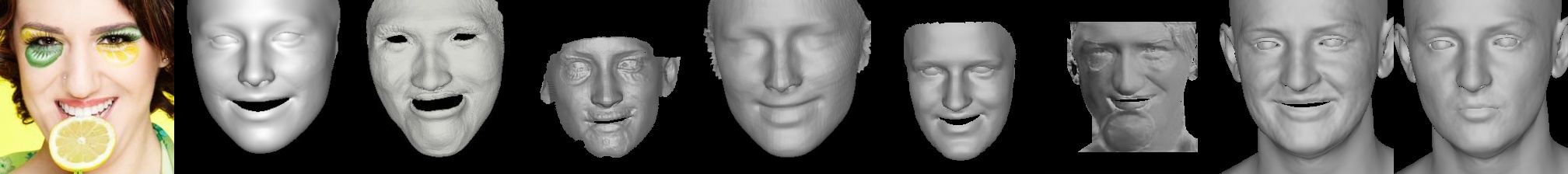}\\ 
    \includegraphics[width=0.97\textwidth]{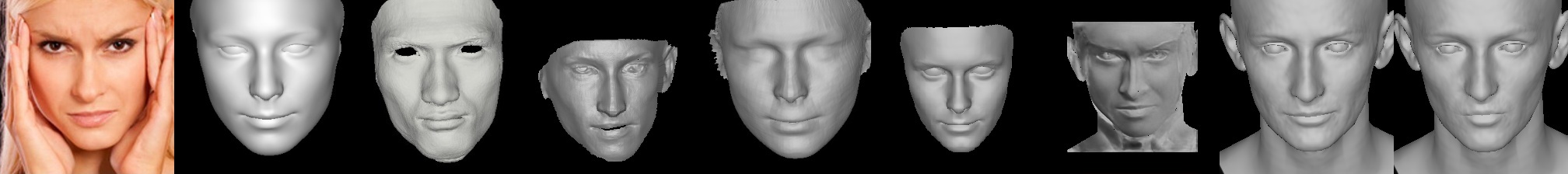}\\ 
    \includegraphics[width=0.97\textwidth]{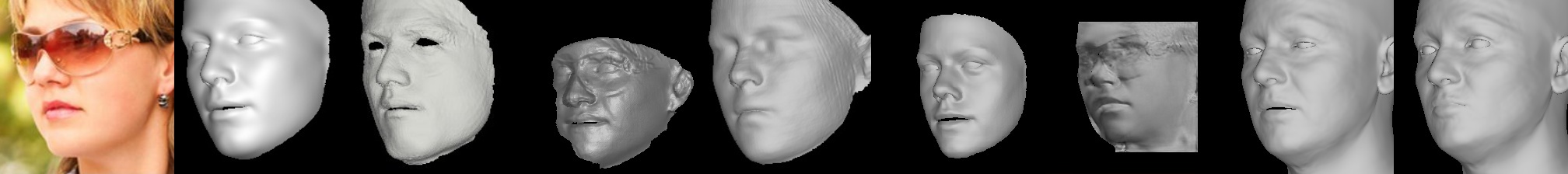}\\ 
    \includegraphics[width=0.97\textwidth]{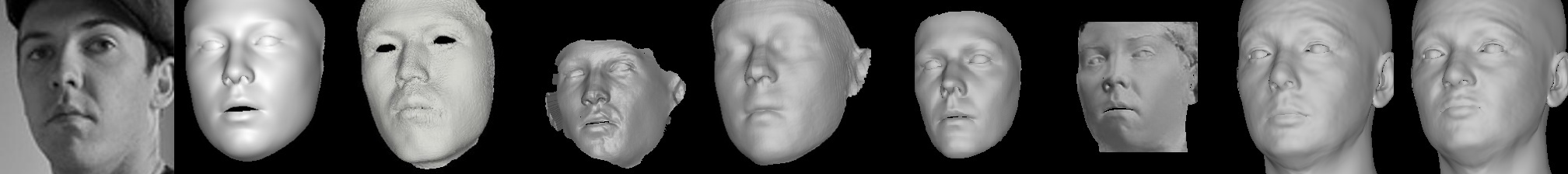}\\ 
    \includegraphics[width=0.97\textwidth]{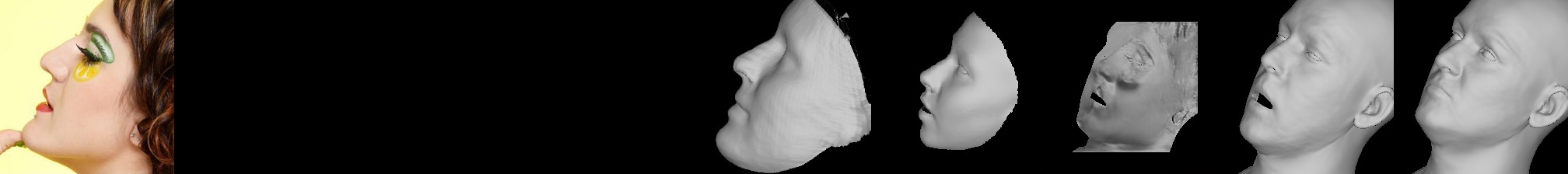}\\ 
    \includegraphics[width=0.97\textwidth]{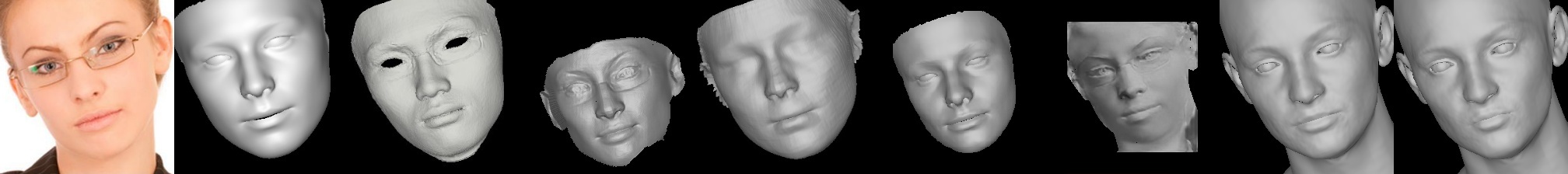}\\ 
    \includegraphics[width=0.97\textwidth]{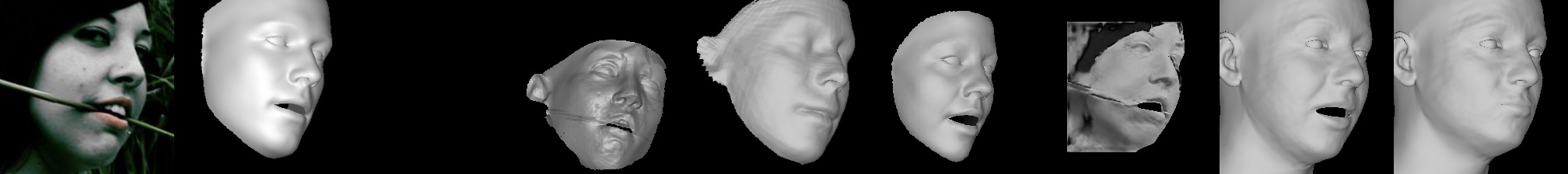} 
    \begin{tabular}{ccccccccc}
      \hspace{0.02\textwidth}  (a) \hspace{\varhspace\textwidth} & (b) \hspace{\varhspace\textwidth} & (c) \hspace{\varhspace\textwidth} & (d) \hspace{\varhspace\textwidth} & (e) \hspace{\varhspace\textwidth} & (f) \hspace{\varhspace\textwidth} & (g) \hspace{\varhspace\textwidth} & (h) \hspace{\varhspace\textwidth} & (i) \hspace{0.02\textwidth}  \\
    \end{tabular}
    \caption{Comparison to previous work, from left to right: (a) Input image, (b) 3DDFA-V2~\cite{guo2020towards}, (c) FaceScape~\cite{yang2020facescape}, (d) Extreme3D~\cite{AnhTran2018}, (e) PRNet~\cite{Feng2018}, (f) Deng et al.~\shortcite{Deng2019}, (g) Cross-modal~\cite{Abrevaya2020}, (h) \modelname detail reconstruction, and (i) reposing (animation) of \modelname's detail reconstruction to a common expression. Blank entries indicate that the particular method did not return any reconstruction.
    Input images are taken from ALFW2000~\cite{AFLW2011,Zhu2015}.
    }
    \label{fig:qualitative_all}
\end{figure*}

Figure~\ref{fig:qualitative_all} shows additional qualitative comparisons to existing coarse and detail reconstruction methods.
\modelname better reconstructs the overall face shape than all existing methods, it reconstructs more details than existing coarse reconstruction methods (e.g. (b), (e), (f)), and it is more robust to occlusions compared with existing detail reconstruction methods (e.g. (c), (d), (g)). 

\begin{figure*}[t]
    \includegraphics[width=\imsize\textwidth]{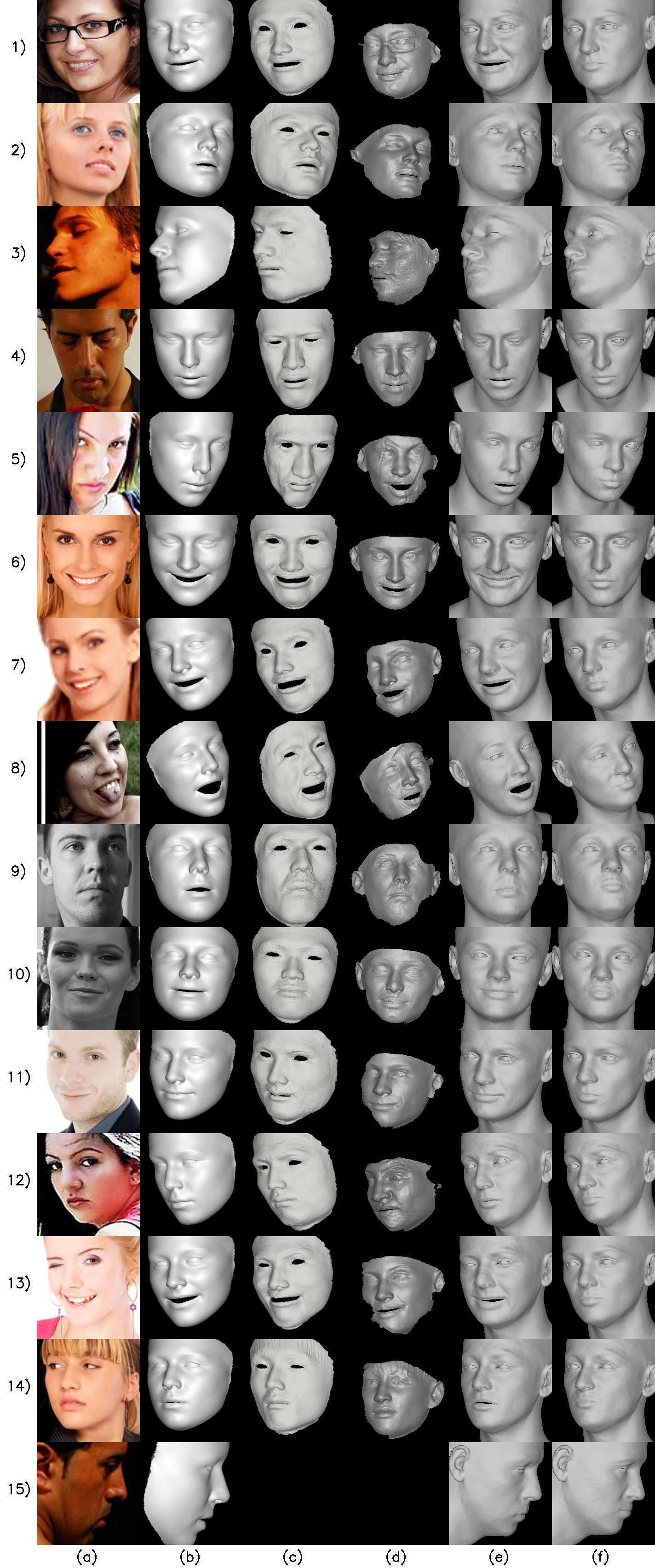}
    \includegraphics[width=\imsize\textwidth]{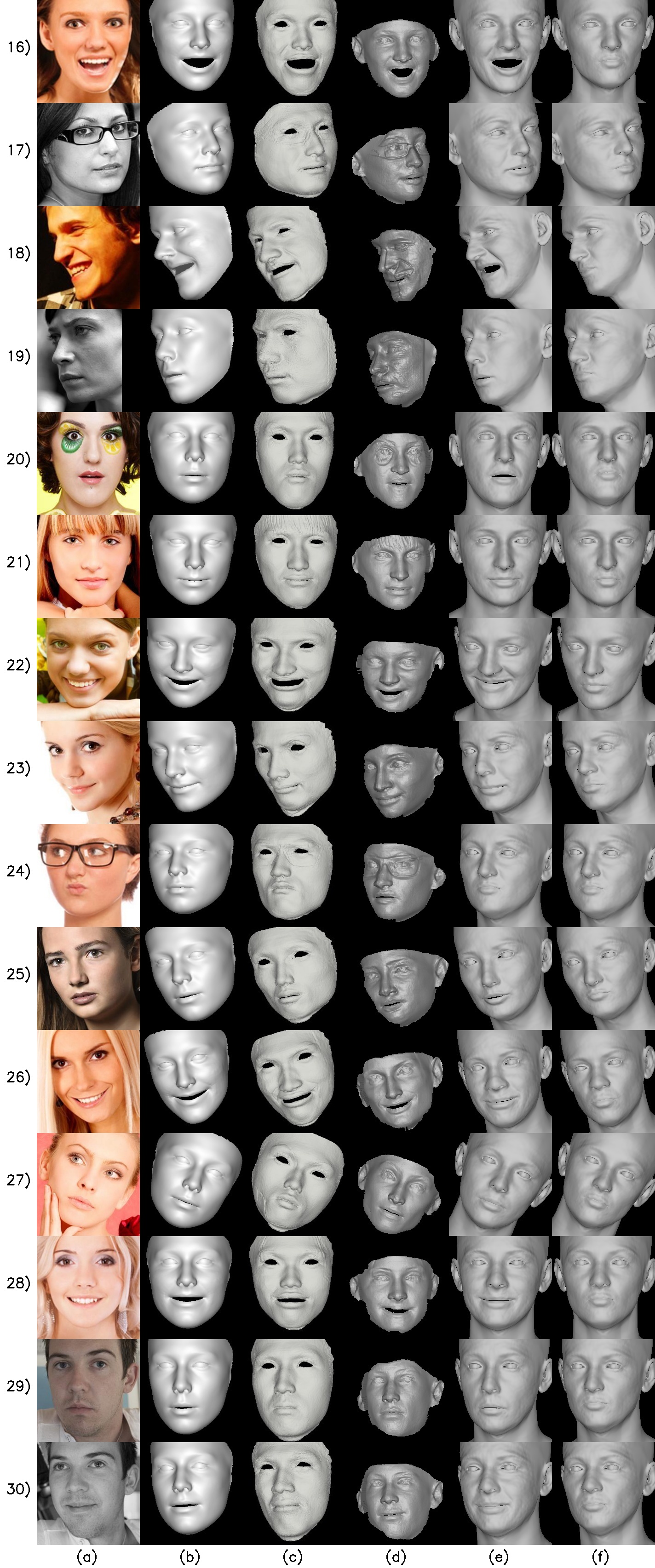}
    \caption{Qualitative comparisons on random ALFW2000~\cite{AFLW2011,Zhu2015} samples. a) Input, b) 3DDFA-V2~\cite{guo2020towards}, c) FaceScape~\cite{yang2020facescape}, d) Extreme3D~\cite{AnhTran2018}, e) \modelname detail reconstruction, and f) reposing (animation) of \modelname's detail reconstruction to a common expression. 
    Blank entries indicate that the particular method did not return any reconstruction.
    }
    \label{fig:ALFW1}
\end{figure*}

\begin{figure*}[t]
    \includegraphics[width=\imsize\textwidth]{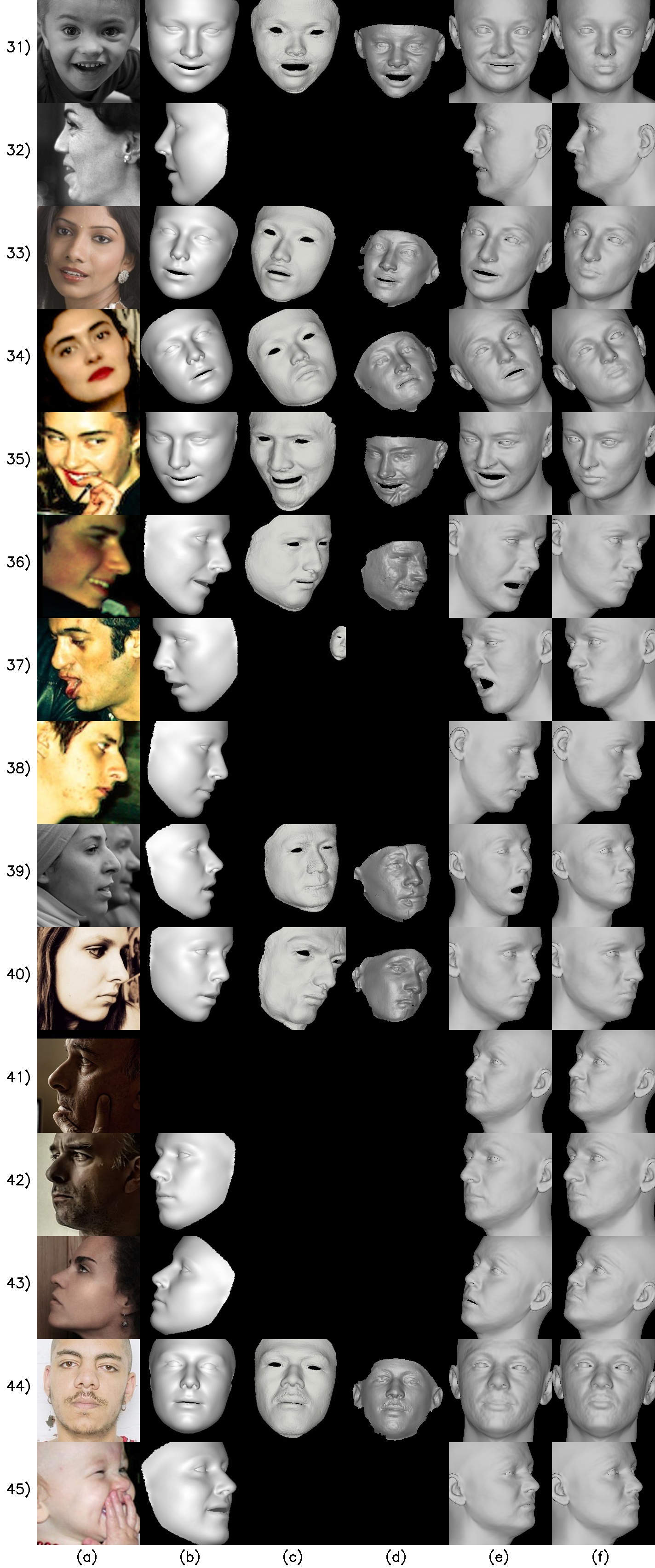}
    \includegraphics[width=\imsize\textwidth]{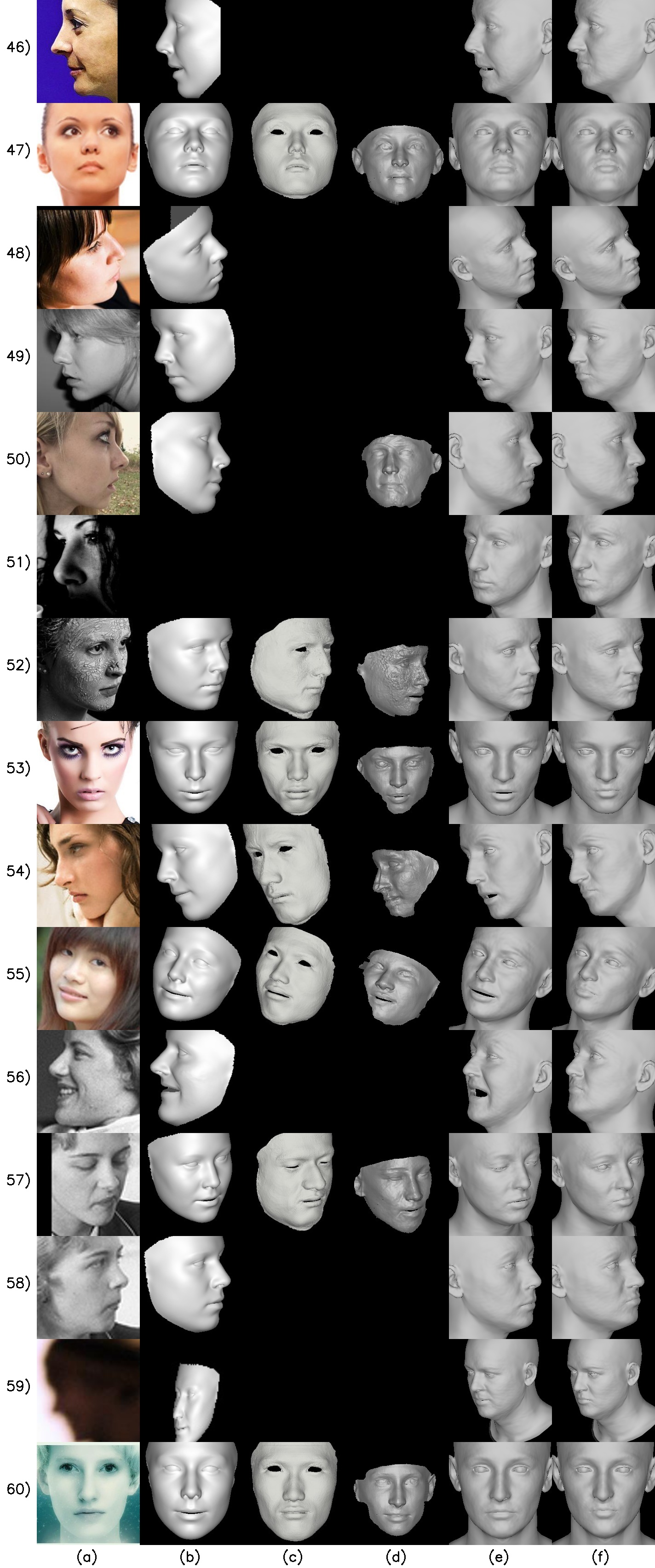}
    \caption{Qualitative comparisons on random ALFW2000~\cite{AFLW2011,Zhu2015} samples. a) Input, b) 3DDFA-V2~\cite{guo2020towards}, c) FaceScape~\cite{yang2020facescape}, d) Extreme3D~\cite{AnhTran2018}, e) \modelname detail reconstruction, and f) reposing (animation) of \modelname's detail reconstruction to a common expression. 
    Blank entries indicate that the particular method did not return any reconstruction.
    }
    \label{fig:ALFW2}
\end{figure*}

\begin{figure*}[t]
    \includegraphics[width=\imsize\textwidth]{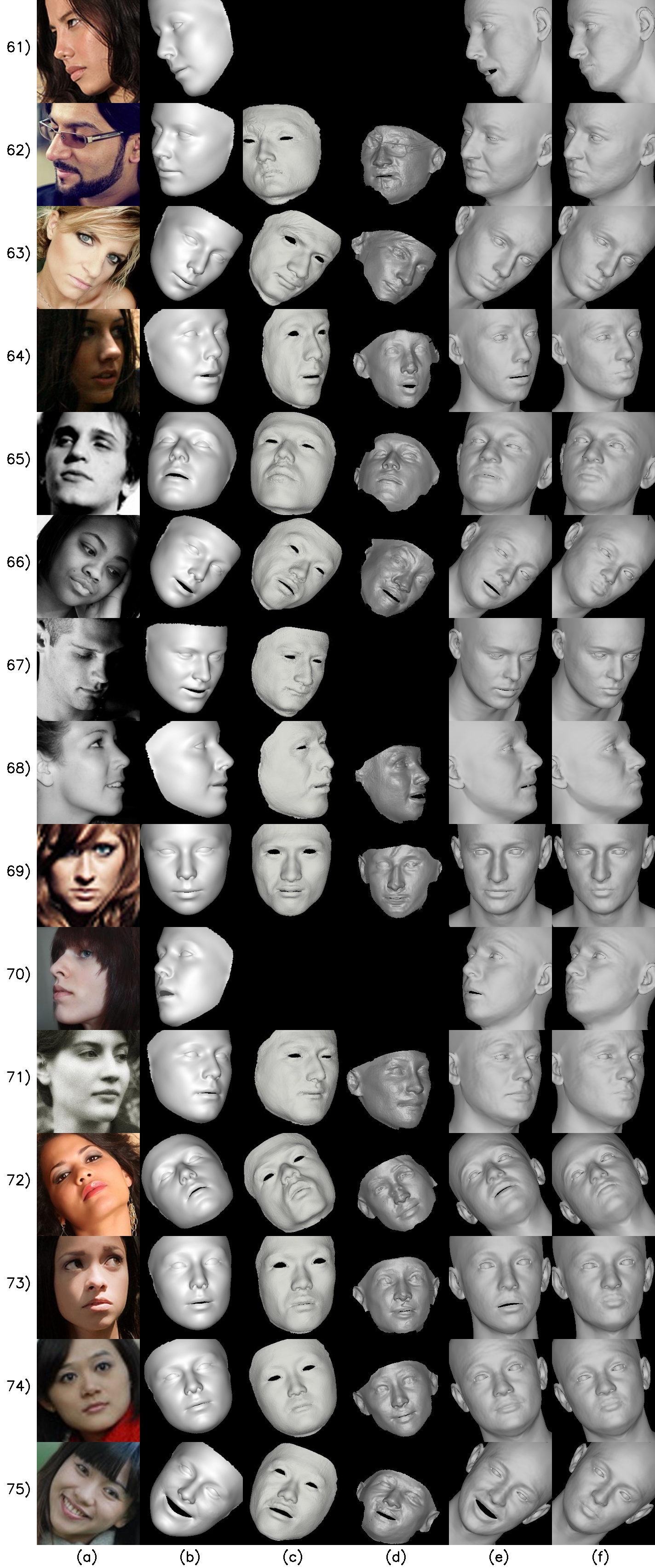}
    \includegraphics[width=\imsize\textwidth]{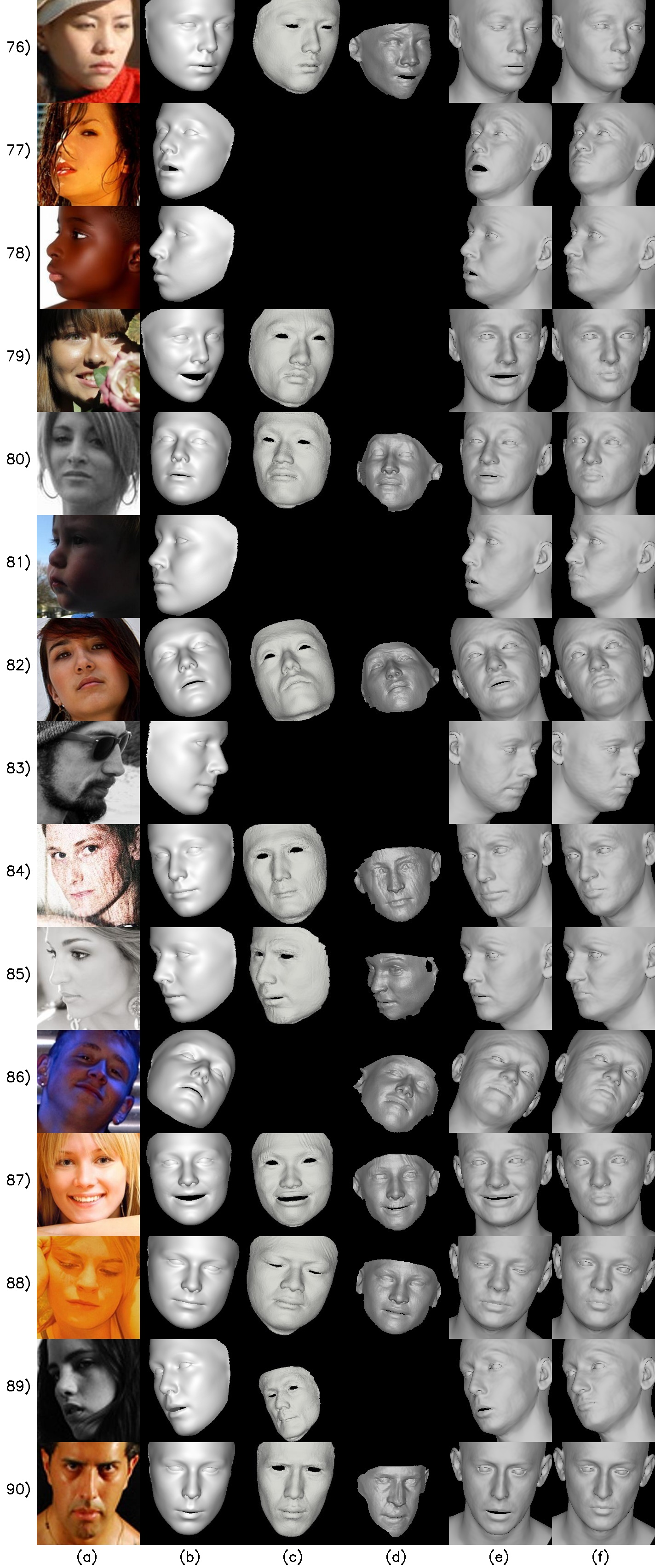}
    \caption{Qualitative comparisons on random ALFW2000~\cite{AFLW2011,Zhu2015} samples. a) Input, b) 3DDFA-V2~\cite{guo2020towards}, c) FaceScape~\cite{yang2020facescape}, d) Extreme3D~\cite{AnhTran2018}, e) \modelname detail reconstruction, and f) reposing (animation) of \modelname's detail reconstruction to a common expression. 
    Blank entries indicate that the particular method did not return any reconstruction.
    }
    \label{fig:ALFW3}    
\end{figure*}

\begin{figure*}[t]
    \includegraphics[width=\imsize\textwidth]{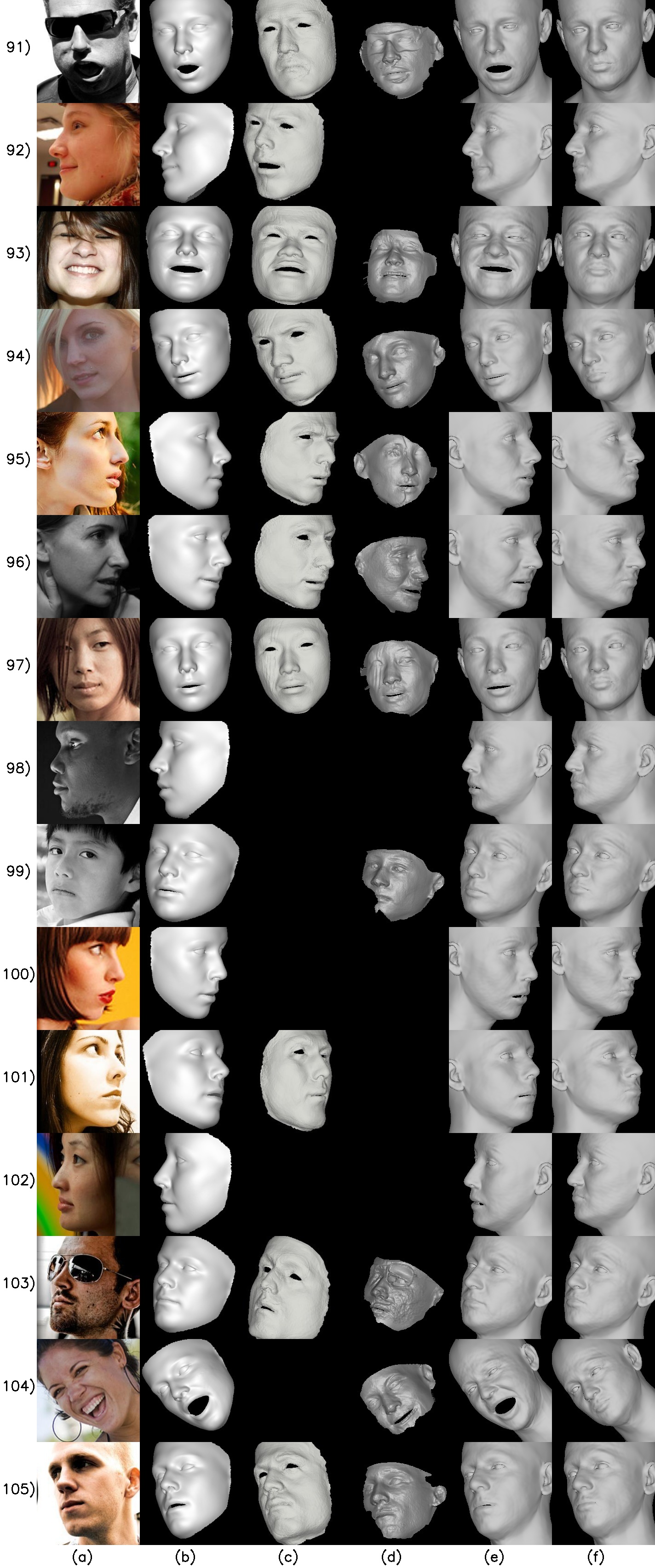}
    \includegraphics[width=\imsize\textwidth]{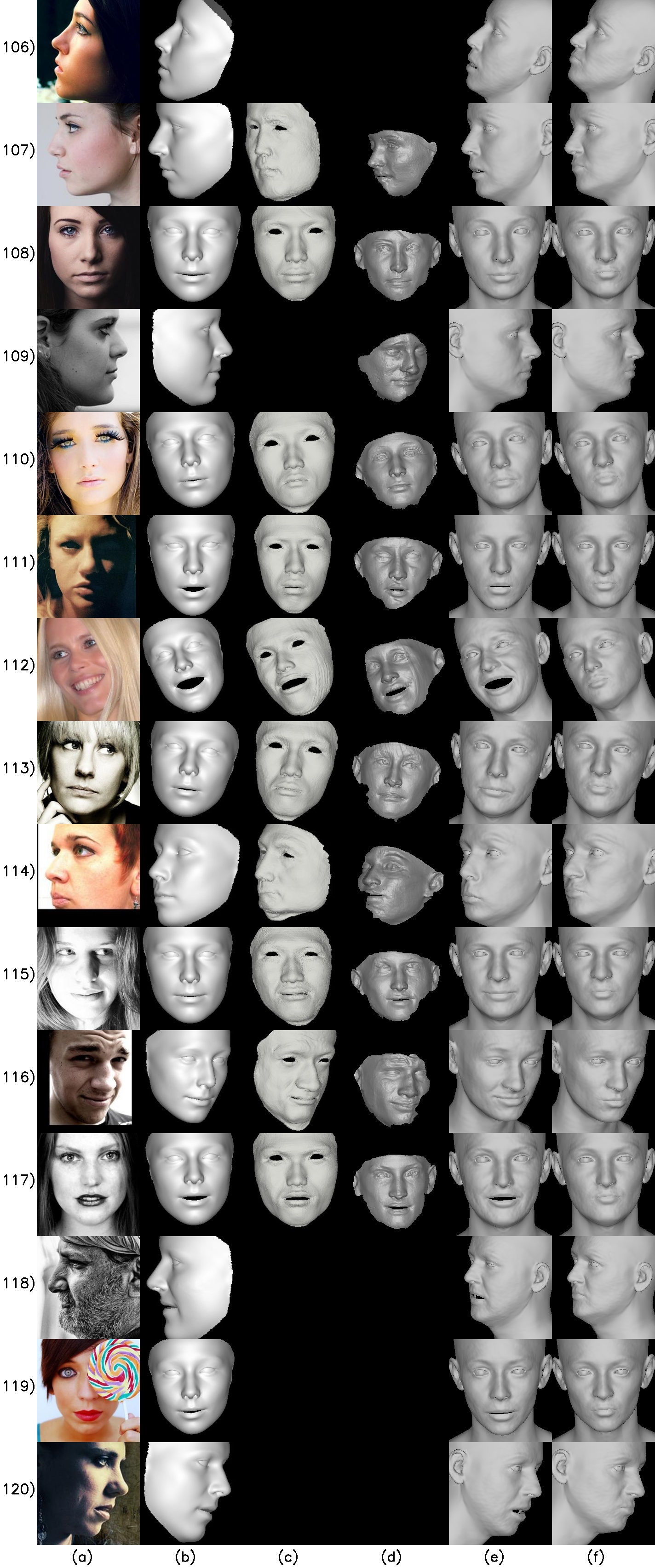}
    \caption{Qualitative comparisons on random ALFW2000~\cite{AFLW2011,Zhu2015} samples. a) Input, b) 3DDFA-V2~\cite{guo2020towards}, c) FaceScape~\cite{yang2020facescape}, d) Extreme3D~\cite{AnhTran2018}, e) \modelname detail reconstruction, and f) reposing (animation) of \modelname's detail reconstruction to a common expression. 
    Blank entries indicate that the particular method did not return any reconstruction.
    }
    \label{fig:ALFW4}    
\end{figure*}

\begin{figure*}[t]
    \includegraphics[width=\imsize\textwidth]{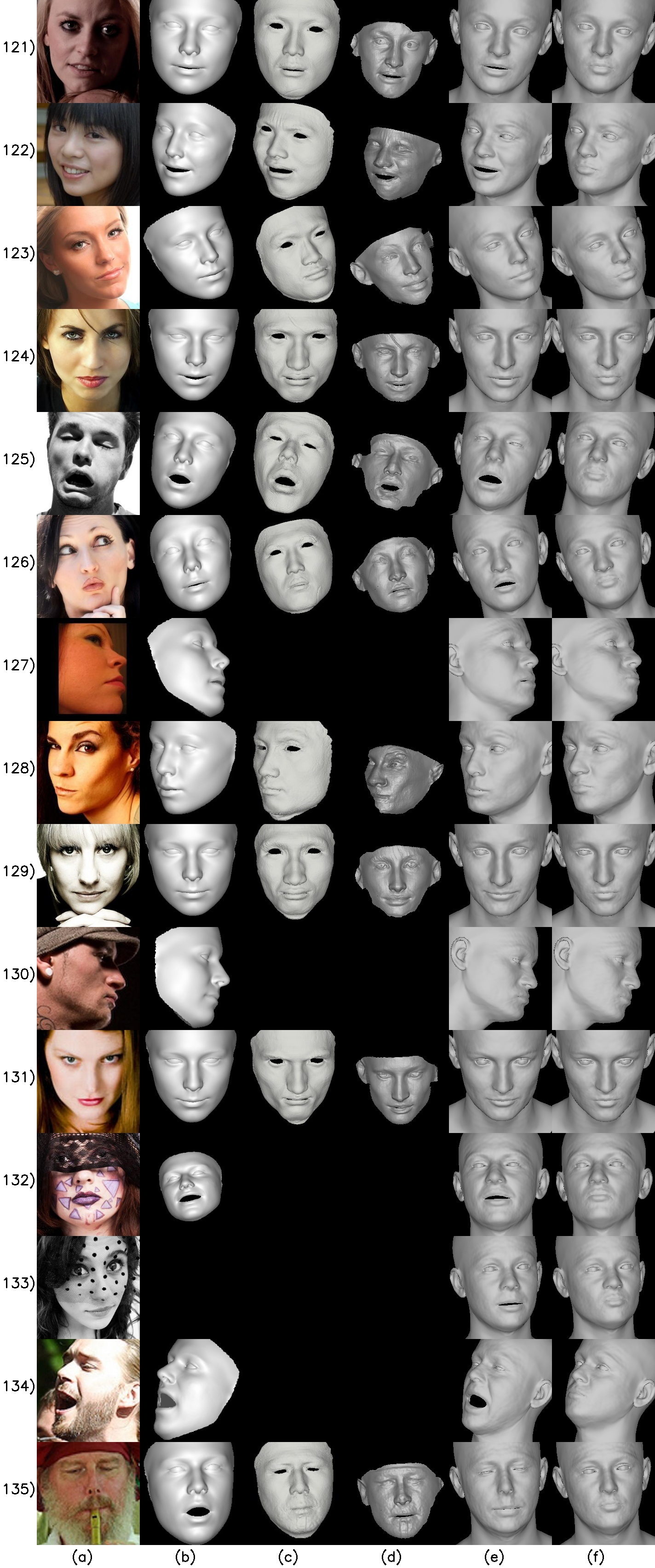}
    \includegraphics[width=\imsize\textwidth]{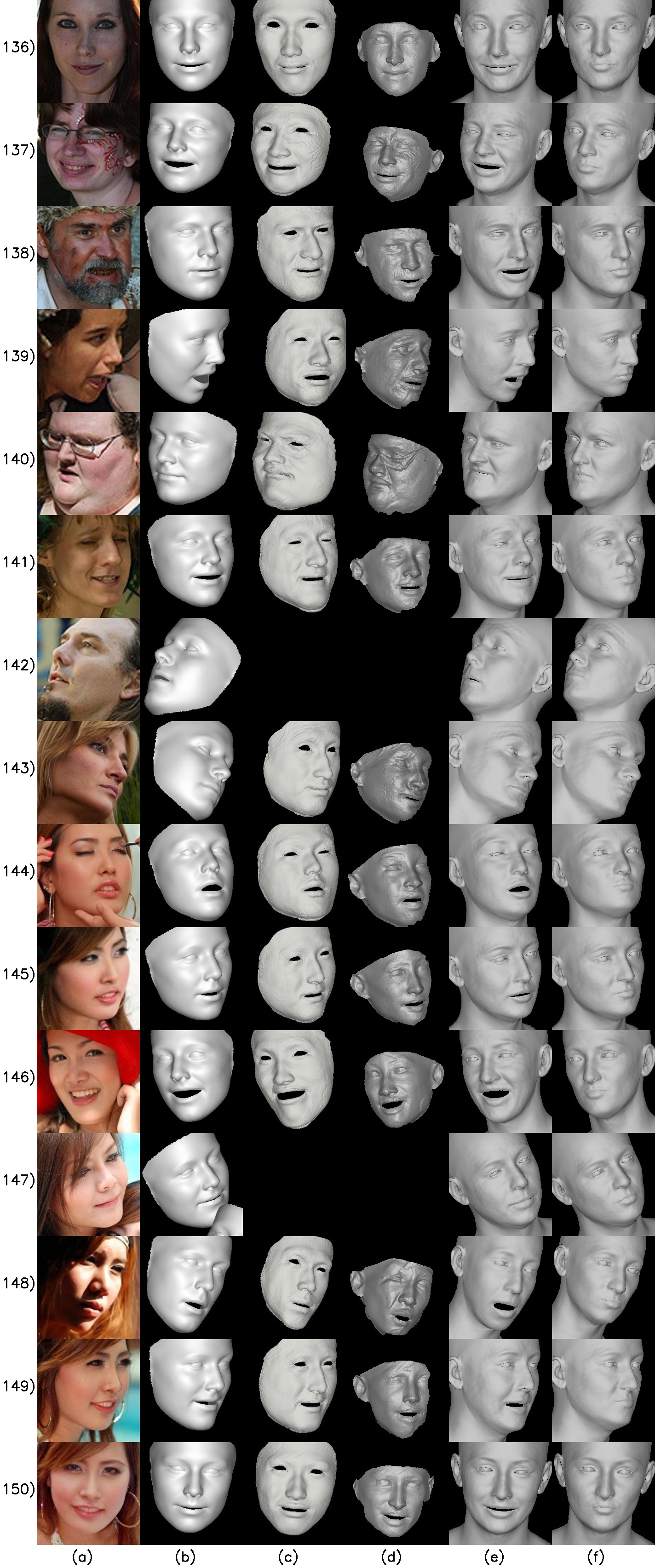}
    \caption{Qualitative comparisons on random ALFW2000~\cite{AFLW2011,Zhu2015} samples. a) Input, b) 3DDFA-V2~\cite{guo2020towards}, c) FaceScape~\cite{yang2020facescape}, d) Extreme3D~\cite{AnhTran2018}, e) \modelname detail reconstruction, and f) reposing (animation) of \modelname's detail reconstruction to a common expression. 
    Blank entries indicate that the particular method did not return any reconstruction.
    }
    \label{fig:ALFW5}
\end{figure*}

\begin{figure*}[t]
    \includegraphics[width=\imsize\textwidth]{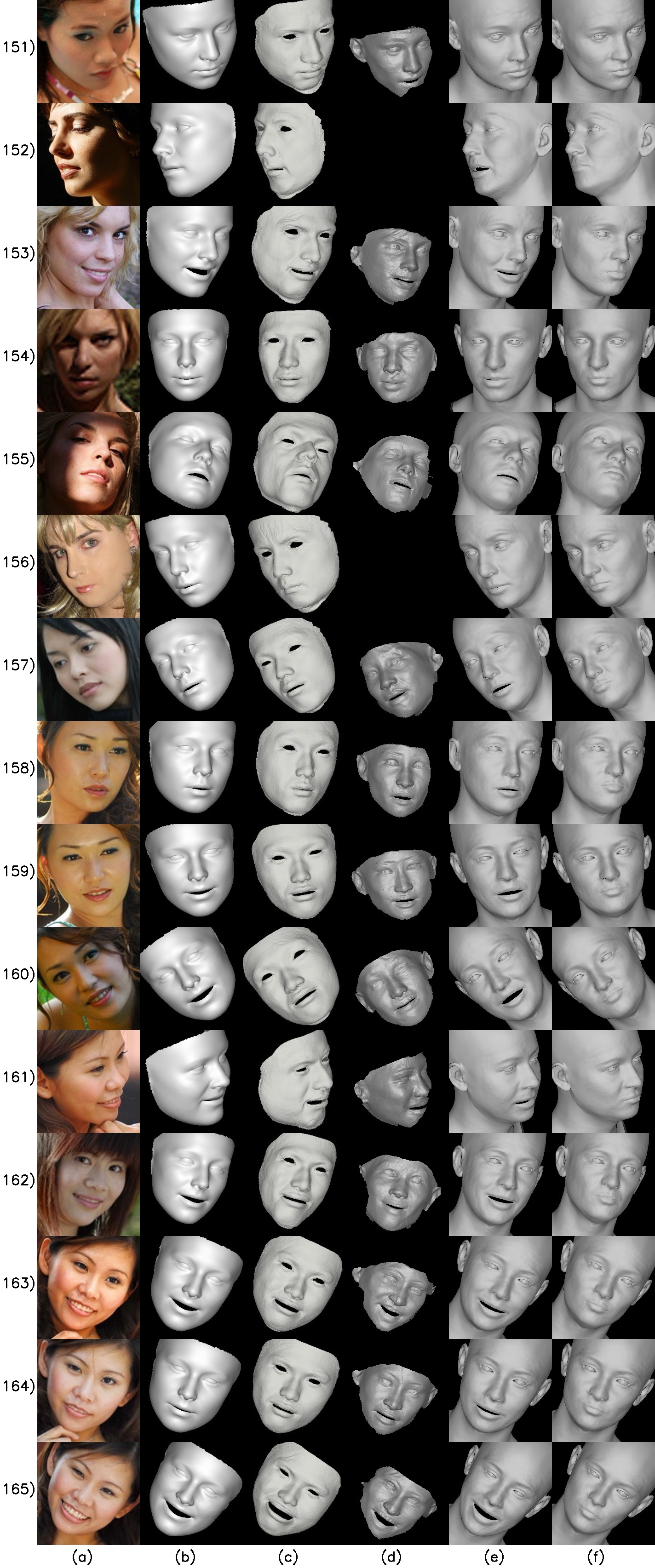}
    \includegraphics[width=\imsize\textwidth]{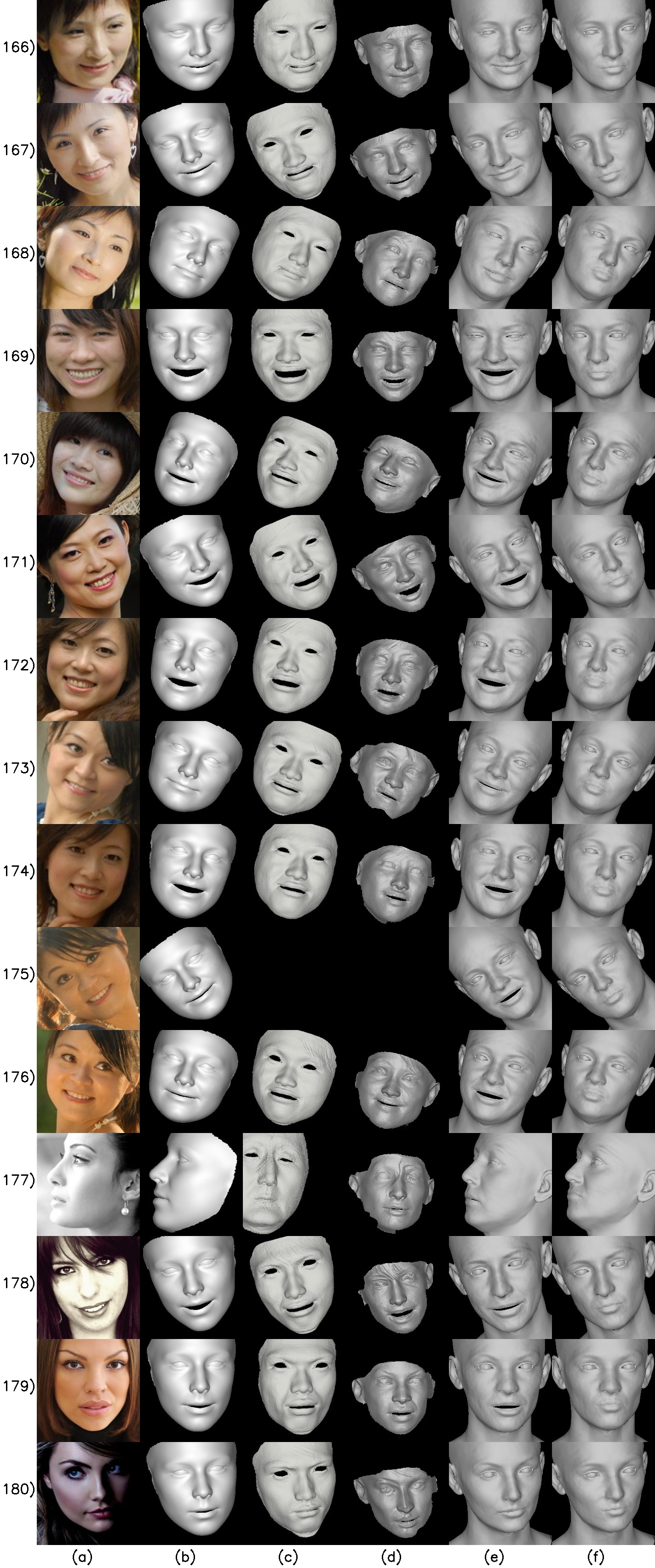}
    \caption{Qualitative comparisons on random ALFW2000~\cite{AFLW2011,Zhu2015} samples. a) Input, b) 3DDFA-V2~\cite{guo2020towards}, c) FaceScape~\cite{yang2020facescape}, d) Extreme3D~\cite{AnhTran2018}, e) \modelname detail reconstruction, and f) reposing (animation) of \modelname's detail reconstruction to a common expression. 
    Blank entries indicate that the particular method did not return any reconstruction.
    }
    \label{fig:ALFW6}
\end{figure*}

\begin{figure*}[t]
    \includegraphics[width=\imsize\textwidth]{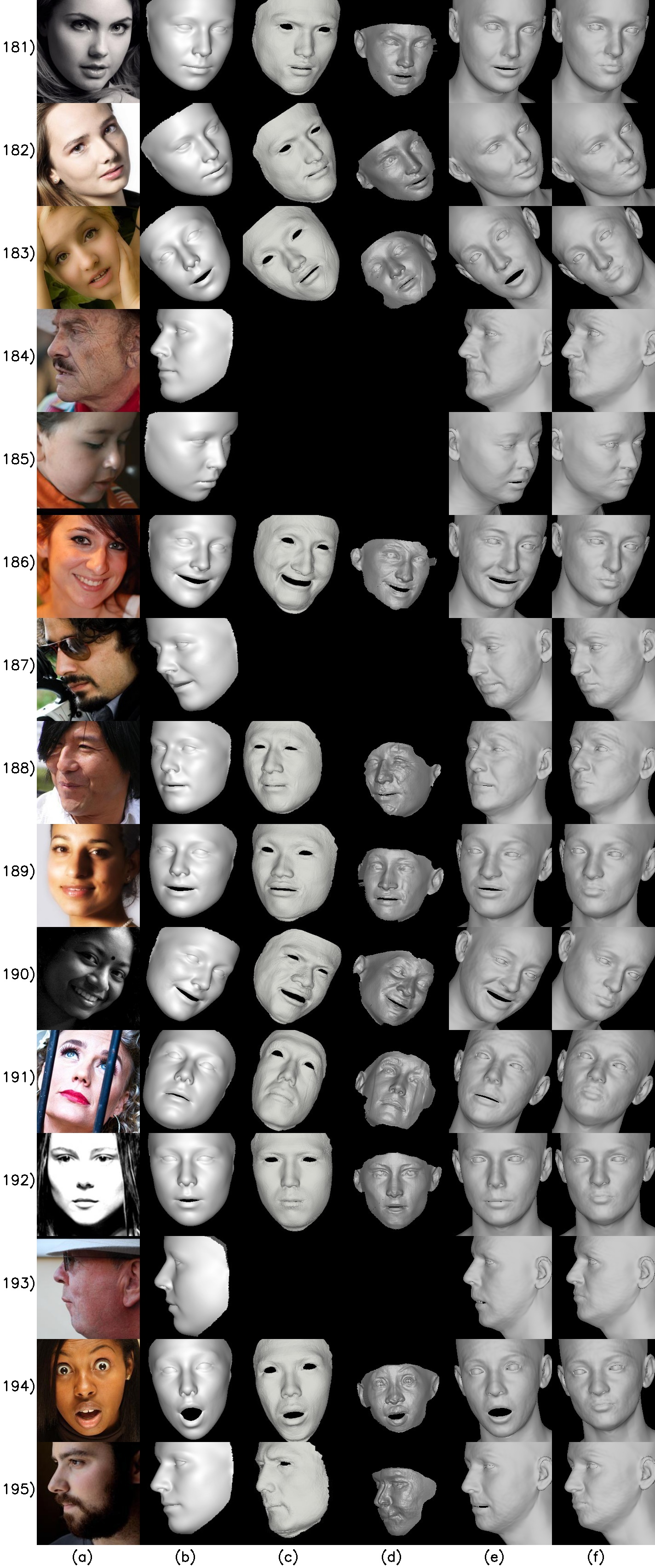}
    \includegraphics[width=\imsize\textwidth]{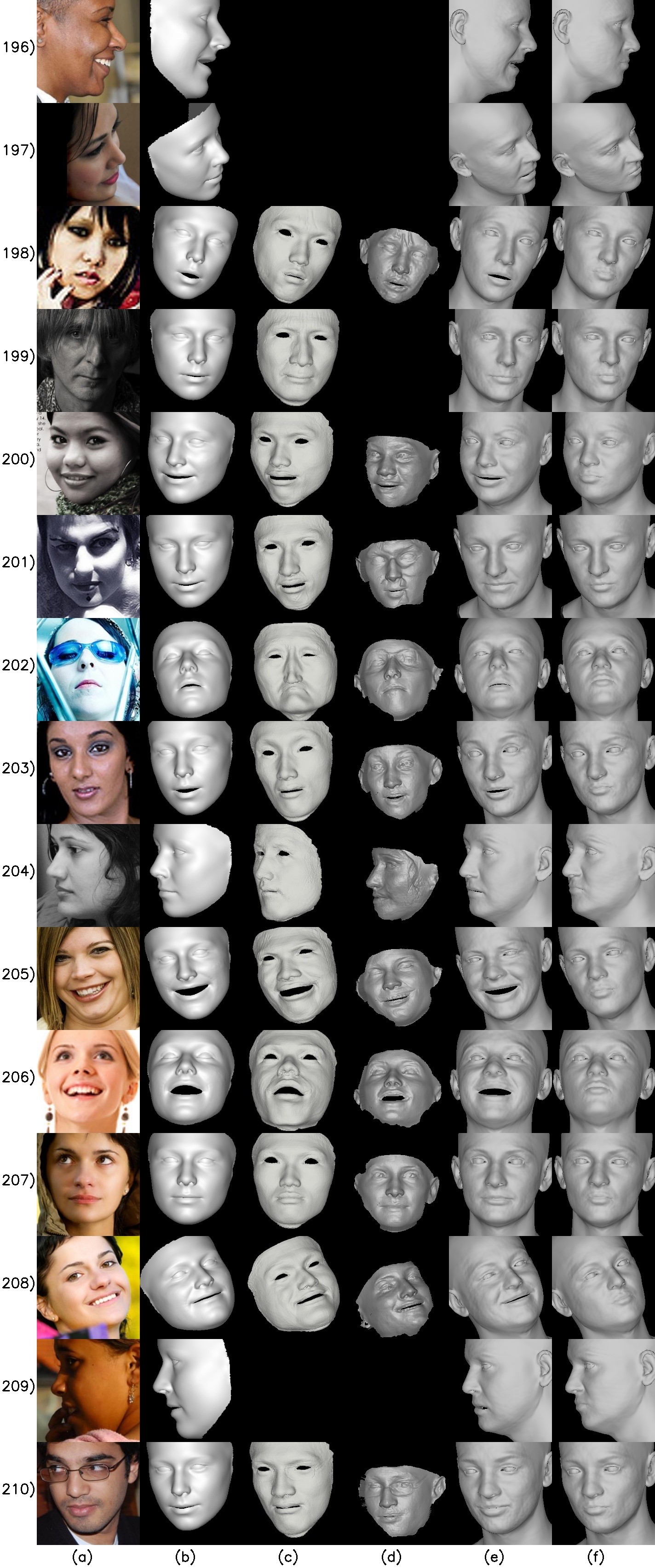}
    \caption{Qualitative comparisons on random ALFW2000~\cite{AFLW2011,Zhu2015} samples. a) Input, b) 3DDFA-V2~\cite{guo2020towards}, c) FaceScape~\cite{yang2020facescape}, d) Extreme3D~\cite{AnhTran2018}, e) \modelname detail reconstruction, and f) reposing (animation) of \modelname's detail reconstruction to a common expression. 
    Blank entries indicate that the particular method did not return any reconstruction.
    }
    \label{fig:ALFW7}
\end{figure*}

As promised in the main paper (e.g.~Section 6.1), we show results for more than 200 randomly selected ALFW2000~\cite{Zhu2015} samples in Figures~\ref{fig:ALFW1}, \ref{fig:ALFW2}, \ref{fig:ALFW3}, \ref{fig:ALFW4}, \ref{fig:ALFW5}, \ref{fig:ALFW6}, and \ref{fig:ALFW7}.
For each sample, we compare \modelname's detail reconstruction (e) with the state-of-the-art coarse reconstruction method 3DDFA-V2~\cite{guo2020towards} (see (b)) and existing detail reconstruction methods, namely FaceScape~\cite{yang2020facescape} (see (c)), and Extreme3D~\cite{AnhTran2018} (see (e)).
In total, \modelname reconstructs more details then 3DDFA-V2, and it is more robust to occlusions than FaceScape and Extreme3D. 
Further, the \modelname retargeting results appear realistic.

\end{document}